\documentclass{article}

\PassOptionsToPackage{numbers,compress}{natbib}
\PassOptionsToPackage{dvipsnames,table}{xcolor}
\usepackage[main,final]{neurips_2026}

\makeatletter
\let\hydraOriginalNotice\@noticestring
\renewcommand{\@noticestring}{\phantom{\hydraOriginalNotice}}
\makeatother

\usepackage[utf8]{inputenc}
\usepackage[T1]{fontenc}
\usepackage{microtype}
\usepackage{xspace}
\usepackage{nicefrac}

\usepackage{amsmath,amssymb,mathtools,amsthm}
\usepackage{bm}

\usepackage{amsmath,amsfonts,bm}

\def\eqref#1{equation~\ref{#1}}

\def\1{\bm{1}}

\DeclareMathAlphabet{\mathsfit}{\encodingdefault}{\sfdefault}{m}{sl}
\SetMathAlphabet{\mathsfit}{bold}{\encodingdefault}{\sfdefault}{bx}{n}

\usepackage{xcolor}
\usepackage{pifont}

\usepackage{graphicx}
\usepackage{float}
\usepackage{wrapfig}
\usepackage{subcaption} 
\usepackage{pgfplots}

\usepackage{array}
\usepackage{booktabs}
\usepackage{multirow}
\usepackage{diagbox}
\usepackage{threeparttable}
\usepackage{siunitx}
\usepackage{tabu}

\usepackage{enumitem}
\usepackage{balance}

\usepackage[disable,textsize=tiny]{todonotes}

\usepackage{url}
\usepackage{hyperref}
\hypersetup{
  colorlinks=true,
  linkcolor=purple,
  citecolor=blue,
  urlcolor=BrickRed
}
\usepackage[capitalize,noabbrev]{cleveref}

\theoremstyle{plain}

\theoremstyle{definition}

\theoremstyle{remark}

\newcolumntype{I}{S[table-format=2.0]}
\newcolumntype{R}{S[table-format=1.2]}
\newcolumntype{A}{S[table-format=1.3]}
\newcolumntype{M}{S[table-format=1.3]}

\definecolor{HydraRow}{RGB}{240,240,240}
\definecolor{HdrBand}{RGB}{245,245,245}

\newcommand{\metriclegend}{%
  {\footnotesize
   \emph{Arrows indicate directionality:}
   $\uparrow$ higher is better, $\downarrow$ lower is better.}%
}

\newcommand{\G}{\mathcal{G}}
 
\newcommand{\methodname}{\textsc{Hydra}\xspace}
\title{Hydra: Towards Transferable\\
Multi-Task Learning on Temporal Graphs}

\author{
Kiarash Shamsi\textsuperscript{1}\quad
\textbf{Farimah Poursafaei}\textsuperscript{2,3} \quad
Tran Gia Bao Ngo\textsuperscript{1} \\
\textbf{Reihaneh Rabbany}\textsuperscript{2,3,6} \quad
\textbf{Baris Coskunuzer}\textsuperscript{4} \quad
\textbf{Guillaume Rabusseau}\textsuperscript{2,5,6} \\
\textbf{Michael M. Bronstein}\textsuperscript{7} \quad
\textbf{Shenyang Huang}\textsuperscript{2,3,7} \quad
\textbf{Cuneyt Gurcan Akcora}\textsuperscript{8} \\
\textsuperscript{1}University of Manitoba, 
\textsuperscript{2}Mila -- Quebec AI Institute, \\
\textsuperscript{3}McGill University,
\textsuperscript{4}University of Texas at Dallas, 
\textsuperscript{5}Universit\'e de Montr\'eal,
\textsuperscript{6}CIFAR AI Chair\\
\textsuperscript{7}University of Oxford,
\textsuperscript{8}University of Central Florida
}

\begin{document}

\maketitle

\newif\ifCRcolors
\CRcolorstrue 
\definecolor{CRorange}{RGB}{190,85,0}
\definecolor{CRgreen}{RGB}{0,115,65}
\definecolor{CRpurple}{RGB}{125,55,165}
\DeclareRobustCommand{\crone}[1]{{\ifCRcolors\color{black}\fi #1}}
\DeclareRobustCommand{\crtwo}[1]{{\ifCRcolors\color{black}\fi #1}}
\DeclareRobustCommand{\crthree}[1]{{\ifCRcolors\color{black}\fi #1}}
\newcommand{\CRtwocolor}{\ifCRcolors\color{black}\fi}
\newcommand{\CRthreecolor}{\ifCRcolors\color{black}\fi}

\newcommand{\ah}[1]{{{\textcolor{blue}{\textbf{AH: }}}{\textcolor{orange}{#1}}}}
\newcommand{\ks}[1]{{{\textcolor{blue}{\textbf{KS: }}}{\textcolor{teal}{#1}}}}
\newcommand{\bao}[1]{{{\textcolor{blue}{\textbf{Bao: }}}{\textcolor{blue}{#1}}}}
\newcommand{\fp}[1]{{{\textcolor{blue}{\textbf{FP: }}}{\textbf{\textcolor{violet}{#1}}}}}
\newcommand{\gr}[1]{{{\textcolor{red}{\textbf{GR: }}}{\textcolor{red}{#1}}}}
\newcommand{\BC}[1]{{{\textcolor{blue}{\textbf{BC: }}}{\bf \textcolor{magenta}{#1}}}}

\newcommand{\etc}{et al.\xspace}
\renewcommand{\vec}[1]{\ensuremath{\mathbf{#1}}}
\newcommand{\vecs}[1]{\ensuremath{\mathbf{\boldsymbol{#1}}}}
\newcommand{\mat}[1]{\ensuremath{\mathbf{#1}}}
\newcommand{\mats}[1]{\ensuremath{\mathbf{\boldsymbol{#1}}}}
\newcommand{\ten}[1]{\mat{\ensuremath{\boldsymbol{\mathcal{#1}}}}}

\newcommand{\vectorize}[1]{\mathrm{vec}\left(#1\right)}
\newcommand{\rank}[1]{\mathrm{rank}(#1)}
%
\newcommand{\kron}{\otimes}
\newcommand{\krao}{\odot}
\newcommand{\outprod}{\circ}
\newcommand{\mlrank}[1]{\mathrm{rank}_{\textit{ml}}(#1)}
\newcommand{\cprank}[1]{\mathrm{rank}_{\textit{CP}}(#1)}
\newcommand{\ttm}[1]{\times_{#1}}
\newcommand{\ttv}[1]{\bullet_{#1}}
\newcommand{\tenmat}[2]{#1_{(#2)}}
\newcommand{\CP}[1]{\llbracket #1 \rrbracket}
\newcommand{\Tucker}[1]{\llbracket #1 \rrbracket}
\newcommand{\TT}[1]{\llangle #1 \rrangle}
\newtheorem{definition}{Definition}

\newcommand{\clds}[1]{\colorbox[HTML]{e7d4e8}{#1}}
\newcommand{\rgds}[1]{\colorbox[HTML]{d9f0d3}{#1}}
\newcommand{\rgs}[1]{%
  \begingroup
  \setlength{\fboxsep}{1pt}%
  \colorbox[HTML]{d9f0d3}{#1}%
  \endgroup
}
\newcommand{\cls}[1]{%
  \begingroup
  \setlength{\fboxsep}{1pt}%
  \colorbox[HTML]{e7d4e8}{#1}%
  \endgroup
}

\newcommand{\revOne}[1]{{{\textcolor{black}{#1}}}}

\newcommand{\revTwo}[1]{{{\textcolor{black}{#1}}}}

\newcommand{\revThree}[1]{{{\textcolor{black}{#1}}}}

\newcommand{\revFour}[1]{{{\textcolor{black}{#1}}}}
\begin{abstract}

Real-world evolving networks are naturally modeled as temporal graphs (TGs), where capturing temporal dynamics is essential for predicting future graph properties that support downstream decision-making. Existing temporal graph methods are primarily designed for a single task while requiring retraining from scratch for each dataset. As a result, little is known about their generalization across tasks or transfer to unseen networks.
In this work, we aim to fill this gap by studying the multi-task graph property prediction setting in TGs. The evolution of graph properties is best captured through a local perspective that models evolving patterns at individual nodes and a global perspective that emphasizes high-level structural changes. To this end, we introduce \methodname, a novel architecture that integrates local connectivity features from temporal GNNs with a spectral learning module that captures global connectivity patterns. We then optimize these two perspectives under a multi-task objective.
In multi-task classification, \methodname achieves an 8.9\% relative gain in AUC over the strongest competitor. In multi-task regression, \methodname obtains the best result in two of three tasks, with an 8.2\% relative gain in MAE over the second-best method, while maintaining competitive performance across all three tasks. By achieving consistent top-ranked performance, \methodname highlights multi-task training on temporal graphs as a promising direction toward adaptable foundation models for temporal graphs. Our code is available at: \href{https://github.com/kiarashamsi/Hydra}{https://github.com/kiarashamsi/Hydra}. 
\end{abstract}

\section{Introduction}
\label{sec:introduction}
Temporal graph prediction addresses the challenge of forecasting the future state and the trajectory of networks that evolve over time and possess an underlying relational structure. Unlike traditional graph analysis that assumes a static topology, temporal graphs are characterized by continuously changing nodes, edges, and features, representing dynamic real-world phenomena such as social network interactions \citep{wang2024temporal,jalili2025the}, traffic flow \citep{sahili2023spatio, ockerman2025pgtiscalingspatiotemporalgnns}, and biological processes \citep{hosseinzadeh2022temporal}. The core difficulty lies in capturing both the complex spatial dependencies encoded in the graph structure and the temporal dynamics that govern its evolution \citep{kazemi2020representation}. Effective solutions must therefore move beyond static graph representation learning to models that can jointly capture spatio-temporal patterns, enabling accurate prediction of future interactions, node properties, and system-wide events. This capability underpins applications such as disease outbreak detection \citep{senthilkumar2022dynamic}, fraud prevention \citep{duan2024cat}, recommendation systems \citep{tang2025tcgc}, and transportation network planning \citep{chen2022graph, uddin2026t3former}. Despite progress in temporal graph learning, two fundamental challenges remain open.

\textbf{Multi-task Temporal Graph Prediction.} The first challenge is multi-task temporal graph property prediction. Existing temporal graph methods commonly focus on one prediction objective at a time, while many real applications require forecasting multiple related graph-level properties from the same evolving network. For example, one may need to predict both community evolution and connectivity growth. By learning these tasks jointly, models can exploit the shared spatio-temporal dynamics that shape them, yielding more robust generalization and a deeper understanding of system-wide behavior. 
\revFour{Multi-task temporal graph prediction requires representations that transfer across heterogeneous graphs, remain stable under temporal drift, and still capture task-specific signals, which existing temporal GNNs and graph-level baselines cannot provide. Our framework overcomes these limits by using a shared trunk that fuses global structural dynamics with local temporal information to support robust multi-graph, multi-task prediction.}

\textbf{Transferability Across Networks.} The second challenge is transferability across networks. In real-world dynamic settings, new networks constantly emerge, data may be sparse or incomplete, and training from scratch is costly. A practical solution is to pre-train temporal graph models on multiple observed networks and transfer them to unseen ones without retraining \citep{levie2021transferability,ruiz2023transferability,wang2025towards, ranjan2026relational}. This requires learning representations that capture generalizable temporal patterns while scaling effectively with data size and model capacity. Unlike static graphs, temporal graphs add the complexity of evolving structures and features, which makes transfer and scaling more difficult.
\revTwo{Yet, no prior work examines how to build a representation trunk that captures cross-network temporal invariances while supporting multiple tasks through flexible, lightweight heads.} The tasks we study match real system needs: forecasting structural stability, connectivity growth, and shrinkage, which are standard aggregate indicators used in risk analysis and network monitoring.

\begin{wrapfigure}{r}{0.65\textwidth}  
  \centering
  \vspace{-18pt}
  \includegraphics[width=0.64\textwidth]{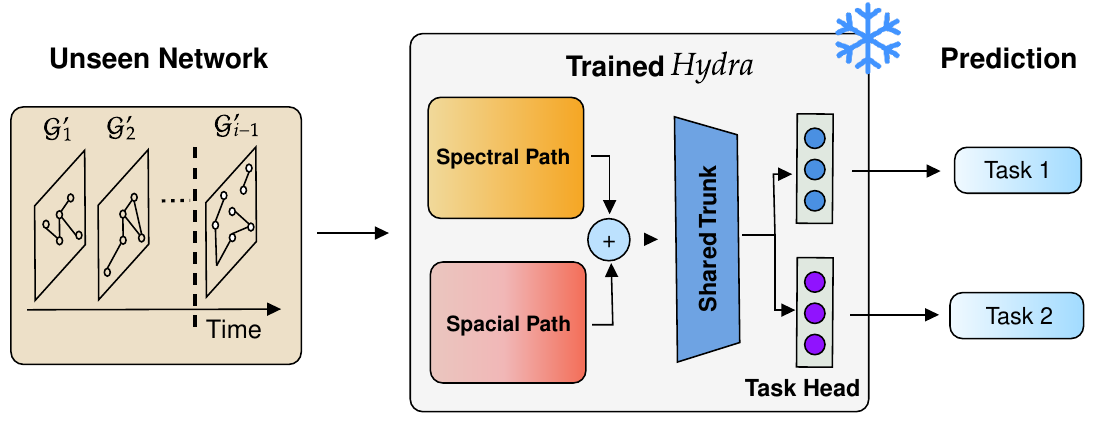}
  \caption{\textbf{\methodname overview.} \methodname is trained once on multiple networks and, without additional training, can be directly applied to unseen networks to predict multiple properties simultaneously.}
  \label{fig:hydra_overview}
   \vspace{-9pt}
\end{wrapfigure}

In this work, we address both challenges by introducing \textbf{\methodname}, a novel temporal multi-task graph neural network (GNN) designed for graph-level multi-property predictions with strong transferability across diverse networks. To capture global structural information, \methodname leverages graph Laplacian descriptors, while a Temporal Graph Neural Network (TGNN) models local node interactions and temporal dynamics with a self-attention pooling layer. 

\revTwo{This trunk–head design separates a shared spectral–spatio–temporal backbone that captures cross-network, task-invariant dynamics from lightweight task-specific heads that flexibly support heterogeneous objectives.}
The resulting spectral and spatio-temporal representations are fused and passed through task-specific prediction heads (Figure~\ref{fig:hydra_overview}). Trained end-to-end in a multi-task setting, \methodname achieves state-of-the-art performance on six benchmark tasks spanning classification and regression. 

Significantly, it generalizes across unseen networks, consistently outperforming models trained from scratch on those networks. We also analyze its scaling behavior, showing that performance \crthree{can improve} as more training networks are incorporated.

Our contributions are summarized as follows:

\begin{itemize}[leftmargin=*, noitemsep, topsep=0.2pt]
\item \textbf{First study on multi-task learning for temporal graphs.} We introduce the first model that jointly handles multiple temporal graph prediction tasks without retraining, achieving state-of-the-art performance while addressing task interference and cross-network invariance. 
\item \textbf{Novel integration of TGNNs with spectral methods.} We incorporate Laplacian descriptors into TGNNs for graph-level prediction, enabling cross-network transfer by capturing structural invariances absent from purely temporal or purely spectral models.

\item \textbf{Comprehensive study of transfer and scaling.} Hydra achieves an 8.9\% AUC gain in classification and 8.2\% MAE gain in two regression tasks over the strongest baselines while remaining competitive on the third task without any task-specific training. Moreover, transferability to unseen networks \crthree{can improve} as the number of training datasets increases.
\end{itemize}
 
\section{Background and Related Work}
\label{sec:related_work}

\textbf{Temporal Graph Property Prediction.} 
Graph property prediction has been widely studied in the static setting, where various methods \citep{luo2025can,chengxuan2024graphormer, yang2023towards} have shown strong performance in tasks such as molecular property prediction \citep{sypetkowski2024scalability,wieder2020compact} and social network analysis \citep{chen2020multi}. 
Building upon these advances, temporal graph property prediction has recently received increasing attention.  
One pioneering approach to extend static models to temporal settings is GraphPulse \citep{shamsi2024graphpulse}, which combines recurrent architectures with topological data analysis tools to capture temporal patterns. 
However, existing work has been limited to training on a single network \citep{huang2023tgbTGB,huang2024benchtemp}, except for MiNT \citep{shamsi2024mint}, which trains across multiple networks and demonstrates transferability to unseen networks during inference. However, MiNT is limited to a single classification task, and its computational cost is high because of its hyperbolic backbone. In this work, we go beyond this limitation by addressing multiple classification and regression tasks in a scalable manner.

\textbf{Multi-Task Learning.} 
Multi-task learning has been extensively studied as a means to improve generalization by jointly optimizing related tasks and leveraging shared information \citep{caruana1997multitask,zhang2018overview,zhang2021survey}. 
In graph learning, several approaches integrate learning strategies to enhance generalization, achieving strong performance across diverse datasets and downstream tasks~\citep{jin2022automated, ju2023multitask, klaser2024minimol, nassif2020multitask}. 
However, while these advances have driven substantial progress in static graph learning, applying multi-task learning to temporal graphs remains unexplored.

\textbf{Temporal Graph Transferability.}
Recent advances in temporal graph learning have produced models tailored for tasks such as link prediction \citep{rossi2020temporal,huang2023temporal,pmlr-v296-cornell25a,shi2026temporal}. While these methods achieve strong results on specific tasks and datasets, their applicability is often narrow: models trained for one task or on one network may fail to generalize to new settings. Unlike static graphs, temporal graphs introduce additional transfer challenges, including evolving topologies, distribution shifts, and non-stationary dynamics. These factors make it difficult for a model to retain predictive power outside its training environment. Addressing transferability is important for real-world use cases where networks are constantly changing, data collection is costly, and retraining for every task or dataset is impractical \citep{wang2020deep,huang2023tgbTGB,qi2023blockchain}. A central open problem is to design models that capture generalizable temporal patterns so that knowledge gained from past networks or tasks can be reused in unseen ones \citep{pan2025towards,liang2025foundation}. Developing such models would enable scalable, efficient, and continuous prediction in dynamic environments.
\crthree{Hydra uses supervised multi-task pre-training with direct zero-shot prediction, rather than self-supervised pre-training followed by target-specific fine-tuning. We view it as a step toward temporal graph foundation models within the evaluated tasks and domains.}
 
\section{\methodname: A Multi-task, Transfer Learning Model}

\subsection{Problem Definition}
 
We define a discrete-time temporal graph as a sequence of snapshots
$\G = \{ \G_t = (V_t, E_t, X_t) \}_{t=1}^T,$ where $V_t$ and $E_t$ denote the node and edge sets at time $t$, and $X_t$ represents associated node or edge features.  In this work, we consider a collection of temporal graphs
$\mathcal{D} = \{ \G^{(i)} \}_{i=1}^N,$ each evolving independently over time. For each temporal graph
$\G^{(i)} = \{ \G_t^{(i)} \}_{t=1}^{T_i}$ we associate task-specific labels $\{ y_t^{(i,k)} \}_{k=1}^K$ for $K$ prediction tasks.

\begin{definition}[Multi-Task Temporal Graph Property Prediction]\label{def:task}
The objective is to learn a model
$$
f_\theta :
\Big\{\,\{\{ \G_t^{(i)} \}_{t=1}^{T-1}\}_ {i=1}^{N} \Big\}
\;\mapsto\;
\Big\{\, \left (\hat{y}_t^{(i,1)}, \ldots, \hat{y}_t^{(i,K)}\right ) \,\Big\}_{i=1}^{N},
$$
that ingests all snapshots of all graphs in $\mathcal{D}$ up to time $t-1$
and jointly predicts $K$ graph-level properties at time $t$ for each graph. The parameters $\theta$ are optimized jointly across $\mathcal{D}$, and the trained model 
is required to generalize to temporal graphs not included in $\mathcal{D}$.
\end{definition}

\methodname is evaluated on two categories of prediction tasks across multiple temporal graphs.  In the \textbf{classification setting}, for $k \in \mathcal{K}_{\mathrm{cls}}$, the objective is to predict binary outcomes  $\hat{y}_t^{(i,k)} \in \{0,1\}$ indicating the directional change of graph-level properties between consecutive snapshots. In the \textbf{regression setting}, for $k \in \mathcal{K}_{\mathrm{reg}}$, the model predicts numerical values $\hat{y}_t^{(i,k)} \in \mathbb{R}$ that quantify the magnitude of graph-level properties at the next snapshot.  In principle, this division allows \methodname to handle both discrete and continuous objectives within a single unified architecture.

Multi-task learning provides the foundation for this setting \citep{zhang2021survey,ruder2017overview}. Instead of training a separate model for each task, it enables simultaneous prediction, allowing shared temporal representations to capture interdependencies between tasks~ \citep{crawshaw2020multi}. \methodname builds on this foundation by unifying diverse temporal prediction tasks under a single architecture.  Specifically, we focus on a \textit{shared trunk} architecture with task-specific heads, as it has proven effective in balancing information sharing 
and mitigating negative transfer \citep{crawshaw2020multi}.  Joint task learning lets the model exploit shared temporal dynamics rather than treating each task independently. Figure~\ref{fig:model_overview} shows the \methodname framework.

\subsection{\methodname: Shared Trunk with Task-Specific Heads}

The central design choice of \methodname is to couple a shared representation trunk with lightweight task-specific heads. We adopt this structure because temporal graph property prediction requires representations that are both transferable across heterogeneous graphs and adaptable to multiple tasks. The name \methodname, from the water serpent in Greek and Roman mythology with multiple heads, emphasizes this shared representation architecture with multiple heads. \revThree{The trunk captures common temporal–structural patterns once, while task-specific heads provide minimal but sufficient specialization.}

\methodname instantiates the predictor $f_\theta$ (Definition~\ref{def:task}) with the shared trunk that produces embeddings $H^{(i)}_t$ for $\G_t^{(i)}$, followed by task-specific heads $\{g_{\psi_k}\}_{k=1}^K$ that map $H_t^{(i)}$ to predictions $\hat y_t^{(i,k)}$.

\revThree{The trunk jointly leverages (i) spectral descriptors that summarize global connectivity patterns and (ii) spatial GNN-based encoders that model local temporal interactions. These components provide complementary global and local information while avoiding redundant representation learning.}

\begin{figure*}[t!]
    \centering
    \includegraphics[width=0.8\linewidth]{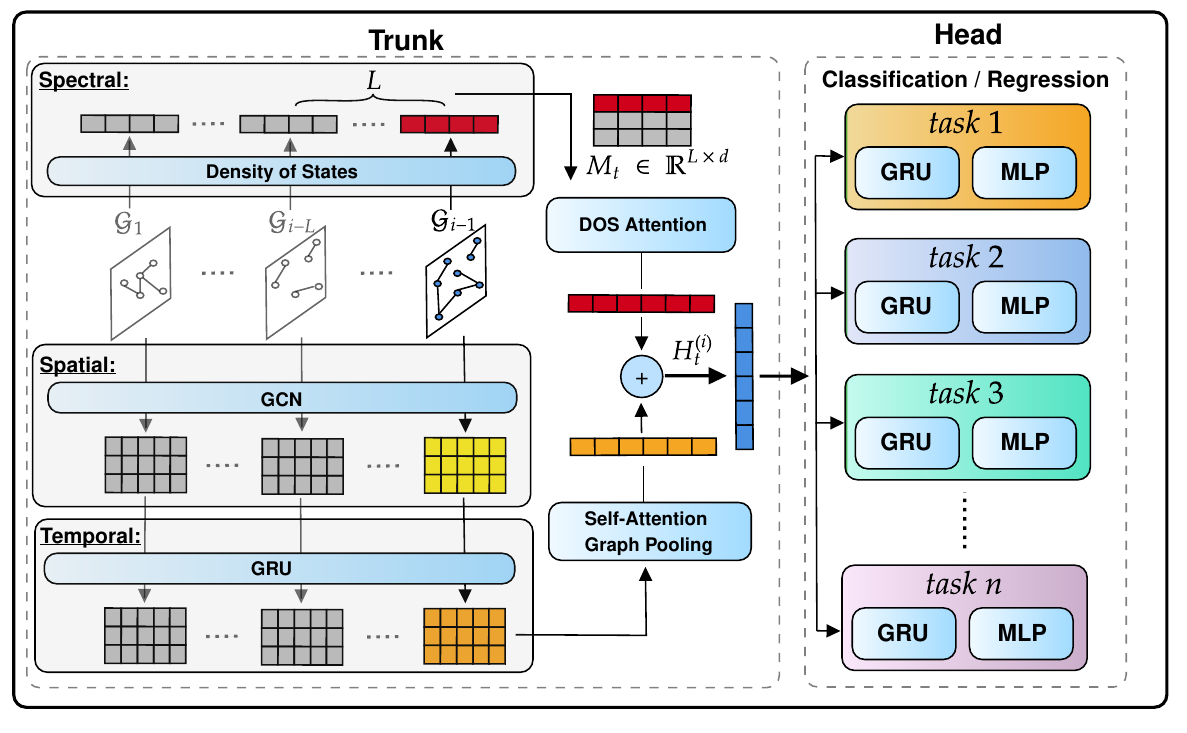}
    \caption{\textbf{\methodname framework.} The architecture integrates spatial and spectral paths with a unified multi-task design, enabling \methodname to learn transferable representations across diverse temporal graph tasks.}
    \label{fig:model_overview}
    \vspace{-12pt}
\end{figure*}

As shown in Figure~\ref{fig:model_overview}, the trunk $H_t^{(i)}$ consists of a spectral path, which encodes global descriptors, and a spatial path, which models local temporal dynamics. \revThree{Their outputs are concatenated into the shared embedding $H_t^{(i)}$,} which is then passed to task-specific decoder heads, $\{g_{\psi_k}\}_{k=1}^K$, each producing a prediction $\hat y_t^{(i,k)}$.

\subsubsection{Spectral Path for Global Connectivity}

Laplacian eigenvalues effectively encode global graph structural properties~\citep{mohar1991laplacian}. For example, in star, path, and fully-connected graphs, the distribution of the Laplacian eigenvalues is distinct and captures their unique topologies. However, computing all eigenvalues of the graph Laplacian matrix is highly expensive and only feasible for small graphs with thousands of nodes. Thus, it is important to efficiently approximate the spectrum of the Laplacian. 

\textbf{Network Density of States.} To capture information from the Laplacian spectrum, we use the network density of states~(DOS), the distribution of the eigenvalues of the graph Laplacian matrix~\citep{dong2019network}, as follows:
 
\begin{definition}[Density of States] 
Given a temporal graph snapshot $\G_t$ with adjacency matrix $\mat{A}$ and degree matrix $\mat{D}$, let the (unnormalized) Laplacian be $\mat{L}=\mat{D}-\mat{A}$. We use the normalized Laplacian $\mat{L}_{\mathrm{sym}}=\mat{I}-\mat{D}^{-\tfrac{1}{2}}\mat{A}\mat{D}^{-\tfrac{1}{2}}$
\crone{with eigenvalues $\lambda_1,\ldots,\lambda_{|V_t|}$.} 
The \emph{density of states} is then defined as a discrete distribution supported on the spectrum of $\mat{L}_{\mathrm{sym}}$: $
\mu(\lambda)=\tfrac{1}{|V_t|}\sum_{i=1}^{|V_t|}\delta(\lambda-\lambda_i)
$
\end{definition}

\crone{Intuitively, DOS summarizes the distribution of Laplacian eigenvalues and thus global connectivity. The Kernel Polynomial Method~\citep{weisse2006kernel} approximates this distribution without explicitly computing eigenvalues or eigenvectors. We discretize the reconstructed density into 20 equal-width bins over $[0,2]$, yielding a fixed-dimensional descriptor irrespective of graph size; implementation details are given in Appendix~\ref{sec:complexity}.}

\textbf{Spectral Memory.}
To stabilize spectral signals and capture short-term trends, we maintain a fixed-length FIFO memory of recent DOS descriptors. 
At snapshot $t$, let $s_t \in \mathbb{R}^d$ denote the spectral feature (e.g., a $d{=}20$ DOS vector). 
For prediction at time $t$, the memory contains the last $L$ descriptors up to $t-1$: $M_t = [s_{t-L};\dots;s_{t-1}] \in \mathbb{R}^{L\times d}.$ After prediction at $t$, the new descriptor $s_t$ is computed and rolled into the memory:
$M_{t+1} = \mathrm{roll}(M_t), 
\quad M_{t+1}[-1,:] = s_t.
$
\crone{The FIFO memory provides a recent structural trajectory rather than a potentially noisy single snapshot, while discarding outdated descriptors.}

\textbf{Attention over Spectral Memory.}
Rather than uniform averaging, we apply attention to emphasize historically similar or predictive patterns. 
With learnable projections $W_q,W_k,W_v \in \mathbb{R}^{d\times d_a}$, we compute $q_t = s_{t-1}W_q,\quad K_t = M_tW_k,\quad V_t = M_tW_v.
$ Attention weights and the attentive spectral embedding are
$\alpha_t = \mathrm{softmax}\!\left(\tfrac{q_t K_t^\top}{\sqrt{d_a}}\right), 
\quad 
z_t = \alpha_t V_t.$ \crone{Unlike uniform averaging, attention can emphasize relevant history and distinguish transient fluctuations from persistent structural changes.} \crthree{The DOS descriptors of each snapshot are fixed, while their temporal aggregation is learned through $W_q,W_k,W_v$.}

\subsubsection{Spatial Path for Local Connectivity}

We integrate spatial and temporal dynamics by combining a Graph Convolutional Network (GCN)~\citep{GCN_kipf2016} with Gated Recurrent Units (GRUs)~\citep{GRU_cho2014}. Similar approaches have proven effective in temporal settings~\citep{TGCN_zhao2019}. At snapshot $t$, a GCN encodes structural features $\mat{Z}_t = \mathrm{GCN}(\mat{X}_{t-1},E_{t-1}),$ and a GRU updates node states with temporal memory 
$\mat{H}_t = \mathrm{GRU}(\mat{Z}_t,\mat{H}_{t-1}).$ The representation $\mat{H}_t$ is thus spatially grounded and temporally enriched.

\textbf{Pooling and Graph Readout.}
We apply Self-Attention Graph Pooling (SAGPool)~\citep{lee2019self} to emphasize structurally important nodes. Node scores are computed as $s = \mathrm{GCN}(\mat{H}_t,E_{t-1}) W_s$, where $W_s \in \mathbb{R}^{d\times 1}$ projects node embeddings to scalar scores.  The top-$k$ nodes are retained to form the pooled graph $(V^{'}, E^{'})$ with embeddings $\mat{H}'_t$. We then compute a graph-level vector using mean pooling over the selected nodes: $g_t = \tfrac{1}{|V^{'}|}\sum_{v \in V^{'}} h_v, \quad h_v \in \mat{H}'_t.$ SAGPool reduces noise from low-activity nodes and highlights structurally central actors, which is particularly relevant in sparse networks where activity is concentrated in a few key nodes~\citep{shamsi2024graphpulse}.

\subsection{Prediction Head}
We fuse spatial and spectral signals to produce a joint temporal state. 
At each step, the shared hidden state of the prediction head
$u_t = \mathrm{GRU}([g_t \| z_t], u_{t-1}),$ where $[g_t \| z_t]$ denotes concatenation of spatial and spectral features. 
The recurrent update captures cross-modal dependencies over time.  Task-specific MLPs then map the shared state to predictions: 
$
\hat{y}_t^{(k)} = \mathrm{MLP}_k(u_t), \quad k=1,\dots,K.
$
This balances efficiency through a shared backbone with flexibility through task-specific heads.

\subsection{End-to-End Training}

\methodname supports both classification and regression objectives. For classification, each head is optimized with Binary Cross-Entropy loss; for regression, with Mean Squared Error. The total loss in each setting is the sum over task-specific losses within that group. Following prior work on loss balancing~\citep{yu2020gradient,sener2018multi}, we train the two groups separately, as naïvely combining heterogeneous objectives degraded optimization stability. This preserves Hydra's multi-task benefits within each group while maintaining transferability across unseen networks.

\section{Experimental Evaluation}

\subsection{Training Setup and Algorithm} \label{sec:exp_setup}

\noindent\textbf{Datasets.} We evaluate \methodname on the MiNT benchmark of 84 ERC-20 temporal transaction networks~\citep{shamsi2024mint}, randomly selecting 64 networks for training/validation and 20 held-out networks for zero-shot transfer evaluation.  Within each network, snapshots are split 70\%–15\%–15\% into training, validation, and test sets. Appendix~\ref{sec:datasetStats} provides detailed dataset statistics. 
This setting lets us evaluate whether \methodname generalizes across network topologies and temporal patterns.  During each epoch, we present networks in random order, and reset historical embeddings at the start of each network to avoid cross-network leakage. The shared trunk is optimized jointly, while all task-specific heads are updated under a multi-task loss with early stopping on validation performance. We train models for $100$ to $250$ epochs with an initial learning rate of $1.5\times 10^{-3}$. We apply early stopping based on validation AUC for classification and MAE for regression, with a patience of $20$ epochs and a tolerance of $5\times 10^{-2}$. We use binary cross-entropy for classification tasks and mean squared error for regression tasks. We optimize using Adam~\citep{kingma2015Adam}. We evaluate each task on 20 unseen test networks across three independent runs, yielding 60 result points per task. We report the average results from three different runs for each task separately, using ROC-AUC for classification objectives and Mean Absolute Error for regression objectives. We pre-train \methodname with 8, 16, and 32 networks to study how scale affects transferability. \crthree{The primary scaling study uses up to 32 networks; the 64-network and capacity analyses are in Appendix~\ref{sec:cr_scaling}.} Unless otherwise specified, all reported results refer to \methodname-32, as this configuration \crthree{offers strong overall performance}.

\noindent\textbf{Tasks.} We train \methodname in two multi-task settings: a classification model with three classification heads and a regression model with three regression heads. The \cls{classification} tasks jointly predict whether the Largest Connected Component (LCC), the number of nodes, or the number of edges will grow or shrink. 
The \rgs{regression} tasks estimate the number of influential nodes (i.e., nodes with degree greater than five), new nodes, and edges. Appendix~\ref{appendix:task_definition} provides full task definitions.

\noindent\textbf{Models.} 
 
We compare \methodname against HTGN~\citep{menglin2021HTGN}, GC-LSTM~\citep{chen2022GCLSTM}, EvolveGCN~\citep{Aldo2020EvolveGCN}, GraphPulse~\citep{shamsi2024graphpulse}, ROLAND~\citep{you2022roland}, TGCN~\citep{ling2020tgcn}, and WinGNN~\citep{yifan2023wingnn}, and additionally MiNT~\citep{shamsi2024mint} for classification (MiNT is the only prior framework explicitly designed for transferability across temporal graphs for classification). \crtwo{Hydra and MiNT are trained only on source networks and evaluated on the same 20 held-out networks without target-network training or fine-tuning. Single-network baselines are trained separately on the training split of each target network.} 
Appendix~\ref{appendix:baselines} provides baseline details.

\textbf{Compute Resources.} Experiments were run on a dedicated cluster compute node equipped with an NVIDIA H100 GPU (80 GB), dual 64-core AMD EPYC CPUs, and 512 GB RAM.

\subsection{Classification and Regression Results}

Table~\ref{tab:classification_summary} presents the aggregate \cls{classification} results across three tasks: LCC Growth/Shrinkage, Node Growth/Shrinkage, and Edge Growth/Shrinkage. For each task, we report three metrics: (\emph{1st}) the number of datasets where the method achieved first place, (\emph{Rank}) the average ranking across datasets, and (\emph{AUC}) the average area under the ROC curve (metric formulations are provided in appendix~\ref{appendix:metrics}).  \methodname consistently achieves the strongest performance across all metrics, outperforming both state-of-the-art specifically trained models and the transferable baseline. \crthree{Per-network means and standard deviations over three runs are reported in Appendix~\ref{appendix:addl_results}.}

\begin{table}[t]
\centering
\caption{\cls{Classification} task results. Across LCC-GS, Node-GS, and Edge-G/S classification, \methodname consistently outperforms both specifically trained and transferable baselines. 
}
\label{tab:classification_summary}

\rowcolors{2}{gray!5}{white}
\resizebox{0.76\linewidth}{!}{%
\begin{tabular}{lccc ccc ccc}
\toprule
  & \multicolumn{3}{c}{\textbf{LCC-G/S}} 
 & \multicolumn{3}{c}{\textbf{Node-G/S}} 
 & \multicolumn{3}{c}{\textbf{Edge-G/S}} \\
\cmidrule(lr){2-4} \cmidrule(lr){5-7} \cmidrule(lr){8-10}
\textbf{} 
& 1st$\uparrow$ & Rank$\downarrow$ & AUC$\uparrow$
& 1st$\uparrow$ & Rank$\downarrow$ & AUC$\uparrow$
& 1st$\uparrow$ & Rank$\downarrow$ & AUC$\uparrow$ \\
\midrule
\multicolumn{10}{l}{\textbf{Single Models}} \\
\hspace{1em}HTGN       & 2 & 4.40 & 0.648 & 0 & 3.65 & 0.615 & 1 & 4.85 & 0.684 \\
\hspace{1em}GC-LSTM    & 0 & 6.00 & 0.572 & 0 & 6.65 & 0.489 & 1 & 5.80 & 0.609 \\
\hspace{1em}EvolveGCN  & 0 & 6.30 & 0.585 & 0 & 7.00 & 0.487 & 0 & 6.10 & 0.626 \\
\hspace{1em}GraphPulse & 3 & 3.50 & 0.686 & 2 & 2.95 & 0.652 & 6 & 3.80 & 0.712 \\
\hspace{1em}ROLAND     & 0 & 6.95 & 0.556 & 0 & 6.70 & 0.480 & 0 & 6.30 & 0.542 \\
\hspace{1em}TGCN       & 1 & 6.10 & 0.583 & 0 & 6.15 & 0.530 & 0 & 5.60 & 0.633 \\
\hspace{1em}WinGNN     & 0 & 6.10 & 0.585 & 0 & 6.20 & 0.528 & 0 & 5.55 & 0.642 \\
\midrule
\multicolumn{10}{l}{\textbf{Transfer Models}} \\
\hspace{1em}MiNT       & 1 & 3.95 & 0.672 & 3 & 3.55 & 0.616 & 4 & 3.30 & 0.727 \\
\rowcolor{gray!15}
\hspace{1em}\methodname (ours) & \textbf{13} & \textbf{1.70} & \textbf{0.759} 
           & \textbf{15} & \textbf{1.95} & \textbf{0.734} 
           & \textbf{8}  & \textbf{2.80} & \textbf{0.753} \\
\bottomrule
\end{tabular}
}

\begin{tablenotes}[flushleft]
\item \metriclegend
\end{tablenotes}
\vspace{-18pt}
\end{table}

\begin{table}[t]
\centering
\caption{\rgs{Regression} task results. 
Across New Node Count, Influential Node Count, and Edge Count regression, \methodname achieves the best performance on two tasks and is competitive on the third, all without additional training.}
\label{tab:regression_summary}

\rowcolors{2}{gray!5}{white}
\resizebox{0.76\linewidth}{!}{%
\begin{tabular}{lccc ccc ccc}
\toprule
 & \multicolumn{3}{c}{\textbf{New Node Count}} 
 & \multicolumn{3}{c}{\textbf{Influential Node Count}} 
 & \multicolumn{3}{c}{\textbf{Edge Count}} \\
\cmidrule(lr){2-4} \cmidrule(lr){5-7} \cmidrule(lr){8-10}
\textbf{} 
& 1st$\uparrow$ & Rank$\downarrow$ & MAE$\downarrow$
& 1st$\uparrow$ & Rank$\downarrow$ & MAE$\downarrow$
& 1st$\uparrow$ & Rank$\downarrow$ & MAE$\downarrow$ \\
\midrule
\multicolumn{10}{l}{\textbf{Single Models}} \\
\hspace{1em}HTGN       & 4 & 3.45 & 0.039 & 4 & \textbf{3.52} & \textbf{0.050} & 2 & 3.95 & 0.049 \\
\hspace{1em}TGCN       & 1 & 4.10 & 0.042 & 1 & 4.85 & 0.059 & 1 & 4.60 & 0.054 \\
\hspace{1em}GCLSTM     & 2 & 4.00 & 0.043 & \textbf{5} & 3.80 & 0.053 & 4 & 4.58 & 0.057 \\
\hspace{1em}ROLAND     & 3 & 4.35 & 0.052 & 2 & 4.55 & 0.057 & 3 & 4.58 & 0.050 \\
\hspace{1em}EGCN       & 0 & 3.90 & 0.040 & 0 & 4.20 & 0.052 & 1 & 5.30 & 0.054 \\
\hspace{1em}GraphPulse & 1 & 5.75 & 0.071 & 1 & 5.90 & 0.078 & 2 & 5.72 & 0.073 \\
\hspace{1em}WinGNN     & 4 & 4.05 & 0.054 & 4 & 4.28 & 0.059 & 5 & 3.93 & 0.057 \\
\midrule
\multicolumn{10}{l}{\textbf{Transfer Model}} \\
\rowcolor{gray!15}
\hspace{1em}\methodname (ours) & \textbf{8} & \textbf{2.50} & \textbf{0.035} 
           & \textbf{5} & 4.90 & 0.056 
           & \textbf{6} & \textbf{3.35} & \textbf{0.046} \\
\bottomrule
\end{tabular}
}

\begin{tablenotes}[flushleft]
\item \metriclegend
\end{tablenotes}
\vspace{-18pt}
\end{table}

Table~\ref{tab:regression_summary} summarizes the \rgs{regression} results across New Node Count, Influential Node Count, and Edge Count Regression. For each task, we report three metrics: (\emph{1st}) the number of datasets where the method achieved first place, (\emph{Rank}) the average ranking across datasets, and (\emph{MAE}) the average mean absolute error. \methodname is the best model on two of the three tasks and remains very close to the best baseline on Influential Node Count, demonstrating strong overall transferability for regression objectives. \methodname produces the most top-ranked results in both classification and regression settings, outperforming all baselines in five out of six tasks. \crtwo{The reported gains average task-wise relative improvements over all three classification tasks (8.9\%) and the two winning regression tasks (8.2\%); Appendix~\ref{appendix:metrics} gives the calculation.} Even in the one regression task where \methodname isn't the leader, it stays competitive with the strongest baseline. Notably, \methodname is always tested on unseen networks, while single models specifically trained for a task are evaluated on the test portion of the same networks they were trained on. Despite this tougher setting, \methodname consistently delivers better or comparable results, making it a reliable and adaptable framework for temporal graph learning.

\begin{wraptable}{r}{0.5\linewidth}
\vspace{-10pt}
\centering
\caption{Meta ablation summary for Edge G/S \cls{classification}. Average AUC and \#1st-Place counts are reported with and without the corresponding component. Per-network results are given in Appendix Tables~\ref{tab:dos_ablation} and \ref{tab:pooling_ablation}.}
\label{tab:meta_ablation}
\scriptsize
\begin{tabular}{l cc cc}
\toprule
\multirow{2}{*}{\textbf{Ablation}} & \multicolumn{2}{c}{\textbf{Avg. AUC}} & \multicolumn{2}{c}{\textbf{\#1st-Place}} \\
 & w/o & w/ & w/o & w/ \\
\midrule
Density of States       & 0.697 & \textbf{0.719} & 6  & \textbf{14} \\
Attention-based Pooling & 0.719 & \textbf{0.753} & 3  & \textbf{17} \\
\bottomrule
\end{tabular}
\vspace{-8pt}
\end{wraptable}

\noindent\textbf{Ablation Studies.} We assess the impact of \methodname's key components by disabling the DOS module and the pooling layer. \crtwo{We provide a comprehensive analysis in Appendix~\ref{appendix:ablation_stury}, including an additional edge-count regression DOS ablation.} As summarized in Table~\ref{tab:meta_ablation}, DOS improves performance from an average AUC of $0.697$ to $0.719$, raising the number of best-performing datasets from 6 to 14 and lowering the mean rank from $1.70$ to $1.30$. Pooling yields an even larger effect, increasing average AUC from $0.719$ to $0.753$ and improving the mean rank from $1.85$ to $1.15$, with gains on 17 out of 20 datasets. These results confirm that DOS provides complementary global spectral information, while pooling enhances the spatial path by emphasizing structurally important nodes, together strengthening \methodname's generalization ability.

\subsection{\methodname Generalization and Efficiency}
\label{sec:scaling_trend}

\begin{figure}[tbh]
\centering
\begin{subfigure}{.33\textwidth}
  \centering
  \includegraphics[width=\linewidth]{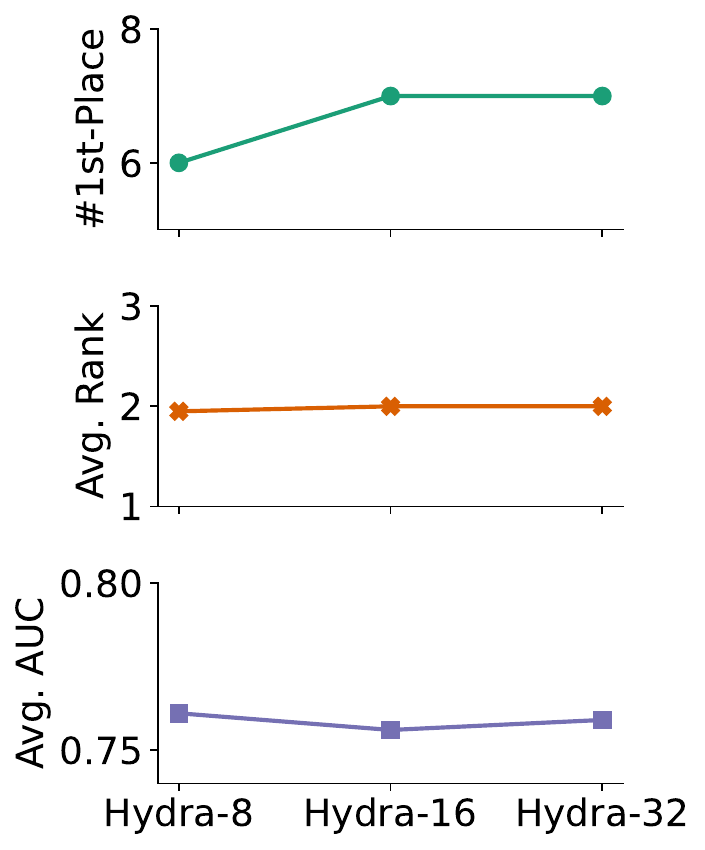}
  \caption{\cls{LLC Growth/Shrinkage}}
  \label{fig_subfig:CLS_llc_GS_AUC}
\end{subfigure}%
\begin{subfigure}{.33\textwidth}
  \centering
  \includegraphics[width=\linewidth]{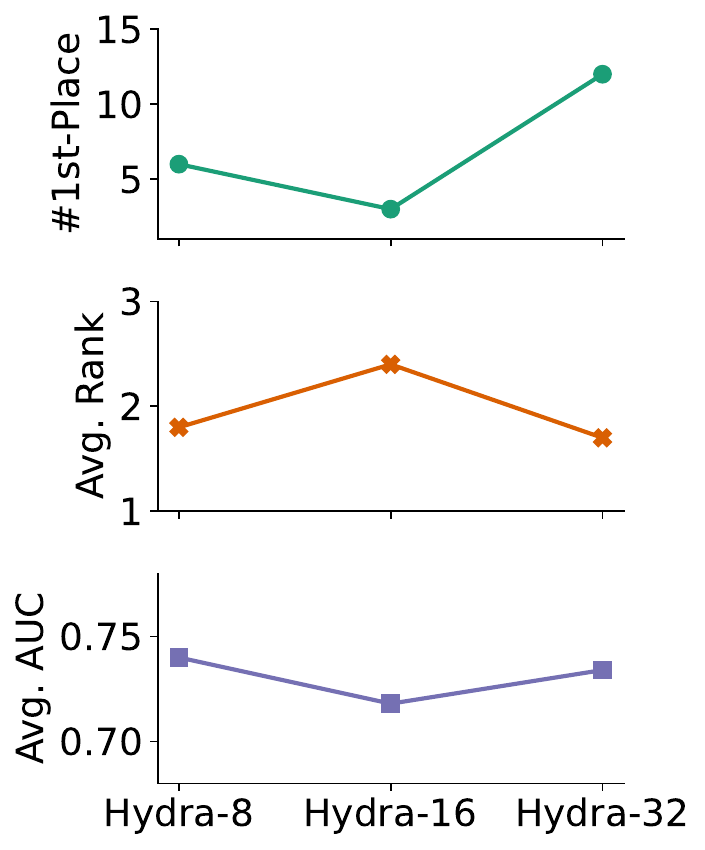}
  \caption{\cls{Node Growth/Shrinkage}} 
  \label{fig_subfig:CLS_node_GS_AUC}
\end{subfigure}
\begin{subfigure}{.33\textwidth}
  \centering
  \includegraphics[width=\linewidth]{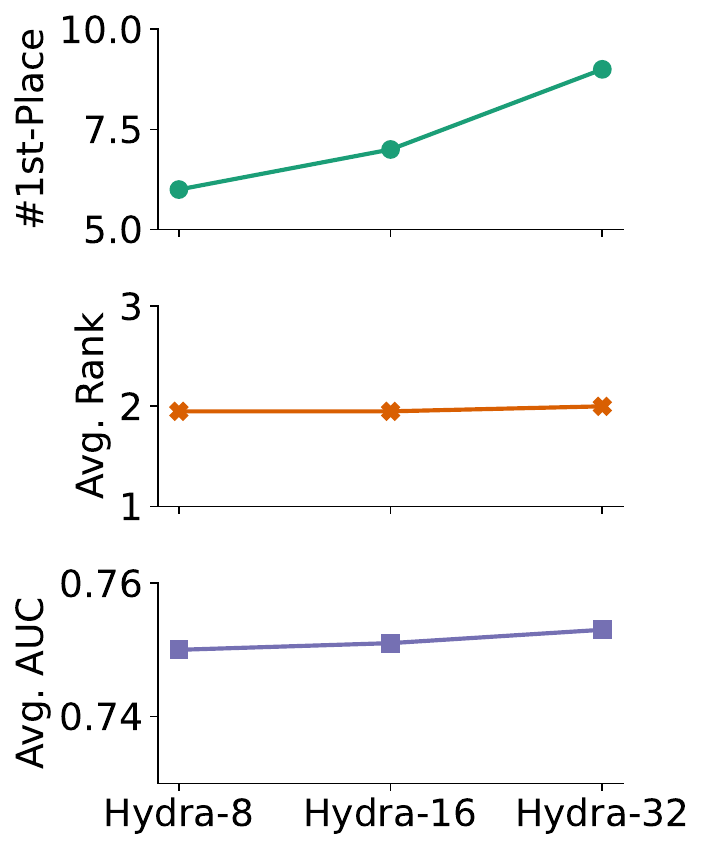}
  \caption{\cls{Edge Growth/Shrinkage}} 
  \label{fig_subfig:CLS_edge_GS_AUC}
\end{subfigure}

\begin{subfigure}{.33\textwidth}
  \centering
  \includegraphics[width=\linewidth]{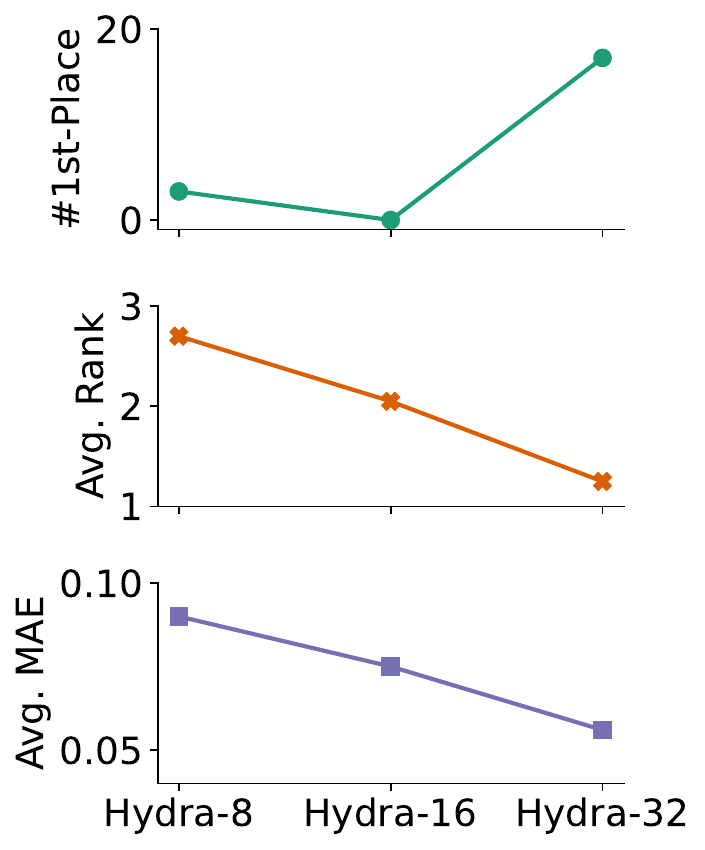}
  \caption{\rgs{\# Influential Node}}
  \label{fig_subfig:RGS_inf_node_count}
\end{subfigure}%
\begin{subfigure}{.33\textwidth}
  \centering
  \includegraphics[width=\linewidth]{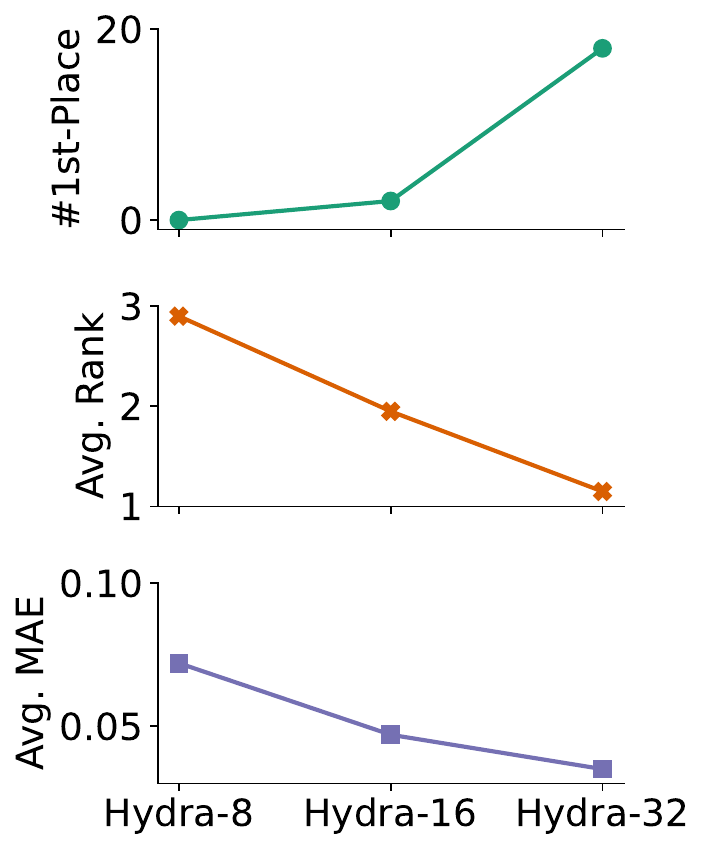}
  \caption{\rgs{\# New Nodes}} 
  \label{fig_subfig:RGS_new_node_count}
\end{subfigure}
\begin{subfigure}{.33\textwidth}
  \centering
  \includegraphics[width=\linewidth]{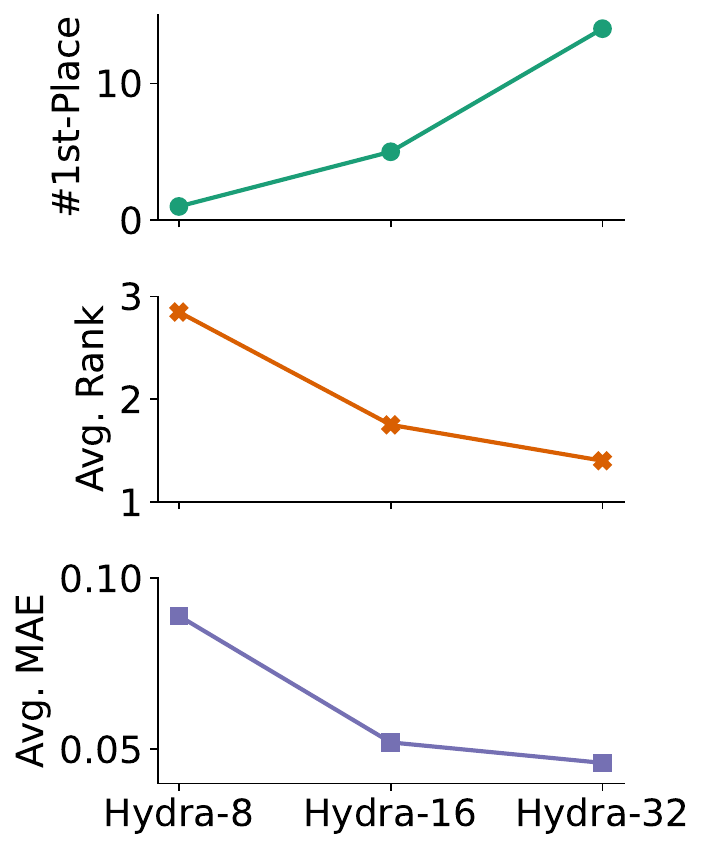}
  \caption{\rgs{\# Edges}} 
  \label{fig_subfig:RGS_edge_count}
\end{subfigure}
\caption{\crone{Scaling results for \methodname-8, \methodname-16, and \methodname-32, pre-trained on 8, 16, and 32 networks, respectively, and evaluated zero-shot on the same 20 held-out networks, across \cls{classification} and \rgs{regression} tasks.}
\crthree{\methodname-32 achieves competitive classification results and the lowest average regression errors among these settings.}
Evaluation measures are \textit{$\#1\textsuperscript{st}$-Place ($\uparrow$)}, \textit{Avg. Rank ($\downarrow$)}, \textit{Avg. AUC ($\uparrow$)}, and \textit{Avg. MAE ($\downarrow$)}.
}
\vspace{-18pt}
\label{fig:scaling_behavior}
\end{figure}

In this section, we analyze how the number of networks used during \methodname pre-training affects its performance in predicting the characteristics of unseen networks at inference time.
Fig.~\ref{fig:scaling_behavior} presents the results. 
For classification tasks, we observe that increasing the number of training networks generally leads to a higher \textit{$\#1\textsuperscript{st}$-Place}. 
This metric is a strong indicator of performance, as it reflects the number of unseen networks on which the model achieves the best results at inference time. 
In contrast, the correlation between the number of pre-training networks and metrics such as Avg. Rank or Avg. AUC is weaker, with only negligible differences across settings. Notably, \methodname-32 consistently achieves the best or on-par performance across all classification benchmarks. 
For regression tasks, we observe a clearer scaling trend: performance steadily improves from \methodname-8 to \methodname-32 across all indicators. 
Specifically, \textit{$\#1\textsuperscript{st}$-Place} increases with scale, while both Avg. Rank and Avg. MAE decrease, indicating better predictive performance. Given its \crthree{overall performance}, we adopt \textbf{\methodname-32} as our primary model and focus our subsequent evaluation on it.
Table~\ref{tab:scaling_all} in Appendix~\ref{appendix:scaling_behavior} also reports the performance of \crthree{the 8-, 16-, and 32-network variants.}

\begin{figure}[t]
    \centering
    \begin{subfigure}[t]{0.4\linewidth}
        \centering
        \includegraphics[width=\linewidth]{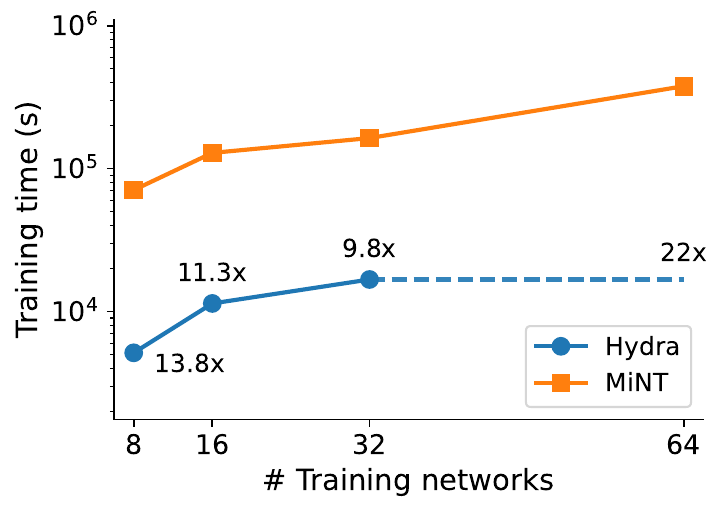}
       \caption{\textbf{Total training cost for three \cls{classification} tasks.} \methodname trains once for all tasks jointly, 
        while MiNT must be retrained separately for each task. \revThree{The dashed line indicates that MiNT is evaluated at 64 networks, while \methodname's best result occurs at 32 networks.}}
        \label{fig:time_transfer}
    \end{subfigure}
    \hfill
    \begin{subfigure}[t]{0.58\linewidth}
        \centering
        \includegraphics[width=\linewidth]{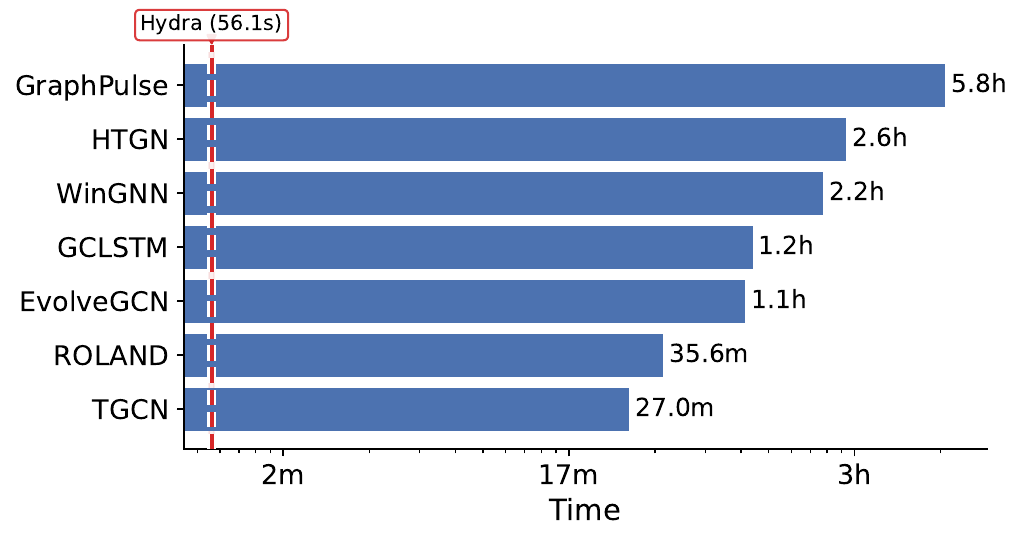}
        \caption{\textbf{Inference cost on TRAC dataset.} \methodname performs a single zero-shot \cls{classification} pass for all three tasks jointly, whereas single-network baselines require full training before they can be used for inference, even on a single task (LCC–G/S here); reported times reflect the cost of inference on an unseen network using \methodname versus single models.}
        \label{fig:time_deploy}
    \end{subfigure}
    \caption{\textbf{Training and inference efficiency of \methodname.} 
    (a) \methodname reduces training cost by avoiding retraining. 
    (b) \methodname drastically lowers inference time by requiring only one forward pass.}
    \label{fig:time_efficiency}
    \vspace{-18pt}
\end{figure}

\noindent\textbf{Computation and Time Efficiency.} The per-snapshot complexity of \methodname reduces to a scalable $\mathcal{O}(N \cdot m + n \log n)$, where $m$ is the number of edges, $n$ is the number of nodes, and $N$ is the number of Chebyshev moments used in the DOS module (see Appendix~\ref{sec:complexity} for details). 
Because MiNT must be retrained separately for each task, training three tasks requires about three full runs on 64 networks. In contrast, \methodname trains all tasks jointly in a single run on 32 networks and achieves a $22\times$ speedup, completing all three tasks in about $16{,}700$ seconds compared to more than $370{,}000$ seconds for MiNT (Figure~\ref{fig:time_transfer}).
This efficiency makes \methodname far more practical and scalable for multi-task training and inference.

\methodname inference time also improves significantly. \methodname requires only a single zero-shot inference pass to predict all three tasks on an unseen network, completing end-to-end evaluation on the largest test network (TRAC) in about $56$ seconds (see Figure~\ref{fig:time_deploy}). 
 \begin{wraptable}{r}{0.7\linewidth} 
\centering
\scriptsize
\renewcommand{\arraystretch}{0.9}
\newcommand{\std}[1]{{\scriptsize\,(#1)}}

\caption{\textcolor{black}{MAE of edge count prediction under one, two, and three task \methodname configurations, reported with standard deviations.}}
\label{tab:multitask_edgecount}

\resizebox{\linewidth}{!}{
\begin{tabular}{lccc}
\toprule
\textbf{Dataset} & \textbf{\methodname (One Task)} & \textbf{\methodname (Two Tasks)} & \textbf{\methodname (Three Tasks)} \\
\midrule
MIR          & 0.057\std{$\pm$ 0.004} & 0.034\std{$\pm$ 0.002} & \textbf{0.016}\std{$\pm$ 0.005} \\
DOGE2.0      & 0.169\std{$\pm$ 0.016} & \textbf{0.150}\std{$\pm$ 0.015} & 0.187\std{$\pm$ 0.008} \\
MUTE         & 0.053\std{$\pm$ 0.020} & 0.067\std{$\pm$ 0.011} & \textbf{0.021}\std{$\pm$ 0.022} \\
EVERMOON     & 0.029\std{$\pm$ 0.021} & 0.042\std{$\pm$ 0.010} & \textbf{0.017}\std{$\pm$ 0.000} \\
DERC         & 0.024\std{$\pm$ 0.016} & 0.035\std{$\pm$ 0.010} & \textbf{0.021}\std{$\pm$ 0.003} \\
ADX          & \textbf{0.023}\std{$\pm$ 0.010} & 0.027\std{$\pm$ 0.006} & 0.025\std{$\pm$ 0.008} \\
HOICHI       & 0.071\std{$\pm$ 0.006} & 0.068\std{$\pm$ 0.011} & \textbf{0.027}\std{$\pm$ 0.020} \\
SDEX         & 0.060\std{$\pm$ 0.023} & \textbf{0.046}\std{$\pm$ 0.007} & 0.095\std{$\pm$ 0.043} \\
BAG          & 0.061\std{$\pm$ 0.018} & 0.067\std{$\pm$ 0.020} & \textbf{0.041}\std{$\pm$ 0.020} \\
XCN          & 0.037\std{$\pm$ 0.017} & \textbf{0.026}\std{$\pm$ 0.010} & 0.062\std{$\pm$ 0.015} \\
ETH2x-FLI    & 0.030\std{$\pm$ 0.013} & 0.038\std{$\pm$ 0.007} & \textbf{0.023}\std{$\pm$ 0.001} \\
stkAAVE      & 0.026\std{$\pm$ 0.008} & \textbf{0.021}\std{$\pm$ 0.003} & 0.051\std{$\pm$ 0.022} \\
GLM          & 0.092\std{$\pm$ 0.009} & \textbf{0.083}\std{$\pm$ 0.017} & 0.087\std{$\pm$ 0.008} \\
QOM          & 0.025\std{$\pm$ 0.003} & \textbf{0.023}\std{$\pm$ 0.005} & 0.036\std{$\pm$ 0.013} \\
WOJAK        & 0.025\std{$\pm$ 0.023} & 0.039\std{$\pm$ 0.017} & \textbf{0.010}\std{$\pm$ 0.004} \\
DINO         & 0.031\std{$\pm$ 0.018} & 0.041\std{$\pm$ 0.010} & \textbf{0.017}\std{$\pm$ 0.007} \\
Metis        & 0.047\std{$\pm$ 0.014} & 0.054\std{$\pm$ 0.016} & \textbf{0.034}\std{$\pm$ 0.010} \\
REPv2        & 0.103\std{$\pm$ 0.002} & \textbf{0.102}\std{$\pm$ 0.003} & 0.111\std{$\pm$ 0.008} \\
TRAC         & 0.029\std{$\pm$ 0.016} & 0.039\std{$\pm$ 0.012} & \textbf{0.021}\std{$\pm$ 0.001} \\
BEPRO        & 0.031\std{$\pm$ 0.025} & 0.046\std{$\pm$ 0.012} & \textbf{0.011}\std{$\pm$ 0.008} \\
\midrule
\textbf{Average}     & 0.051 & 0.052 & \textbf{0.046} \\
\textbf{Best Count}  & 1     & 7     & \textbf{12} \\
\bottomrule
\end{tabular}
}
\vspace{-1pt}
\end{wraptable}

Single-network baselines, in contrast, require retraining from scratch for every dataset and task before they can be used for inference, with training times ranging from several thousand seconds to over $9{,}000$ seconds even for a single task. Extending these models to multiple tasks or multiple networks multiplies the cost linearly, making them infeasible at scale. 
\methodname's design eliminates the need for repeated training and heavy preprocessing, providing unique efficiency in both training and inference.

\crthree{\noindent\textbf{Transfer beyond ERC-20.} We additionally train Hydra-Social on six social networks and Hydra-Mix on those networks plus 32 ERC-20 networks, holding out MathOverflow and RedditB. On RedditB, Edge G/S AUC rises from 0.604 (Social) to 0.769 (Mix); on MathOverflow, Hydra-Mix reaches 0.630, below MiNT-Mix's 0.782. These results support mixed-domain transfer but show that gains are not uniform (Appendix~\ref{sec:cr_social}).}

\section{Impact of Multi-Task Training}
\label{sec:multitask_edgecount}

To assess how multi-task learning shapes the shared temporal representation, we examine edge count prediction as a reference task and train three model variants: single-task (edge count only), two-task (edge count and new node count), and the full three-task \methodname configuration. Table~\ref{tab:multitask_edgecount} shows the results.

The trend is non-monotonic: average MAE is 0.051, 0.052, and 0.046 for one, two, and three tasks, respectively. The three-task model achieves the best results on 12 out of 20 datasets and the lowest overall average MAE, compared to 7 and 1 for the two-task and single-task variants respectively. \crtwo{These gains are consistent with additional tasks providing complementary supervisory signals} that encourage the shared trunk to learn richer temporal and structural patterns, \crtwo{suggesting benefits from multi-task supervision overall.}

\section{Conclusion}
\label{sec:conclusion}

We have introduced \methodname, a novel temporal graph-level multi-task and multi-network model designed to address the complexities of dynamic network analysis. \methodname combines a spatial path, which captures local connectivity through temporal GNNs, along with a spectral path that attends to global structural patterns, forming a new architectural design for temporal graph learning. This unified framework enables \methodname to handle diverse prediction tasks simultaneously while transferring effectively to unseen networks. Empirically, \methodname consistently outperforms state-of-the-art baselines across benchmarks, achieving strong performance without requiring additional training on target networks. These results highlight \methodname as the first architecture to bring together spectral and spatial pathways for transferable, multi-task learning on temporal graphs. Looking ahead, this work opens up a promising research direction toward more generalizable temporal models that can support a wide range of tasks.

\subsection*{Acknowledgments}
This work was partially supported by the Canadian NSERC Discovery Grant RGPIN-2020-05665: Data Science on Blockchains, the National Science Foundation under grant DMS-2229417, and the Simons Foundation under grant SFM-21269. This research is also supported by the Canadian Institute for Advanced Research (CIFAR AI chair program) and the EPSRC Turing AI World-Leading Research Fellowship No.EP/X040062/1 and EPSRC AI Hub No. EP/Y028872/1. The authors acknowledge the UCF Advanced Research Computing Center and Texas Advanced Computing Center (TACC) at UT Austin.

\bibliography{references}
\bibliographystyle{abbrvnat} 

\clearpage
\appendix

\appendix
\centerline{\bf \Large Appendix}

\section{Limitations} \label{app:limit}

In this work, we propose \methodname, a novel architecture for multi-task temporal graph property prediction. \crthree{Our evaluation covers ERC-20 transaction networks~\citep{shamsi2024mint} and two unseen social networks. Transfer to traffic, citation, and biological temporal graphs remains untested; broader domain coverage is needed.} 

Here, we also focus on the graph property prediction task, as we expect the model to learn the evolution of the global graph structure and transfer that learning across networks. One limitation is that we explore fewer other tasks, such as link prediction and node classification. \crtwo{Training networks have different sequence lengths, but we do not independently vary temporal duration in a controlled experiment.} This is also partly due to the nature of the dataset, where many nodes are transient and only appear once in a network; it is more difficult to generalize and predict the behavior of these one-time nodes.

\crthree{DOS is a fixed, binned spectral summary and may discard structural detail; learned attention does not make the descriptor itself learnable. Our ablations establish the value of the spectral path, not superiority over learned spectral alternatives. We do not include matched spectral-transformer or other recent graph-pretraining baselines, which require adaptation to this multi-network graph-level setting. Increasing pre-training diversity may also introduce task-dependent negative transfer or capacity constraints (Appendix~\ref{sec:cr_scaling}). Future directions include broader multi-domain pre-training, self-supervised objectives, and prompt- or head-based adaptation.}

\section{Broader Impact} \label{app:broader_impact}

This work advances temporal graph learning by enabling a single model to handle multiple tasks and transfer across networks without retraining. This capability reduces computational cost and improves the transferability of TGNNs, with potential impact in areas such as network security, financial fraud detection, and real-time social network analysis, where evolving systems require continuously adaptive, multi-objective modeling.

Moreover, the principles of multi-network and multi-task pre-training introduced here may inspire similar advances in other temporal data domains, including healthcare, transportation, and climate science. More broadly, this work opens a new direction toward generalizable temporal graph models that can be trained across distinct networks, taking an important step toward Temporal Graph Foundation Models.

\section{Computational Complexity of \methodname}
\label{sec:complexity}
\crone{\noindent\textbf{DOS computation.} We rescale the normalized Laplacian as $\widetilde{\mat L}=\mat{L}_{\mathrm{sym}}-\mat I$, mapping its spectrum from $[0,2]$ to $[-1,1]$. KPM estimates 50 Chebyshev moments using 50 random probing vectors and stochastic trace estimation. The recurrence $T_j(x)=2xT_{j-1}(x)-T_{j-2}(x)$ requires only sparse matrix--vector products. Jackson damping smooths the reconstructed density, which we map back to $[0,2]$ and discretize into 20 equal-width bins. The complexity below uses $N$ for the moment count and treats the probe count as fixed.}

For a snapshot $G_t = (V_t,E_t)$ with $n = |V_t|$ nodes and $m = |E_t|$ edges, the per-snapshot complexity of \methodname is  
$$
\mathcal{O}\!\big(m \cdot d + n \cdot d^2 + n \cdot \log n + N \cdot m + K \cdot d\big),
$$  
where $d$ is the hidden dimension, $K$ is the number of task-specific heads, and $N$ is the number of Chebyshev moments used to approximate the DOS with a kernel polynomial method. The term $m \cdot d$ arises from spatial message passing, since each edge propagates a $d$-dimensional embedding. The term $n \cdot d^2$ comes from recurrent or attention-based updates. Pooling with top-$k$ selection contributes $n \cdot \log n$, reflecting attention-score computation for all nodes followed by partial sorting. The spectral DOS module adds $N \cdot m$, as each moment requires one sparse matrix–vector multiplication. Finally, the cost of task-specific heads is $K \cdot d$. Over $|\mathcal{D}|$ networks with $T$ snapshots each, the overall training cost scales as  
$$
\mathcal{O}\!\left( |\mathcal{D}| \cdot T \,[\, m \cdot (d + N) + n \cdot d^2 + n \cdot \log n + K \cdot d \,] \right).
$$

In our implementation, the spatial hidden dimension is fixed to $16$, the DOS descriptor dimension is $20$, and we train with $K=3$ task-specific heads. Substituting these values, the per-snapshot complexity of \methodname becomes  
$$
\mathcal{O}\!\big(16 \cdot m + 256 \cdot n + n \log n + N \cdot m + 108\big),
$$  
where $m=|E_t|$ and $n=|V_t|$. The terms correspond to spatial message passing ($16 \cdot m$), recurrent/attention updates ($256 \cdot n$), pooling with top-$k$ selection ($n \log n$), DOS spectral approximation with $N$ Chebyshev moments ($N \cdot m$), and three task heads on a $36$-dimensional joint embedding ($108$). For each snapshot, the overall training complexity scales as  
$$
\mathcal{O}\!\left(  \big((16 + N)\,m + 256\,n + n \log n  \big)\right).
$$  

In practice, with hidden size and Chebyshev moment parameters fixed, the per-snapshot complexity of \methodname reduces to 
$\mathcal{O}(N \cdot m + n \log n)$, dominated by sparse matrix–vector multiplications in the DOS module.

\section{Dataset}
\label{sec:datasetStats}

We evaluate \methodname on temporal transaction networks from the MiNT benchmark \citep{shamsi2024mint}, the first large-scale dataset designed for training and evaluating temporal graph models across multiple heterogeneous networks. MiNT contains 84 ERC-20 token transaction graphs collected from the Ethereum blockchain between 2017 and 2023, spanning more than six years of activity. The biggest  MiNT token network contains \(128,159\) unique addresses and \(554,705\) transactions, while the smallest token network has \(1,454\) nodes.  Each network is represented as a sequence of weekly snapshots, where nodes correspond to wallet addresses and edges represent token transfers. Across their full duration, networks range from tens of thousands to over 100K unique nodes and up to several million edges, reflecting the scale of real-world token ecosystems.

A distinctive feature of MiNT is its strong inductiveness: new addresses and transactions appear continuously, making prediction inherently open-world. This reflects the real behavior of blockchain networks, where adoption cycles, liquidity shocks, and market events constantly introduce novel participants and structural patterns. The long temporal duration further amplifies this effect, as networks exhibit bursts of rapid growth, periods of stagnation, and sudden fragmentation or collapse. Such novelty and surprise factors make MiNT particularly challenging, as models must generalize across highly dynamic trajectories rather than stationary or repetitive patterns.

\begin{table*}[htbp]
  \centering
  \caption{All token networks' statistics.}
  \label{tab:TGS_all_stats}
  \resizebox{\linewidth}{!}{%
    \begin{tabular}{lrrclll | lrrclll}
    \toprule 
  
      Token  & \#Node & \#Transaction & \#Snapshots (days) & Growth rate & Novelty & Surprise & Token &\#Node & \#Transaction & \#Snapshots (days) & Growth rate & Novelty & Surprise  \\ \midrule

      ARC & 11325 & 70968 & 606 & 0.43 & 0.32 & 0.88 & Metis & 52586 & 343141 & 907 & 0.44 & 0.48 & 0.89\\
CELR & 65350 & 235807 & 1691 & 0.49 & 0.56 & 0.96 & cDAI & 52753 & 358050 & 1437 & 0.45 & 0.46 & 0.9\\
CMT & 86895 & 205961 & 309 & 0.45 & 0.72 & 0.92 & BITCOIN & 34051 & 347054 & 178 & 0.48 & 0.39 & 0.63\\
DRGN & 113453 & 341849 & 2164 & 0.44 & 0.57 & 0.97 & INJ & 60472 & 312822 & 1113 & 0.46 & 0.52 & 0.98\\
GHST & 35156 & 180955 & 1146 & 0.43 & 0.51 & 0.93 & MIM & 23038 & 269366 & 885 & 0.44 & 0.4 & 0.89\\
INU & 8556 & 66315 & 154 & 0.27 & 0.41 & 0.59 & GLM & 53385 & 234912 & 1080 & 0.5 & 0.53 & 0.96\\
IOTX & 63079 & 288469 & 1993 & 0.45 & 0.56 & 0.99 & Mog & 14590 & 240680 & 107 & 0.37 & 0.38 & 0.55\\
QSP & 117977 & 299671 & 2178 & 0.45 & 0.67 & 0.99 &DPI & 40627 & 234246 & 1150 & 0.49 & 0.5 & 0.86\\

REP & 83282 & 224843 & 346 & 0.46 & 0.69 & 0.96 &LINA & 45342 & 227147 & 1144 & 0.45 & 0.46 & 0.95\\

RFD & 23208 & 173695 & 169 & 0.3 & 0.39 & 0.6 & Yf-DAI & 22466 & 226875 & 1158 & 0.42 & 0.31 & 0.87\\

TNT & 88247 & 316352 & 1216 & 0.43 & 0.55 & 0.93 & BOB & 42806 & 212099 & 199 & 0.35 & 0.48 & 0.73\\

TRAC & 71667 & 299181 & 2110 & 0.46 & 0.54 & 0.97 & RGT & 35277 & 211932 & 1110 & 0.44 & 0.46 & 0.98\\

RLB & 28033 & 240291 & 129 & 0.43 & 0.49 & 0.76 & TVK & 42539 & 208082 & 1062 & 0.41 & 0.48 & 0.93\\
 
steCRV & 19079 & 211538 & 1033 & 0.45 & 0.53 & 0.9 & RSR & 50645 & 205906 & 659 & 0.47 & 0.62 & 0.91\\

ALBT & 63042 & 434881 & 1152 & 0.43 & 0.44 & 0.89 & WOJAK & 34341 & 198653 & 201 & 0.37 & 0.48 & 0.73\\

POLS & 128159 & 554705 & 1132 & 0.45 & 0.61 & 0.94 & ANT & 36517 & 200262 & 1107 & 0.47 & 0.46 & 0.93\\

SWAP & 69230 & 509769 & 1213 & 0.46 & 0.45 & 0.79 & LADYS & 37486 & 192176 & 181 & 0.37 & 0.52 & 0.79\\

SUPER & 83299 & 502030 & 986 & 0.47 & 0.46 & 0.85 & ETH2x-FLI & 11008 & 199088 & 965 & 0.47 & 0.28 & 0.84\\

RARI & 87186 & 502960 & 1207 & 0.43 & 0.47 & 0.91 & TURBO & 38638 & 189048 & 189 & 0.33 & 0.48 & 0.72\\

KP3R & 39323 & 493258 & 1102 & 0.43 & 0.33 & 0.88 & REPv2 & 39061 & 191367 & 1194 & 0.48 & 0.5 & 0.97\\

MIR & 79984 & 444998 & 1066 & 0.45 & 0.43 & 0.92 & NOIA & 29798 & 185528 & 1133 & 0.46 & 0.37 & 0.7\\

aUSDC & 23742 & 475680 & 1067 & 0.46 & 0.4 & 0.73 & 0x0 & 21531 & 182430 & 283 & 0.51 & 0.46 & 0.81\\

LUSD & 25852 & 430473 & 943 & 0.48 & 0.36 & 0.87 & PSYOP & 25450 & 168896 & 169 & 0.32 & 0.39 & 0.59\\

PICKLE & 28498 & 430262 & 1149 & 0.48 & 0.34 & 0.69 & ShibDoge & 40023 & 134697 & 680 & 0.43 & 0.53 & 0.8\\

DODO & 47046 & 390443 & 1131 & 0.47 & 0.45 & 0.91 & ADX & 14567 & 123755 & 1188 & 0.44 & 0.4 & 0.91\\

YFII & 43964 & 391984 & 1196 & 0.44 & 0.44 & 0.96 & BAG & 11860 & 122634 & 298 & 0.31 & 0.44 & 0.87\\

STARL & 71590 & 369913 & 856 & 0.46 & 0.48 & 0.86 & QOM & 21757 & 118292 & 598 & 0.46 & 0.41 & 0.81\\
 
LQTY & 34687 & 374230 & 943 & 0.45 & 0.34 & 0.91 & BEPRO & 26521 & 120261 & 1132 & 0.46 & 0.48 & 0.87\\

FEG & 118294 & 367584 & 1007 & 0.4 & 0.62 & 0.92 & AIOZ & 29231 & 119926 & 947 & 0.43 & 0.49 & 0.89\\

AUDIO & 91218 & 362685 & 1108 & 0.45 & 0.58 & 0.95 & PRE & 40476 & 118625 & 1113 & 0.5 & 0.55 & 0.86\\

OHM & 45728 & 377068 & 690 & 0.43 & 0.46 & 0.88 & CRU & 19990 & 117712 & 1144 & 0.5 & 0.43 & 0.95\\

WOOL & 16874 & 351178 & 716 & 0.41 & 0.18 & 0.41 & POOH & 27245 & 111641 & 193 & 0.26 & 0.49 & 0.69\\

DERC & 24277 & 111205 & 824 & 0.45 & 0.49 & 0.83 & aDAI & 13648 & 187050 & 1068 & 0.45 & 0.46 & 0.82\\

stkAAVE & 37355 & 110924 & 1128 & 0.42 & 0.57 & 0.71 & ORN & 44010 & 239451 & 1134 & 0.46 & 0.47 & 0.87\\

BTRFLY & 8450 & 108371 & 453 & 0.48 & 0.34 & 0.44 & DOGE2.0 & 7664 & 79047 & 123 & 0.45 & 0.38 & 0.66\\

SDEX & 9127 & 104869 & 240 & 0.41 & 0.44 & 0.75 & HOICHI & 5075 & 77361 & 436 & 0.36 & 0.32 & 0.71\\

XCN & 20085 & 104185 & 607 & 0.46 & 0.42 & 0.84 & EVERMOON & 7552 & 79868 & 163 & 0.24 & 0.35 & 0.52\\

HOP & 37004 & 102650 & 514 & 0.41 & 0.6 & 0.88& MUTE & 12426 & 82345 & 977 & 0.43 & 0.46 & 0.95\\

MAHA & 18401 & 96180 & 749 & 0.43 & 0.47 & 0.91 & crvUSD & 2950 & 88647 & 174 & 0.61 & 0.37 & 0.73\\

DINO & 15837 & 94140 & 358 & 0.44 & 0.44 & 0.74& SLP & 6675 & 95368 & 1151 & 0.43 & 0.36 & 0.91\\

bendWETH & 1454 & 96898 & 593 & 0.51 & 0.21 & 0.51& sILV2 & 12838 & 92905 & 611 & 0.4 & 0.34 & 0.48\\

PUSH & 14501 & 93103 & 936 & 0.46 & 0.38 & 0.83 & SPONGE & 25852 & 90468 & 184 & 0.31 & 0.66 & 0.81\\

\bottomrule
    \end{tabular}
    }
\end{table*}

We summarize detailed statistics of each token network in the MiNT datasets in Table \ref{tab:TGS_all_stats}.   Most networks have more than 10k nodes and over 100k edges. The lifespan of MiNT networks varies from \(107\) days to \(6\) years, and each day has at least one transaction. As the table shows, the token networks have quite high surprise values with an average of $0.82$.  

For our experiments, we follow the MiNT protocol and split the dataset into 64 financial networks for training and 20 additional networks for unseen testing. This setup lets \methodname learn transferable temporal representations from diverse source networks and evaluate zero-shot generalization on new target networks available only at inference time.

\section{Baselines}
\label{appendix:baselines}
In this section, we provide further details on the temporal graph learning models we used as baselines for our work.

\textbf{TGCN} ~\citep{ling2020tgcn} is a combination of GCN and GRU. In particular, GCN is used to learn complex topological structures, while GRU is used to model embeddings dynamically to capture temporal dependence.

\textbf{HTGN}~\citep{menglin2021HTGN} leverages the power of hyperbolic geometry, which is well-suited for capturing hierarchical structures and complex relationships in temporal networks. HTGN maps the temporal graph into hyperbolic space and utilizes hyperbolic graph neural networks and hyperbolic gated recurrent neural networks to model the evolving dynamics. It incorporates two key modules that are hyperbolic temporal contextual self-attention (HTA) and hyperbolic temporal consistency (HTC)-to ensure that temporal dependencies are effectively captured and that the model is both stable and generalizable across various tasks.

\textbf{GraphPulse}~\citep{shamsi2024graphpulse} addresses the challenge of learning from nodes and edges with different timestamps, which many existing models struggle with. It combines two key techniques: the Mapper method from topological data analysis to extract clustering information from graph nodes and Recurrent Neural Networks (RNNs) for temporal reasoning. This principled approach helps capture both the structure and dynamics of evolving graphs.

\textbf{GCLSTM}~\citep{chen2022GCLSTM} combines a Graph Convolutional Network (GCN) and Long Short-Term Memory (LSTM) units to handle both the structural and temporal aspects of evolving networks. The GCN is used to capture the local structural properties of the network at each snapshot, while the LSTM learns the temporal evolution of these snapshots over time.

\textbf{EvolveGCN}~\citep{Aldo2020EvolveGCN} is designed to capture the temporal dynamics of graph-structured data. Instead of relying on static node embeddings, EvolveGCN evolves the parameters of a graph convolutional network (GCN) over time. By using a recurrent neural network (RNN) to adapt the GCN parameters, this model is capable of dynamically adjusting during both training and testing, allowing it to handle evolving graphs, even when node sets vary significantly across different time steps.

\textbf{ROLAND}~\citep{you2022roland} is a dynamic graph learning framework that models node representations as hierarchical states, updated recurrently to capture temporal dependencies in evolving graphs. It supports scalable training using techniques like truncated backpropagation through time and meta-learning. In our DTDG setting, we use ROLAND to benchmark its performance and adaptability across diverse temporal networks.

\textbf{WinGNN}~\citep{yifan2023wingnn} uses a simple GNN to model topological information from the graph as other models existing in the literature. However, to model temporal dependencies, WinGNN proposes a novel mechanism of random gradient aggregation and a meta-learning strategy. In particular, WinGNN computes the frame-wise loss of the current snapshot and passes the loss gradient to the next to model graph dynamics without using RNN-based modules. Then it introduces the randomized sliding window to acquire the window-aware gradient on consecutive snapshots, and the two types of gradients are aggregated to update the GNN modules.

\section{Task Formalizations}
\label{appendix:task_definition}
In this section, we provide detailed definitions for all classification and regression tasks considered in \methodname. Each temporal snapshot corresponds to a 7-day interval, and the property prediction setup follows GraphPulse~\citep{shamsi2024graphpulse}. We also note that some of these tasks and the corresponding labels were processed specifically for this work, ensuring consistency and comparability across networks.

Setting \(n = 7\), \(\delta_1 = 3\), and \(\delta_2 = 10\) days, we establish a practical graph property with a 7-day prediction window. This choice is particularly relevant in financial contexts, such as Ethereum asset networks, where it can guide investment decisions~\citep{abay2019chainnet}.

\subsection{Classification Tasks}

\textbf{Node Growth/Shrinkage (Node G/S).}  
Let $V(t_1,t_n)$ denote the set of unique nodes active between times $t_1$ and $t_n$. The task predicts whether the number of nodes increases in the prediction interval $[t_{n+\delta_1},t_{n+\delta_2}]$:
\begin{equation}
P_{\text{nodes}}(\mathcal{G}, t_1, t_n, \delta_1, \delta_2) =
\begin{cases}
1, & \text{if } |V(t_{n+\delta_1}, t_{n+\delta_2})| > |V(t_1,t_n)|, \\
0, & \text{otherwise.}
\end{cases}
\end{equation}
\textbf{Importance.} Node growth measures adoption, reflecting the entry of new addresses, while shrinkage signals attrition. In token ecosystems, this corresponds to market expansion or decline.

\textbf{Edge Growth/Shrinkage (Edge G/S).}  
Let $E(t_1,t_n)$ denote the set of transactions in $[t_1,t_n]$. The model predicts whether transaction activity grows in the next interval:
\begin{equation}
P_{\text{edges}}(\mathcal{G}, t_1, t_n, \delta_1, \delta_2) =
\begin{cases}
1, & \text{if } |E(t_{n+\delta_1}, t_{n+\delta_2})| > |E(t_1,t_n)|, \\
0, & \text{otherwise.}
\end{cases}
\end{equation}
\textbf{Importance.} Edge growth captures changes in liquidity and market engagement, with direct implications for trading activity and token valuation.

\textbf{Largest Connected Component Growth/Shrinkage (LCC G/S).}  
Let $C(t_1,t_n)$ denote the size of the largest connected component during $[t_1,t_n]$. The task is to predict whether connectivity expands:
\begin{equation}
P_{\text{LCC}}(\mathcal{G}, t_1, t_n, \delta_1, \delta_2) =
\begin{cases}
1, & \text{if } |C(t_{n+\delta_1}, t_{n+\delta_2})| > |C(t_1,t_n)|, \\
0, & \text{otherwise.}
\end{cases}
\end{equation}
\textbf{Importance.} LCC growth reflects stronger structural integration, improving liquidity and market stability in blockchain networks.

\subsection{Regression Tasks}

\textbf{New Node Count.}  
Let $V_{\text{new}}(t_{n+\delta_1}, t_{n+\delta_2})$ denote the set of nodes that appear for the first time in the prediction interval. The regression target is:
\begin{equation}
\hat{y}_{\text{new-nodes}}(\mathcal{G}, t_1, t_n, \delta_1, \delta_2) = |V_{\text{new}}(t_{n+\delta_1}, t_{n+\delta_2})|.
\end{equation}
\textbf{Importance.} New node prediction quantifies adoption and user acquisition, a key indicator of ecosystem growth.

\textbf{Edge Count.}  
The task is to predict the number of transactions in the future interval:
\begin{equation}
\hat{y}_{\text{edges}}(\mathcal{G}, t_1, t_n, \delta_1, \delta_2) = |E(t_{n+\delta_1}, t_{n+\delta_2})|.
\end{equation}
\textbf{Importance.} Transaction forecasts provide fine-grained estimates of liquidity and demand surges in decentralized markets.

\textbf{Influential Node Count.}  
Define influential nodes as those with a degree greater than 5 in the prediction interval. Empirically, degree distributions of ERC-20 token graphs are heavy-tailed~\citep{shamsi2024mint}: most nodes appear only once or twice, and a small fraction appear hundreds or thousands of times. When we plot the cumulative distribution of node degrees, there is usually a sharp drop in frequency after degree 1–2, followed by a long but thinner tail. Setting the threshold at 5 sits just beyond this long tail cutoff, ensuring that only the top 10–20\% of nodes in activity are retained as influential. In practice, this excludes wallets that perform only a handful of transfers while capturing the repeat participants who actually shape liquidity and flow. 

Let $V_{\text{inf}}(t_{n+\delta_1}, t_{n+\delta_2}) = \{ v \in V : \deg(v) > 5 \}$. The regression target is:
\begin{equation}
\hat{y}_{\text{inf}}(\mathcal{G}, t_1, t_n, \delta_1, \delta_2) = |V_{\text{inf}}(t_{n+\delta_1}, t_{n+\delta_2})|.
\end{equation}
\textbf{Importance.} Influential nodes represent hubs such as active traders, liquidity providers, or contracts, which shape the stability and price dynamics of token ecosystems.

\subsection{Importance of Temporal Property Prediction}
\revThree{Monitoring global structural dynamics in temporal graphs is essential in domains where the evolution of the entire network, rather than individual edges, drives operational decisions. In blockchain transaction networks, most transactions occur only once and do not repeat, so link-level prediction is not informative. Graph properties instead track global indicators such as changes in connectivity, activity, and influential participants, which often serve as early signals of liquidity risk, instability, or ecosystem decline~\citep{abay2019chainnet,gurcan2021blockchain}. Since thousands of Ethereum-based tokens evolve with their own user bases and activity patterns~\citep{ZhuKA24}, training a separate temporal model for each network is not practical. A single model that learns from many heterogeneous networks and jointly predicts several graph properties offers a more scalable and informative solution.
Similar considerations arise in communication, payment, and social systems, where forecasting network-level behaviors helps detect anomalies, anticipate demand, and understand system health~\citep{kazemi2020representation}. A multi-task approach is particularly valuable because these properties move together and provide complementary signals, leading to more reliable and actionable predictions than solving each task independently.}
\section{Evaluation Metrics}  
\label{appendix:metrics}  

We evaluate models using both performance scores and ranking-based statistics. Below we provide formal definitions of each metric used in the main paper.  

\paragraph{First-Place Count.}  
For a given task, the \emph{first-place count} measures how often a method achieves the top performance across datasets. Let $\mathcal{D}$ denote the set of datasets and $\mathcal{M}$ the set of methods. For dataset $d \in \mathcal{D}$, let
\(
m^*(d) = \arg\max_{m \in \mathcal{M}} \mathrm{Perf}(m, d),
\) 
where $\mathrm{Perf}(m,d)$ is the task-specific score (AUC for classification, $\mathrm{MAE}$ for regression). Then the first-place count for method $m$ is
\(
\mathrm{First}(m) = \sum_{d \in \mathcal{D}} \mathbf{1}[m = m^*(d)].
\)  

\paragraph{Average Rank.}  
Each method is ranked per dataset according to performance. Let $\mathrm{rank}(m,d)$ be the rank of method $m$ on dataset $d$ (lower is better). The average rank of method $m$ is
\(
\mathrm{AvgRank}(m) = \frac{1}{|\mathcal{D}|} \sum_{d \in \mathcal{D}} \mathrm{rank}(m,d).
\) 

\paragraph{Average AUC (Classification).}  
For binary classification, the Area Under the ROC Curve (AUC) for dataset $d$ is
\(
\mathrm{AUC}(d) = \Pr\!\big(\hat{y}_i > \hat{y}_j \,\big|\, y_i=1,\, y_j=0 \big),
\)
where $\hat{y}_i$ are predicted scores and $y_i \in \{0,1\}$ are ground-truth labels. The average AUC of method $m$ is
\(
\mathrm{AvgAUC}(m) = \frac{1}{|\mathcal{D}|} \sum_{d \in \mathcal{D}} \mathrm{AUC}(m,d).
\)  

\paragraph{Average MAE (Regression).}  
For regression, the Mean Absolute Error (MAE) on dataset $d$ is
\(
\mathrm{MAE}(d) = \frac{1}{N_d} \sum_{i=1}^{N_d} |\hat{y}_i - y_i|,
\)
where $\hat{y}_i$ are predictions and $y_i$ ground truth labels. The average MAE of method $m$ is
\(
\mathrm{AvgMAE}(m) = \frac{1}{|\mathcal{D}|} \sum_{d \in \mathcal{D}} \mathrm{MAE}(m,d).
\)  

\paragraph{Relative Gain Across Tasks.}  
Relative gain quantifies \methodname's improvement over the strongest baseline by comparing the average metric values across all 20 test datasets for each task. For classification (where higher AUC is better), the gain for task $t$ is defined as
\[
\mathrm{Gain}_t = \frac{\overline{\mathrm{AUC}}_{\text{Hydra}}(t) - \max_{m \in \mathcal{B}} \overline{\mathrm{AUC}}_m(t)}{\max_{m \in \mathcal{B}} \overline{\mathrm{AUC}}_m(t)} \times 100\%,
\]
where $\overline{\mathrm{AUC}}_m(t)$ denotes the average AUC of method $m$ on task $t$ across all test datasets.  

For regression (where lower MAE is better), the gain is
\[
\mathrm{Gain}_t = \frac{\min_{m \in \mathcal{B}} \overline{\mathrm{MAE}}_m(t) - \overline{\mathrm{MAE}}_{\text{Hydra}}(t)}{\min_{m \in \mathcal{B}} \overline{\mathrm{MAE}}_m(t)} \times 100\%,
\]
where $\overline{\mathrm{MAE}}_m(t)$ denotes the average MAE of method $m$ on task $t$ across all test datasets.  

Finally, the overall gain across a group of tasks $\mathcal{T}$ (e.g., all classification tasks) is computed as the mean of the per-task gains:
\(
\mathrm{Gain}_{\mathcal{T}} = \frac{1}{|\mathcal{T}|} \sum_{t \in \mathcal{T}} \mathrm{Gain}_t.
\)

\crtwo{For LCC, node, and edge growth/shrinkage, the relative AUC gains are 10.64\%, 12.58\%, and 3.58\%, respectively (mean 8.9\%). For new-node and edge-count regression, the relative MAE reductions are 10.26\% and 6.12\% (mean 8.2\%); this regression summary excludes influential-node count, where Hydra is not the best model, although it remains competitive without any specific training on tested networks.}

\section{Ablation Studies}
\label{appendix:ablation_stury}

\begin{table*}[ht]
\centering

\begin{minipage}[t]{0.47\textwidth}
\centering
\caption{DOS ablation study on task Edge G/S. 
AUC values are reported. Best results per dataset are in \textbf{bold}.}
\label{tab:dos_ablation}
\scriptsize
\renewcommand{\arraystretch}{1.05}
\setlength{\tabcolsep}{6pt}
\begin{tabular*}{\textwidth}{@{\extracolsep{\fill}} lcc @{}}
\toprule
\textbf{Dataset} & \methodname \textbf{w/o DOS} & \methodname \textbf{w DOS} \\
\midrule
MIR       & \textbf{0.791 (±0.014)} & 0.755 (±0.013) \\
DOGE2.0   & 0.500 (±0.102) & \textbf{0.732 (±0.044)} \\
MUTE      & 0.675 (±0.030) & \textbf{0.699 (±0.011)} \\
EVERMOON  & 0.426 (±0.129) & \textbf{0.796 (±0.219)} \\
DERC      & 0.748 (±0.029) & \textbf{0.761 (±0.030)} \\
ADX       & 0.678 (±0.019) & \textbf{0.703 (±0.006)} \\
HOICHI    & \textbf{0.792 (±0.052)} & 0.572 (±0.016) \\
SDEX      & \textbf{0.599 (±0.134)} & 0.261 (±0.129) \\
BAG       & 0.759 (±0.294) & \textbf{0.874 (±0.017)} \\
XCN       & 0.822 (±0.073) & \textbf{0.825 (±0.038)} \\
ETH2x-FLI & 0.716 (±0.020) & \textbf{0.736 (±0.026)} \\
stkAAVE   & 0.708 (±0.021) & \textbf{0.719 (±0.006)} \\
GLM       & 0.755 (±0.194) & \textbf{0.810 (±0.015)} \\
QOM       & 0.639 (±0.015) & \textbf{0.698 (±0.024)} \\
WOJAK     & 0.548 (±0.081) & \textbf{0.633 (±0.081)} \\
DINO      & 0.672 (±0.044) & \textbf{0.843 (±0.031)} \\
Metis     & \textbf{0.785 (±0.050)} & 0.713 (±0.027) \\
REPv2     & \textbf{0.761 (±0.039)} & 0.736 (±0.028) \\
TRAC      & \textbf{0.794 (±0.041)} & 0.732 (±0.007) \\
BEPRO     & 0.775 (±0.022) & \textbf{0.786 (±0.009)} \\
\midrule
\textbf{$1\textsuperscript{st}$-Place Count$\uparrow$} & 6 & \textbf{14} \\
\textbf{Avg. Rank $\downarrow$} & 1.70 & \textbf{1.30} \\
\textbf{Avg. AUC $\uparrow$}    & 0.697 & \textbf{0.719} \\
\bottomrule
\end{tabular*}
\end{minipage}%
\hfill
\begin{minipage}[t]{0.47\textwidth}
\centering
\caption{SAG pooling ablation study on task Edge G/S. 
AUC values are reported. Best results per dataset are in \textbf{bold}.}
\label{tab:pooling_ablation}
\scriptsize
\renewcommand{\arraystretch}{1.05}
\setlength{\tabcolsep}{6pt}
\begin{tabular*}{\textwidth}{@{\extracolsep{\fill}} lcc @{}}
\toprule
\textbf{Dataset} & \methodname \textbf{w/o Pooling} & \methodname \textbf{w Pooling} \\
\midrule
MIR       & 0.755 (±0.013) & \textbf{0.793 (±0.002)} \\
DOGE2.0   & 0.732 (±0.044) & \textbf{0.897 (±0.089)} \\
MUTE      & 0.699 (±0.011) & \textbf{0.701 (±0.098)} \\
EVERMOON  & 0.796 (±0.219) & \textbf{0.818 (±0.046)} \\
DERC      & 0.761 (±0.030) & \textbf{0.839 (±0.008)} \\
ADX       & 0.703 (±0.006) & \textbf{0.722 (±0.087)} \\
HOICHI    & 0.572 (±0.016) & \textbf{0.591 (±0.103)} \\
SDEX      & 0.261 (±0.129) & \textbf{0.348 (±0.066)} \\
BAG       & 0.874 (±0.017) & \textbf{0.969 (±0.008)} \\
XCN       & 0.825 (±0.038) & \textbf{0.844 (±0.012)} \\
ETH2x-FLI & \textbf{0.736 (±0.026)} & 0.712 (±0.045) \\
stkAAVE   & 0.719 (±0.006) & \textbf{0.732 (±0.011)} \\
GLM       & 0.810 (±0.015) & \textbf{0.850 (±0.009)} \\
QOM       & 0.698 (±0.024) & \textbf{0.745 (±0.003)} \\
WOJAK     & \textbf{0.633 (±0.081)} & 0.585 (±0.067) \\
DINO      & 0.843 (±0.031) & \textbf{0.895 (±0.003)} \\
Metis     & 0.713 (±0.027) & \textbf{0.733 (±0.004)} \\
REPv2     & 0.736 (±0.028) & \textbf{0.772 (±0.016)} \\
TRAC      & \textbf{0.732 (±0.007)} & 0.722 (±0.001) \\
BEPRO     & 0.786 (±0.009) & \textbf{0.800 (±0.002)} \\
\midrule
\textbf{$1\textsuperscript{st}$-Place Count$\uparrow$}  & 3 & \textbf{17} \\
\textbf{Avg. Rank $\downarrow$} & 1.85 & \textbf{1.15} \\
\textbf{Avg. AUC $\uparrow$}    & 0.719 & \textbf{0.753} \\
\bottomrule
\end{tabular*}
\end{minipage}

\end{table*}

To better understand \methodname components, we conduct an ablation study on the edge growth/shrinkage task. The ablations reflect different stages in the model’s incremental development. First, we compare the initial version of \methodname without pooling, evaluating the impact of adding Density of States (DOS) features in the spectral path. Next, we consider models that already include DOS and assess the effect of introducing attention-based pooling in the spatial path. Results are reported in Appendix Tables~\ref{tab:dos_ablation} and \ref{tab:pooling_ablation}. Together, these comparisons show that DOS enhances the capture of global spectral structure, while pooling improves the selection of informative subgraphs, confirming their complementary roles in \methodname's design and performance.  

\noindent\textbf{Impact of DOS.} To assess the contribution of the DOS module, we compare \methodname with and without DOS features (Table~\ref{tab:dos_ablation}). This setup isolates the effect of spectral information on predictive performance, holding all other components fixed. The results show that incorporating DOS substantially improves performance on most datasets: \methodname with DOS achieves the best results in 14 out of 20 networks, compared to 6 without DOS. On average, DOS raises AUC from 0.697 to 0.719 and improves the mean rank from 1.70 to 1.30. Gains are particularly large on challenging datasets such as \textit{EVERMOON}, \textit{DOGE2.0}, and \textit{DINO}, where DOS more effectively captures global spectral structure. These findings highlight that DOS provides complementary information to the spatial path, leading to stronger generalization across temporal networks.  

\begin{wraptable}{r}{0.54\textwidth}
\centering
\caption{\crtwo{DOS ablation for edge-count regression.
Values are means with standard deviations in parentheses
over three runs. Lower is better; best means per dataset
are in bold.}}
\label{tab:cr_dos_reg}
\vspace{-8pt}
\scriptsize
\renewcommand{\arraystretch}{1.05}
\setlength{\tabcolsep}{3pt}

\newcommand{\dosstd}[1]{{\tiny ($\pm$#1)}}

\begin{tabular*}{\linewidth}
{@{\extracolsep{\fill}}lcc@{}}
\toprule
\textbf{Dataset}
& \shortstack{\methodname\\\textbf{w/o DOS}}
& \shortstack{\methodname\\\textbf{w DOS}} \\
\midrule
MIR
& 0.017 \dosstd{0.005}
& \textbf{0.013} \dosstd{0.004} \\
DOGE2.0
& 0.183 \dosstd{0.015}
& \textbf{0.092} \dosstd{0.008} \\
MUTE
& \textbf{0.025} \dosstd{0.012}
& \textbf{0.025} \dosstd{0.005} \\
EVERMOON
& 0.015 \dosstd{0.004}
& \textbf{0.012} \dosstd{0.005} \\
DERC
& 0.016 \dosstd{0.008}
& \textbf{0.015} \dosstd{0.004} \\
ADX
& 0.024 \dosstd{0.013}
& \textbf{0.016} \dosstd{0.005} \\
HOICHI
& \textbf{0.029} \dosstd{0.007}
& \textbf{0.029} \dosstd{0.008} \\
SDEX
& 0.088 \dosstd{0.006}
& \textbf{0.063} \dosstd{0.006} \\
BAG
& \textbf{0.041} \dosstd{0.002}
& 0.052 \dosstd{0.002} \\
XCN
& 0.060 \dosstd{0.020}
& \textbf{0.009} \dosstd{0.002} \\
ETH2x-FLI
& \textbf{0.024} \dosstd{0.006}
& 0.030 \dosstd{0.004} \\
stkAAVE
& \textbf{0.045} \dosstd{0.012}
& 0.078 \dosstd{0.017} \\
GLM
& \textbf{0.085} \dosstd{0.006}
& 0.094 \dosstd{0.007} \\
QOM
& 0.034 \dosstd{0.018}
& \textbf{0.012} \dosstd{0.001} \\
WOJAK
& 0.013 \dosstd{0.008}
& \textbf{0.009} \dosstd{0.000} \\
DINO
& \textbf{0.018} \dosstd{0.004}
& \textbf{0.018} \dosstd{0.007} \\
Metis
& \textbf{0.030} \dosstd{0.002}
& 0.034 \dosstd{0.010} \\
REPv2
& 0.106 \dosstd{0.011}
& \textbf{0.066} \dosstd{0.004} \\
TRAC
& \textbf{0.021} \dosstd{0.003}
& 0.022 \dosstd{0.005} \\
BEPRO
& 0.014 \dosstd{0.013}
& \textbf{0.004} \dosstd{0.001} \\
\midrule
\textbf{Strict wins $\uparrow$}
& 6 & \textbf{11} \\
\textbf{Reported avg. $\downarrow$}
& 0.044 & \textbf{0.034} \\
\bottomrule

\end{tabular*}
\vspace{-20pt}
\end{wraptable}

\noindent\textbf{Impact of Attention-based Pooling.} The ablation on the edge growth/shrinkage task (Table~\ref{tab:pooling_ablation}) shows the contribution of the pooling mechanism within \methodname's spatial path. When pooling is enabled, the model achieves higher AUC on 17 out of 20 datasets, with the average score improving from 0.719 to 0.753 and the mean rank from 1.85 to 1.15. The improvement is particularly noticeable when pooling helps the spatial path focus on structurally important nodes, resulting in stronger graph-level representations. These findings indicate that pooling is an essential component for producing compact, informative graph-level representations and strengthen \methodname's ability to capture temporal dynamics.

\{paragraph{Regression DOS ablation.} The additional edge-count ablation reduces reported average MAE from 0.044 to 0.034 with DOS. DOS wins on 11 networks, the variant without DOS on six, and three are tied at the reported precision (Table~\ref{tab:cr_dos_reg}). This complements the classification ablation, while improvements remain dataset-dependent.

\section{\crthree{Transfer to Social Interaction Networks}}
\label{sec:cr_social}

\begin{wraptable}{R}{0.4\linewidth}
    \centering
    \vspace{-12pt}
    \scriptsize
    \setlength{\tabcolsep}{4pt}
    \renewcommand{\arraystretch}{1.10}
    \caption{\crthree{Edge G/S on unseen social networks:
    mean AUC ($\pm$ standard deviation) over three runs.
    Hydra variants transfer zero-shot; HTGN is
    target-trained. Best mean AUC per network is bold.}}
    \label{tab:cr_social}

    \resizebox{\linewidth}{!}{%
    \begin{tabular}{@{}lcc@{}}
    \toprule
    \textbf{Model} & \textbf{MathOverflow} & \textbf{RedditB} \\
    \midrule
    HTGN
        & $\mathbf{0.788\,(\pm 0.051)}$
        & $0.656\,(\pm 0.040)$ \\
    Hydra-Social
        & $0.584\,(\pm 0.049)$
        & $0.604\,(\pm 0.084)$ \\
    Hydra-Mix
        & $0.630\,(\pm 0.016)$
        & $\mathbf{0.769\,(\pm 0.015)}$ \\
    \bottomrule
    \end{tabular}%
    }
\end{wraptable}

To assess Hydra's applicability beyond transaction
networks, we adopt a cross-domain evaluation setting similar
to MiNT, using the same social datasets and training/test
split. Hydra-Social is trained on LastFM, SuperUser,
Email-Eu, AskUbuntu, CollegeMsg, and StackOverflow;
MathOverflow and RedditB are held out for zero-shot testing.
Hydra-Mix
combines social and transaction networks, using the same
six social training networks together with 32 ERC-20
training networks. Both Hydra variants are evaluated
without target-network training or fine-tuning, whereas
HTGN is trained separately on each target network.

As shown in Table~\ref{tab:cr_social}, Hydra-Mix
improves over Hydra-Social on both held-out social networks.
On RedditB, it achieves a mean AUC of 0.769 compared with
0.656 for target-trained HTGN, demonstrating effective
transfer to an unseen social network without
target-specific training. On MathOverflow, mixed-domain
training improves Hydra's mean AUC from 0.584 to 0.630,
although it remains below HTGN's 0.788. These results
support Hydra's applicability beyond transaction networks
and the potential benefit of mixed-domain pre-training,
while showing that transfer performance remains dependent
on the target network.

\section{\crthree{Further Analysis of Scaling and Model Capacity}}
\label{sec:cr_scaling}

\begin{figure}[tb]
    \centering
    \includegraphics[width=0.5\textwidth]
    {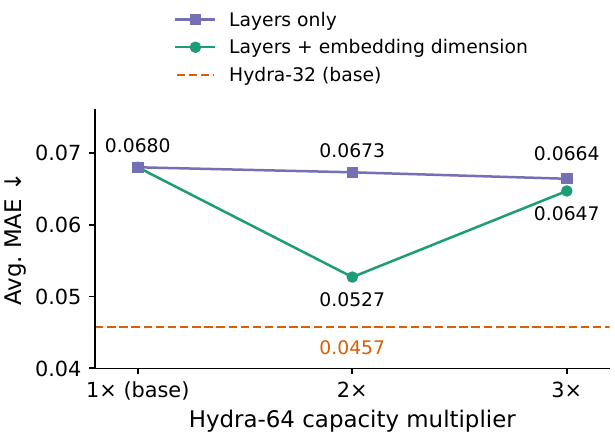}
    \caption{Effect of model capacity on Hydra-64
    edge-count prediction. Values are means over three runs,
    averaged across 20 held-out networks; variability is
    not shown. The dashed line indicates Hydra-32's base
    configuration. Lower MAE is better.}
    \label{fig:cr_capacity}
    \vspace{-2pt}
\end{figure}

\begin{wraptable}{R}{0.43\linewidth}
    \centering
    \scriptsize
    \setlength{\tabcolsep}{3pt}
    \renewcommand{\arraystretch}{1.10}

    \sisetup{
        mode = text,
        detect-weight = true,
        table-format = 1.3,
        table-model-setup = \bfseries
    }

    \caption{\crthree{Edge G/S on the 20 held-out networks:
    mean AUC ($\pm$ standard deviation) over three runs.
    Best mean AUC per network is bold.}}
    \label{tab:cr_class64_sd}

    \resizebox{\linewidth}{!}{%
    \begin{tabular}{
        @{}l
        S@{\hspace{2pt}}l
        S@{\hspace{2pt}}l
        @{}
    }
    \toprule
    \textbf{Dataset}
        & \multicolumn{2}{c}{\textbf{Hydra-32}}
        & \multicolumn{2}{c}{\textbf{Hydra-64}} \\
    \midrule
    MIR
        & 0.793 & $(\pm 0.002)$
        & \bfseries 0.796 & $(\pm 0.006)$ \\
    DOGE2.0
        & \bfseries 0.897 & $(\pm 0.089)$
        & 0.841 & $(\pm 0.009)$ \\
    MUTE
        & 0.701 & $(\pm 0.098)$
        & \bfseries 0.764 & $(\pm 0.008)$ \\
    EVERMOON
        & \bfseries 0.818 & $(\pm 0.046)$
        & 0.784 & $(\pm 0.102)$ \\
    DERC
        & \bfseries 0.839 & $(\pm 0.008)$
        & 0.825 & $(\pm 0.011)$ \\
    ADX
        & 0.722 & $(\pm 0.087)$
        & \bfseries 0.746 & $(\pm 0.031)$ \\
    HOICHI
        & 0.591 & $(\pm 0.103)$
        & \bfseries 0.643 & $(\pm 0.050)$ \\
    SDEX
        & \bfseries 0.348 & $(\pm 0.066)$
        & 0.343 & $(\pm 0.013)$ \\
    BAG
        & \bfseries 0.969 & $(\pm 0.008)$
        & 0.960 & $(\pm 0.019)$ \\
    XCN
        & \bfseries 0.844 & $(\pm 0.012)$
        & 0.836 & $(\pm 0.016)$ \\
    ETH2x-FLI
        & 0.712 & $(\pm 0.045)$
        & \bfseries 0.732 & $(\pm 0.012)$ \\
    stkAAVE
        & 0.732 & $(\pm 0.011)$
        & \bfseries 0.755 & $(\pm 0.017)$ \\
    GLM
        & \bfseries 0.850 & $(\pm 0.009)$
        & 0.840 & $(\pm 0.003)$ \\
    QOM
        & 0.745 & $(\pm 0.003)$
        & \bfseries 0.746 & $(\pm 0.014)$ \\
    WOJAK
        & 0.585 & $(\pm 0.067)$
        & \bfseries 0.608 & $(\pm 0.026)$ \\
    DINO
        & \bfseries 0.895 & $(\pm 0.003)$
        & 0.881 & $(\pm 0.018)$ \\
    Metis
        & \bfseries 0.733 & $(\pm 0.004)$
        & 0.711 & $(\pm 0.028)$ \\
    REPv2
        & 0.772 & $(\pm 0.016)$
        & \bfseries 0.774 & $(\pm 0.015)$ \\
    TRAC
        & \bfseries 0.722 & $(\pm 0.001)$
        & 0.719 & $(\pm 0.005)$ \\
    BEPRO
        & \bfseries 0.800 & $(\pm 0.002)$
        & 0.787 & $(\pm 0.012)$ \\
    \bottomrule
    \end{tabular}%
    }
    \vspace{-60pt}
\end{wraptable}

To explore future scaling directions, we extend
pre-training beyond the primary Hydra-32 configuration to
64 networks and evaluate on the same 20 unseen networks.
With the base architecture, Hydra-64 maintains a similar
mean Edge G/S AUC (0.7546 versus 0.7534), while mean
edge-count MAE increases from 0.0457 to 0.0680.
This task-dependent response motivates examining model
capacity alongside pre-training scale. Per-network
classification results, including standard deviations,
are reported in Table~\ref{tab:cr_class64_sd}.

We increase the number of learnable layers alone
or jointly increase the number of layers and the embedding
dimension by factors of two and three.
As shown in Figure~\ref{fig:cr_capacity}, doubling both
reduces Hydra-64's mean edge-count MAE from 0.0680 to
0.0527, a 22.5\% reduction, partially closing the gap to
Hydra-32. Increasing layers alone yields smaller gains,
and tripling both is less effective than doubling them.
Hydra-32 therefore remains the stronger regression
configuration in this comparison.

These findings highlight the importance of model
capacity when expanding pre-training: jointly increasing depth
and embedding dimension partially recovers regression performance,
while increasing depth alone yields smaller gains. This provides
a concrete basis for investigating how architectural choices
interact with pre-training scale. A valuable future direction
is to jointly study model capacity and training-network
composition to identify when broader pre-training translates
into improved transfer across tasks.

\section{Scaling Behavior in \methodname}
\label{appendix:scaling_behavior}

We conducted scaling-trend experiments to systematically evaluate \methodname under different training pack sizes. \crthree{Three} model variations were trained using 8, 16, and 32 networks, enabling us to observe how increasing the number of networks in the training loop affects performance. This setup provides a consistent way to study \methodname's behavior under varying amounts of training data across multiple tasks. For each task, we present results for all \crthree{three} \methodname variants. These results demonstrate how \methodname scales with the number of training networks and are shown in the following tables. Each table provides a detailed breakdown by task and model variation, offering a comprehensive view of performance under various scaling setups. Table~\ref {tab:scaling_all} reports the detailed results for each \methodname variation trained with different datapacks across all six tasks. Subsections (a)–(c) present classification tasks (Node G/S, Edge G/S, LCC G/S), while (d)–(f) correspond to regression tasks (New Node Count, Influential Node Count, Edge Count).

\begin{table*}[h]
\centering
\caption{Performance of \methodname across six tasks as the number of training networks increases from 8 to 32.}
\label{tab:scaling_all}
\scriptsize
\renewcommand{\arraystretch}{0.90}
\begin{subtable}[t]{0.49\textwidth}
\caption{\cls{Classification}: Node G/S}
\resizebox{\linewidth}{!}{\begin{tabular}{lccc}
\toprule
\textbf{Dataset} & \methodname-8 & \methodname-16 & \methodname-32 \\
\midrule
MIR       & \textbf{0.769 \scriptsize (±0.003)} & \textbf{0.769 \scriptsize (±0.007)} & \underline{0.764 \scriptsize (±0.001)} \\
DOGE2.0   & \underline{0.613 \scriptsize (±0.083)} & 0.573 \scriptsize (±0.133) & \textbf{0.633 \scriptsize (±0.064)} \\
MUTE      & \underline{0.755 \scriptsize (±0.016)} & \textbf{0.763 \scriptsize (±0.003)} & 0.748 \scriptsize (±0.025) \\
EVERMOON  & \underline{0.624 \scriptsize (±0.005)} & 0.582 \scriptsize (±0.031) & \textbf{0.655 \scriptsize (±0.040)} \\
DERC      & \underline{0.734 \scriptsize (±0.010)} & 0.725 \scriptsize (±0.006) & \textbf{0.742 \scriptsize (±0.004)} \\
ADX       & \underline{0.753 \scriptsize (±0.027)} & \textbf{0.767 \scriptsize (±0.004)} & 0.718 \scriptsize (±0.051) \\
HOICHI    & \textbf{0.582 \scriptsize (±0.029)} & 0.552 \scriptsize (±0.050) & \underline{0.558 \scriptsize (±0.047)} \\
SDEX      & \textbf{0.762 \scriptsize (±0.008)} & \underline{0.748 \scriptsize (±0.068)} & 0.743 \scriptsize (±0.026) \\
BAG       & \underline{0.963 \scriptsize (±0.010)} & 0.952 \scriptsize (±0.021) & \textbf{0.969 \scriptsize (±0.009)} \\
XCN       & \underline{0.862 \scriptsize (±0.017)} & 0.821 \scriptsize (±0.070) & \textbf{0.878 \scriptsize (±0.009)} \\
ETH2x-FLI & \textbf{0.717 \scriptsize (±0.030)} & \underline{0.710 \scriptsize (±0.009)} & 0.678 \scriptsize (±0.031) \\
stkAAVE   & \underline{0.776 \scriptsize (±0.009)} & \underline{0.776 \scriptsize (±0.003)} & \textbf{0.779 \scriptsize (±0.007)} \\
GLM       & \underline{0.750 \scriptsize (±0.011)} & 0.746 \scriptsize (±0.010) & \textbf{0.763 \scriptsize (±0.005)} \\
QOM       & 0.703 \scriptsize (±0.012) & \underline{0.707 \scriptsize (±0.013)} & \textbf{0.719 \scriptsize (±0.012)} \\
WOJAK     & \textbf{0.502 \scriptsize (±0.103)} & 0.352 \scriptsize (±0.157) & \underline{0.412 \scriptsize (±0.060)} \\
DINO      & 0.903 \scriptsize (±0.029) & \underline{0.905 \scriptsize (±0.021)} & \textbf{0.910 \scriptsize (±0.013)} \\
Metis     & \underline{0.685 \scriptsize (±0.002)} & 0.632 \scriptsize (±0.091) & \textbf{0.693 \scriptsize (±0.007)} \\
REPv2     & \textbf{0.728 \scriptsize (±0.022)} & \underline{0.721 \scriptsize (±0.023)} & 0.689 \scriptsize (±0.013) \\
TRAC      & \underline{0.756 \scriptsize (±0.005)} & 0.745 \scriptsize (±0.014) & \textbf{0.765 \scriptsize (±0.003)} \\
BEPRO     & \underline{0.858 \scriptsize (±0.023)} & 0.806 \scriptsize (±0.094) & \textbf{0.865 \scriptsize (±0.010)} \\
\midrule
\textbf{$1\textsuperscript{st}$-Place Count$\uparrow$} & 6 & 3 & \textbf{12} \\
\textbf{Avg. Rank$\downarrow$} & 1.85 & 2.45 & \textbf{1.70} \\
\textbf{Avg. AUC$\uparrow$}    & \textbf{0.740} & 0.718 & 0.734 \\
\bottomrule
\end{tabular}}
\end{subtable}\hfill
\begin{subtable}[t]{0.49\textwidth}
\caption{\cls{Classification}: Edge G/S}
\resizebox{\linewidth}{!}{\begin{tabular}{lccc}
\toprule
\textbf{Dataset} & \methodname-8 & \methodname-16 & \methodname-32 \\
\midrule
MIR       & \textbf{0.800 (±0.007)} & \underline{0.796 (±0.001)} & 0.793 (±0.002) \\
DOGE2.0   & \underline{0.859 (±0.022)} & \textbf{0.897 (±0.089)} & \textbf{0.897 (±0.089)} \\
MUTE      & \underline{0.757 (±0.004)} & \textbf{0.764 (±0.004)} & 0.701 (±0.098) \\
EVERMOON  & 0.698 (±0.180) & \underline{0.750 (±0.024)} & \textbf{0.818 (±0.046)} \\
DERC      & \underline{0.823 (±0.018)} & 0.826 (±0.002) & \textbf{0.839 (±0.008)} \\
ADX       & \underline{0.760 (±0.004)} & \textbf{0.768 (±0.015)} & 0.722 (±0.087) \\
HOICHI    & 0.606 (±0.015) & \textbf{0.617 (±0.026)} & 0.591 (±0.103) \\
SDEX      & 0.288 (±0.023) & \underline{0.331 (±0.049)} & \textbf{0.348 (±0.066)} \\
BAG       & \textbf{0.969 (±0.009)} & 0.955 (±0.015) & \textbf{0.969 (±0.008)} \\
XCN       & \textbf{0.847 (±0.006)} & \underline{0.845 (±0.010)} & 0.844 (±0.012) \\
ETH2x-FLI & \underline{0.735 (±0.006)} & \textbf{0.738 (±0.005)} & 0.712 (±0.045) \\
stkAAVE   & \textbf{0.744 (±0.008)} & 0.743 (±0.007) & 0.732 (±0.011) \\
GLM       & \underline{0.849 (±0.003)} & 0.841 (±0.015) & \textbf{0.850 (±0.009)} \\
QOM       & \underline{0.755 (±0.015)} & \textbf{0.762 (±0.008)} & 0.745 (±0.003) \\
WOJAK     & \textbf{0.627 (±0.014)} & 0.561 (±0.101) & \underline{0.585 (±0.067)} \\
DINO      & 0.875 (±0.030) & \underline{0.889 (±0.009)} & \textbf{0.895 (±0.003)} \\
Metis     & \textbf{0.735 (±0.004)} & 0.697 (±0.062) & \underline{0.733 (±0.004)} \\
REPv2     & \underline{0.778 (±0.001)} & \textbf{0.784 (±0.009)} & 0.772 (±0.016) \\
TRAC      & \underline{0.713 (±0.010)} & 0.711 (±0.011) & \textbf{0.722 (±0.001)} \\
BEPRO     & \underline{0.783 (±0.009)} & 0.743 (±0.085) & \textbf{0.800 (±0.002)} \\
\midrule
\textbf{$1\textsuperscript{st}$-Place Count$\uparrow$}   & 6 & 7 & \textbf{9} \\
\textbf{Avg. Rank$\downarrow$} & \textbf{1.95} & \textbf{1.95} & 2 \\
\textbf{Avg. AUC$\uparrow$}    & 0.75 & 0.751 & \textbf{0.753} \\
\bottomrule
\end{tabular}}
\end{subtable}\hfill

\vspace{0.4cm}

\begin{subtable}[t]{0.49\textwidth}
\caption{\cls{Classification}: LCC G/S}
\resizebox{\linewidth}{!}{\begin{tabular}{lccc}
\toprule
\textbf{Dataset} & \methodname-8 & \methodname-16 & \methodname-32 \\
\midrule
MIR       & \textbf{0.819 (±0.004)} & \underline{0.817 (±0.001)} & 0.815 (±0.007) \\
DOGE2.0   & \textbf{0.762 (±0.010)} & 0.667 (±0.175) & \underline{0.702 (±0.160)} \\
MUTE      & \underline{0.714 (±0.049)} & \textbf{0.756 (±0.006)} & 0.704 (±0.085) \\
EVERMOON  & 0.600 (±0.057) & \underline{0.639 (±0.029)} & \textbf{0.667 (±0.045)} \\
DERC      & \textbf{0.819 (±0.003)} & \underline{0.813 (±0.005)} & 0.808 (±0.022) \\
ADX       & 0.691 (±0.066) & \textbf{0.770 (±0.021)} & \underline{0.727 (±0.093)} \\
HOICHI    & 0.610 (±0.044) & \textbf{0.640 (±0.041)} & \underline{0.638 (±0.075)} \\
SDEX      & 0.784 (±0.017) & \underline{0.810 (±0.048)} & \textbf{0.816 (±0.025)} \\
BAG       & \underline{0.968 (±0.013)} & 0.957 (±0.021) & \textbf{0.976 (±0.006)} \\
XCN       & \underline{0.863 (±0.011)} & \underline{0.863 (±0.029)} & \textbf{0.887 (±0.014)} \\
ETH2x-FLI & \textbf{0.708 (±0.011)} & \underline{0.703 (±0.001)} & 0.687 (±0.032) \\
stkAAVE   & \underline{0.753 (±0.011)} & \textbf{0.757 (±0.003)} & 0.748 (±0.012) \\
GLM       & \textbf{0.853 (±0.006)} & 0.844 (±0.020) & \underline{0.848 (±0.012)} \\
QOM       & 0.725 (±0.009) & \textbf{0.730 (±0.002)} & \underline{0.729 (±0.010)} \\
WOJAK     & \textbf{0.558 (±0.032)} & 0.458 (±0.142) & \underline{0.500 (±0.083)} \\
DINO      & \underline{0.878 (±0.021)} & \textbf{0.883 (±0.032)} & 0.818 (±0.049) \\
Metis     & \underline{0.727 (±0.006)} & 0.669 (±0.100) & \textbf{0.731 (±0.001)} \\
REPv2     & \underline{0.781 (±0.004)} & \textbf{0.785 (±0.009)} & 0.764 (±0.020) \\
TRAC      & \underline{0.772 (±0.006)} & 0.768 (±0.008) & \textbf{0.781 (±0.004)} \\
BEPRO     & \underline{0.826 (±0.015)} & 0.800 (±0.058) & \textbf{0.830 (±0.006)} \\
\midrule
\textbf{$1\textsuperscript{st}$-Place Count$\uparrow$}   & 6 & \textbf{7} & \textbf{7} \\
\textbf{Avg. Rank$\downarrow$} & \textbf{1.95} & 2 & 2 \\
\textbf{Avg. AUC$\uparrow$}    & \textbf{0.761} & 0.756 & 0.759 \\
\bottomrule
\end{tabular}}
\end{subtable}\hfill
\begin{subtable}[t]{0.49\textwidth}
\caption{\rgs{Regression}: New Node Count }
\resizebox{\linewidth}{!}{\begin{tabular}{lccc}
\toprule
\textbf{Dataset} & \methodname-8 & \methodname-16 & \methodname-32 \\
\midrule
MIR       & 0.056 \scriptsize (±0.057) & \underline{0.040 \scriptsize (±0.042)} & \textbf{0.013 \scriptsize (±0.004)} \\
DOGE2.0   & \underline{0.068 \scriptsize (±0.023)} & \textbf{0.060 \scriptsize (±0.027)} & 0.092 \scriptsize (±0.008) \\
MUTE      & 0.073 \scriptsize (±0.036) & \underline{0.036 \scriptsize (±0.025)} & \textbf{0.025 \scriptsize (±0.005)} \\
EVERMOON  & 0.070 \scriptsize (±0.053) & \underline{0.038 \scriptsize (±0.045)} & \textbf{0.012 \scriptsize (±0.005)} \\
DERC      & 0.071 \scriptsize (±0.049) & \underline{0.038 \scriptsize (±0.040)} & \textbf{0.015 \scriptsize (±0.004)} \\
ADX       & 0.069 \scriptsize (±0.047) & \underline{0.029 \scriptsize (±0.035)} & \textbf{0.016 \scriptsize (±0.005)} \\
HOICHI    & 0.064 \scriptsize (±0.038) & \underline{0.048 \scriptsize (±0.030)} & \textbf{0.029 \scriptsize (±0.008)} \\
SDEX      & 0.074 \scriptsize (±0.021) & \underline{0.072 \scriptsize (±0.011)} & \textbf{0.063 \scriptsize (±0.006)} \\
BAG       & 0.072 \scriptsize (±0.023) & \underline{0.056 \scriptsize (±0.009)} & \textbf{0.052 \scriptsize (±0.002)} \\
XCN       & 0.067 \scriptsize (±0.056) & \underline{0.036 \scriptsize (±0.044)} & \textbf{0.009 \scriptsize (±0.002)} \\
ETH2x-FLI & 0.053 \scriptsize (±0.034) & \underline{0.032 \scriptsize (±0.023)} & \textbf{0.030 \scriptsize (±0.004)} \\
stkAAVE   & 0.085 \scriptsize (±0.037) & \textbf{0.057 \scriptsize (±0.024)} & \underline{0.078 \scriptsize (±0.017)} \\
GLM       & \underline{0.097 \scriptsize (±0.013)} & 0.103 \scriptsize (±0.008) & \textbf{0.094 \scriptsize (±0.007)} \\
QOM       & 0.073 \scriptsize (±0.054) & \underline{0.034 \scriptsize (±0.040)} & \textbf{0.012 \scriptsize (±0.001)} \\
WOJAK     & 0.076 \scriptsize (±0.059) & \underline{0.034 \scriptsize (±0.042)} & \textbf{0.009 \scriptsize (±0.000)} \\
DINO      & 0.065 \scriptsize (±0.053) & \underline{0.046 \scriptsize (±0.042)} & \textbf{0.018 \scriptsize (±0.007)} \\
Metis     & 0.072 \scriptsize (±0.030) & \underline{0.037 \scriptsize (±0.016)} & \textbf{0.034 \scriptsize (±0.010)} \\
REPv2     & 0.083 \scriptsize (±0.024) & \underline{0.074 \scriptsize (±0.015)} & \textbf{0.066 \scriptsize (±0.004)} \\
TRAC      & 0.074 \scriptsize (±0.042) & \underline{0.029 \scriptsize (±0.026)} & \textbf{0.022 \scriptsize (±0.005)} \\
BEPRO     & 0.081 \scriptsize (±0.067) & \underline{0.034 \scriptsize (±0.046)} & \textbf{0.004 \scriptsize (±0.001)} \\
\midrule
\textbf{$1\textsuperscript{st}$-Place Count$\uparrow$}   & 0 & 2 & \textbf{18} \\
\textbf{Avg. Rank$\downarrow$} & 2.90 & 1.95 & \textbf{1.15} \\
\textbf{Avg. MAE$\downarrow$}  & 0.072 & 0.047 & \textbf{0.035} \\
\bottomrule
\end{tabular}}
\end{subtable}\hfill

\vspace{0.4cm}

\begin{subtable}[t]{0.49\textwidth}
\caption{\rgs{Regression}: Influential Node Count}
\resizebox{\linewidth}{!}{\begin{tabular}{lccc}
\toprule
\textbf{Dataset} & \methodname-8 & \methodname-16 & \methodname-32 \\
\midrule
MIR       & 0.080 \scriptsize (±0.059) & \underline{0.059 \scriptsize (±0.004)} & \textbf{0.039 \scriptsize (±0.005)} \\
DOGE2.0   & \textbf{0.065 \scriptsize (±0.034)} & \underline{0.094 \scriptsize (±0.003)} & 0.126 \scriptsize (±0.032) \\
MUTE      & 0.103 \scriptsize (±0.070) & \underline{0.078 \scriptsize (±0.034)} & \textbf{0.045 \scriptsize (±0.027)} \\
EVERMOON  & 0.107 \scriptsize (±0.067) & \underline{0.072 \scriptsize (±0.018)} & \textbf{0.038 \scriptsize (±0.024)} \\
DERC      & 0.063 \scriptsize (±0.065) & \underline{0.056 \scriptsize (±0.012)} & \textbf{0.033 \scriptsize (±0.017)} \\
ADX       & 0.084 \scriptsize (±0.070) & \underline{0.054 \scriptsize (±0.009)} & \textbf{0.030 \scriptsize (±0.001)} \\
HOICHI    & 0.115 \scriptsize (±0.066) & \underline{0.088 \scriptsize (±0.045)} & \textbf{0.055 \scriptsize (±0.028)} \\
SDEX      & 0.087 \scriptsize (±0.027) & \underline{0.083 \scriptsize (±0.068)} & \textbf{0.042 \scriptsize (±0.019)} \\
BAG       & 0.141 \scriptsize (±0.064) & \underline{0.109 \scriptsize (±0.045)} & \textbf{0.075 \scriptsize (±0.027)} \\
XCN       & \textbf{0.071 \scriptsize (±0.033)} & \underline{0.096 \scriptsize (±0.008)} & 0.126 \scriptsize (±0.033) \\
ETH2x-FLI & 0.066 \scriptsize (±0.059) & \underline{0.052 \scriptsize (±0.007)} & \textbf{0.033 \scriptsize (±0.020)} \\
stkAAVE   & 0.067 \scriptsize (±0.049) & \underline{0.063 \scriptsize (±0.008)} & \textbf{0.042 \scriptsize (±0.013)} \\
GLM       & 0.108 \scriptsize (±0.045) & \underline{0.088 \scriptsize (±0.014)} & \textbf{0.075 \scriptsize (±0.017)} \\
QOM       & 0.066 \scriptsize (±0.055) & \underline{0.055 \scriptsize (±0.004)} & \textbf{0.039 \scriptsize (±0.020)} \\
WOJAK     & 0.116 \scriptsize (±0.068) & \underline{0.074 \scriptsize (±0.006)} & \textbf{0.036 \scriptsize (±0.027)} \\
DINO      & 0.085 \scriptsize (±0.066) & \underline{0.059 \scriptsize (±0.009)} & \textbf{0.034 \scriptsize (±0.002)} \\
Metis     & 0.082 \scriptsize (±0.056) & \underline{0.072 \scriptsize (±0.022)} & \textbf{0.046 \scriptsize (±0.008)} \\
REPv2     & \textbf{0.127 \scriptsize (±0.015)} & 0.130 \scriptsize (±0.004) & \underline{0.129 \scriptsize (±0.022)} \\
TRAC      & 0.069 \scriptsize (±0.053) & \underline{0.059 \scriptsize (±0.003)} & \textbf{0.043 \scriptsize (±0.023)} \\
BEPRO     & 0.105 \scriptsize (±0.068) & \underline{0.064 \scriptsize (±0.005)} & \textbf{0.033 \scriptsize (±0.018)} \\
\midrule
\textbf{$1\textsuperscript{st}$-Place Count$\uparrow$} & 3 & 0 & \textbf{17} \\
\textbf{Avg. Rank$\downarrow$} & 2.70 & 2.05 & \textbf{1.25} \\
\textbf{Avg. MAE$\downarrow$}  & 0.090 & 0.075 & \textbf{0.056} \\
\bottomrule
\end{tabular}}
\end{subtable}\hfill
\begin{subtable}[t]{0.49\textwidth}
\caption{\rgs{Regression}: Edge Count}
\resizebox{\linewidth}{!}{\begin{tabular}{lccc}
\toprule
\textbf{Dataset} & \methodname-8 & \methodname-16 & \methodname-32 \\
\midrule
MIR       & 0.087 \scriptsize (±0.057) & \underline{0.035 \scriptsize (±0.027)} & \textbf{0.016 \scriptsize (±0.005)} \\
DOGE2.0   & \textbf{0.115 \scriptsize (±0.023)} & \underline{0.135 \scriptsize (±0.032)} & 0.187 \scriptsize (±0.008) \\
MUTE      & 0.092 \scriptsize (±0.036) & \underline{0.048 \scriptsize (±0.017)} & \textbf{0.021 \scriptsize (±0.022)} \\
EVERMOON  & 0.085 \scriptsize (±0.053) & \underline{0.041 \scriptsize (±0.029)} & \textbf{0.017 \scriptsize (±0.000)} \\
DERC      & 0.082 \scriptsize (±0.049) & \underline{0.033 \scriptsize (±0.024)} & \textbf{0.021 \scriptsize (±0.003)} \\
ADX       & 0.082 \scriptsize (±0.047) & \underline{0.029 \scriptsize (±0.018)} & \textbf{0.025 \scriptsize (±0.008)} \\
HOICHI    & 0.090 \scriptsize (±0.038) & \underline{0.062 \scriptsize (±0.006)} & \textbf{0.027 \scriptsize (±0.020)} \\
SDEX      & \underline{0.080 \scriptsize (±0.021)} & \textbf{0.077 \scriptsize (±0.018)} & 0.095 \scriptsize (±0.043) \\
BAG       & 0.099 \scriptsize (±0.023) & \underline{0.065 \scriptsize (±0.005)} & \textbf{0.041 \scriptsize (±0.020)} \\
XCN       & 0.066 \scriptsize (±0.056) & \textbf{0.027 \scriptsize (±0.009)} & \underline{0.062 \scriptsize (±0.015)} \\
ETH2x-FLI & 0.084 \scriptsize (±0.034) & \underline{0.035 \scriptsize (±0.019)} & \textbf{0.023 \scriptsize (±0.001)} \\
stkAAVE   & 0.065 \scriptsize (±0.037) & \textbf{0.025 \scriptsize (±0.004)} & \underline{0.051 \scriptsize (±0.022)} \\
GLM       & 0.106 \scriptsize (±0.013) & \underline{0.101 \scriptsize (±0.007)} & \textbf{0.087 \scriptsize (±0.008)} \\
QOM       & 0.074 \scriptsize (±0.054) & \textbf{0.032 \scriptsize (±0.019)} & \underline{0.036 \scriptsize (±0.013)} \\
WOJAK     & 0.090 \scriptsize (±0.059) & \underline{0.034 \scriptsize (±0.035)} & \textbf{0.010 \scriptsize (±0.004)} \\
DINO      & 0.085 \scriptsize (±0.053) & \underline{0.040 \scriptsize (±0.028)} & \textbf{0.017 \scriptsize (±0.007)} \\
Metis     & 0.087 \scriptsize (±0.030) & \underline{0.041 \scriptsize (±0.005)} & \textbf{0.034 \scriptsize (±0.010)} \\
REPv2     & 0.127 \scriptsize (±0.024) & \textbf{0.105 \scriptsize (±0.003)} & \underline{0.111 \scriptsize (±0.008)} \\
TRAC      & 0.086 \scriptsize (±0.042) & \underline{0.029 \scriptsize (±0.020)} & \textbf{0.021 \scriptsize (±0.001)} \\
BEPRO     & 0.094 \scriptsize (±0.067) & \underline{0.037 \scriptsize (±0.036)} & \textbf{0.011 \scriptsize (±0.008)} \\
\midrule

\textbf{$1\textsuperscript{st}$-Place Count$\uparrow$} & 1 & 5 & \textbf{14} \\
\textbf{Avg. Rank$\downarrow$} & 2.85 & 1.75 & \textbf{1.40} \\
\textbf{Avg. MAE$\downarrow$}  & 0.089 & 0.052 & \textbf{0.046} \\
\bottomrule
\end{tabular}}
\end{subtable}

\end{table*}

 \section{Extended Network and Task Results}
\label{appendix:addl_results}

This section reports the full per-network results for all classification and regression tasks, using the same evaluation protocol as in the main text. These tables complement the summary figures by showing the complete distribution of baseline and \methodname performance across datasets.  

\noindent\textbf{Classification.} Extended \cls{classification} results are provided for \textit{Edge-G/S}, \textit{LLC-G/S}, and \textit{Node-G/S} in Table~\ref{table:C_edge_GS}, Table~\ref{tab:llc_full_table}, and Table~\ref{tab:nod_gs_full_table}, respectively.  
\begin{table*}[ht]
\centering
\caption{AUC results for the Edge Growth/Shrinkage prediction task (\cls{classification}). Best results are in \textbf{bold}, second best are \underline{underlined}.}
\label{table:C_edge_GS}
\resizebox{\textwidth}{!}{%
\begin{tabular}{l  | ccccccc | cc}
\toprule
 & \multicolumn{7}{c|}{\textbf{Single Model on Individual Networks}} & \multicolumn{2}{c}{\textbf{Transfer Models}} \\
 \midrule
\textbf{Dataset} & \textbf{HTGN} & \textbf{GC-LSTM} & \textbf{EvolveGCN} & \textbf{GraphPulse} & \textbf{ROLAND} & \textbf{TGCN} & \textbf{WinGNN} & \textbf{MiNT} & \methodname \textbf{(Ours)} \\
\midrule
MIR       & 0.750 \scriptsize$\pm$0.005 & 0.768 \scriptsize$\pm$0.026 & 0.745 \scriptsize$\pm$0.015 & 0.689 \scriptsize$\pm$0.097 & 0.228 \scriptsize$\pm$0.060 & 0.749 \scriptsize$\pm$0.026 & 0.742 \scriptsize$\pm$0.015 & \textbf{0.836 \scriptsize$\pm$0.016} & \underline{0.793 \scriptsize$\pm$0.002} \\
DOGE2.0   & \underline{0.590 \scriptsize$\pm$0.059} & 0.538 \scriptsize$\pm$0.000 & 0.551 \scriptsize$\pm$0.022 & 0.384 \scriptsize$\pm$0.180 & 0.513 \scriptsize$\pm$0.022 & 0.487 \scriptsize$\pm$0.044 & 0.577 \scriptsize$\pm$0.038 & 0.538 \scriptsize$\pm$0.038 & \textbf{0.897 \scriptsize$\pm$0.089} \\
MUTE      & 0.649 \scriptsize$\pm$0.015 & 0.593 \scriptsize$\pm$0.030 & 0.617 \scriptsize$\pm$0.010 & \textbf{0.779 \scriptsize$\pm$0.004} & 0.289 \scriptsize$\pm$0.042 & 0.557 \scriptsize$\pm$0.068 & 0.593 \scriptsize$\pm$0.054 & 0.673 \scriptsize$\pm$0.013 & \underline{0.701 \scriptsize$\pm$0.098} \\
EVERMOON  & 0.512 \scriptsize$\pm$0.023 & \underline{0.562 \scriptsize$\pm$0.179} & 0.451 \scriptsize$\pm$0.046 & 0.519 \scriptsize$\pm$0.130 & 0.349 \scriptsize$\pm$0.119 & 0.463 \scriptsize$\pm$0.149 & 0.525 \scriptsize$\pm$0.114 & 0.517 \scriptsize$\pm$0.039 & \textbf{0.818 \scriptsize$\pm$0.046} \\
DERC      & 0.683 \scriptsize$\pm$0.013 & 0.703 \scriptsize$\pm$0.022 & 0.669 \scriptsize$\pm$0.009 & 0.769 \scriptsize$\pm$0.040 & 0.405 \scriptsize$\pm$0.357 & 0.743 \scriptsize$\pm$0.077 & 0.674 \scriptsize$\pm$0.044 & \underline{0.798 \scriptsize$\pm$0.027} & \textbf{0.839 \scriptsize$\pm$0.008} \\
ADX       & \underline{0.769 \scriptsize$\pm$0.018} & 0.723 \scriptsize$\pm$0.002 & 0.718 \scriptsize$\pm$0.004 & \textbf{0.784 \scriptsize$\pm$0.002} & 0.761 \scriptsize$\pm$0.011 & 0.674 \scriptsize$\pm$0.034 & 0.733 \scriptsize$\pm$0.023 & 0.679 \scriptsize$\pm$0.024 & 0.722 \scriptsize$\pm$0.087 \\
HOICHI    & \underline{0.807 \scriptsize$\pm$0.047} & \textbf{0.857 \scriptsize$\pm$0.000} & 0.856 \scriptsize$\pm$0.001 & 0.714 \scriptsize$\pm$0.010 & 0.815 \scriptsize$\pm$0.036 & 0.836 \scriptsize$\pm$0.034 & 0.769 \scriptsize$\pm$0.101 & 0.765 \scriptsize$\pm$0.018 & 0.591 \scriptsize$\pm$0.103 \\
SDEX      & \textbf{0.762 \scriptsize$\pm$0.034} & 0.720 \scriptsize$\pm$0.002 & 0.733 \scriptsize$\pm$0.028 & 0.436 \scriptsize$\pm$0.030 & 0.483 \scriptsize$\pm$0.254 & \underline{0.759 \scriptsize$\pm$0.039} & 0.726 \scriptsize$\pm$0.000 & 0.614 \scriptsize$\pm$0.020 & 0.348 \scriptsize$\pm$0.066 \\
BAG       & 0.673 \scriptsize$\pm$0.227 & 0.196 \scriptsize$\pm$0.179 & 0.329 \scriptsize$\pm$0.040 & \underline{0.934 \scriptsize$\pm$0.020} & 0.418 \scriptsize$\pm$0.016 & 0.334 \scriptsize$\pm$0.171 & 0.485 \scriptsize$\pm$0.105 & 0.931 \scriptsize$\pm$0.028 & \textbf{0.969 \scriptsize$\pm$0.008} \\
XCN       & 0.668 \scriptsize$\pm$0.099 & 0.306 \scriptsize$\pm$0.092 & 0.512 \scriptsize$\pm$0.067 & 0.821 \scriptsize$\pm$0.004 & 0.765 \scriptsize$\pm$0.015 & 0.703 \scriptsize$\pm$0.037 & 0.586 \scriptsize$\pm$0.029 & \textbf{0.851 \scriptsize$\pm$0.043} & \underline{0.844 \scriptsize$\pm$0.012} \\
ETH2x-FLI & 0.610 \scriptsize$\pm$0.059 & 0.670 \scriptsize$\pm$0.009 & 0.688 \scriptsize$\pm$0.010 & 0.666 \scriptsize$\pm$0.047 & 0.621 \scriptsize$\pm$0.023 & 0.647 \scriptsize$\pm$0.020 & 0.617 \scriptsize$\pm$0.056 & \textbf{0.729 \scriptsize$\pm$0.015} & \underline{0.712 \scriptsize$\pm$0.045} \\
stkAAVE   & 0.702 \scriptsize$\pm$0.042 & 0.368 \scriptsize$\pm$0.011 & 0.397 \scriptsize$\pm$0.022 & \textbf{0.743 \scriptsize$\pm$0.006} & 0.591 \scriptsize$\pm$0.122 & 0.577 \scriptsize$\pm$0.129 & 0.572 \scriptsize$\pm$0.018 & 0.709 \scriptsize$\pm$0.022 & \underline{0.732 \scriptsize$\pm$0.011} \\
GLM       & 0.830 \scriptsize$\pm$0.029 & 0.451 \scriptsize$\pm$0.003 & 0.501 \scriptsize$\pm$0.033 & 0.769 \scriptsize$\pm$0.018 & 0.559 \scriptsize$\pm$0.357 & 0.531 \scriptsize$\pm$0.008 & 0.530 \scriptsize$\pm$0.004 & \underline{0.831 \scriptsize$\pm$0.024} & \textbf{0.850 \scriptsize$\pm$0.009} \\
QOM       & 0.633 \scriptsize$\pm$0.017 & 0.612 \scriptsize$\pm$0.001 & 0.618 \scriptsize$\pm$0.002 & \textbf{0.775 \scriptsize$\pm$0.011} & 0.641 \scriptsize$\pm$0.003 & 0.647 \scriptsize$\pm$0.032 & 0.645 \scriptsize$\pm$0.099 & 0.647 \scriptsize$\pm$0.019 & \underline{0.745 \scriptsize$\pm$0.003} \\
WOJAK     & 0.479 \scriptsize$\pm$0.005 & 0.484 \scriptsize$\pm$0.000 & 0.505 \scriptsize$\pm$0.023 & 0.467 \scriptsize$\pm$0.030 & \underline{0.529 \scriptsize$\pm$0.005} & 0.516 \scriptsize$\pm$0.021 & 0.511 \scriptsize$\pm$0.026 & 0.524 \scriptsize$\pm$0.027 & \textbf{0.585 \scriptsize$\pm$0.067} \\
DINO      & 0.730 \scriptsize$\pm$0.195 & \underline{0.874 \scriptsize$\pm$0.028} & 0.868 \scriptsize$\pm$0.029 & 0.801 \scriptsize$\pm$0.020 & 0.497 \scriptsize$\pm$0.092 & 0.544 \scriptsize$\pm$0.314 & 0.628 \scriptsize$\pm$0.251 & 0.779 \scriptsize$\pm$0.113 & \textbf{0.895 \scriptsize$\pm$0.003} \\
Metis     & 0.715 \scriptsize$\pm$0.122 & 0.646 \scriptsize$\pm$0.023 & 0.688 \scriptsize$\pm$0.027 & \textbf{0.812 \scriptsize$\pm$0.011} & 0.696 \scriptsize$\pm$0.108 & 0.709 \scriptsize$\pm$0.033 & 0.690 \scriptsize$\pm$0.039 & \underline{0.760 \scriptsize$\pm$0.025} & 0.733 \scriptsize$\pm$0.004 \\
REPv2     & 0.760 \scriptsize$\pm$0.012 & 0.725 \scriptsize$\pm$0.014 & 0.709 \scriptsize$\pm$0.002 & \textbf{0.830 \scriptsize$\pm$0.001} & 0.751 \scriptsize$\pm$0.003 & 0.696 \scriptsize$\pm$0.035 & 0.744 \scriptsize$\pm$0.026 & \underline{0.789 \scriptsize$\pm$0.020} & 0.772 \scriptsize$\pm$0.016 \\
TRAC      & 0.712 \scriptsize$\pm$0.071 & 0.748 \scriptsize$\pm$0.000 & 0.748 \scriptsize$\pm$0.000 & \underline{0.767 \scriptsize$\pm$0.001} & 0.495 \scriptsize$\pm$0.223 & 0.741 \scriptsize$\pm$0.012 & 0.752 \scriptsize$\pm$0.007 & \textbf{0.785 \scriptsize$\pm$0.008} & 0.722 \scriptsize$\pm$0.001 \\
BEPRO     & 0.655 \scriptsize$\pm$0.038 & 0.632 \scriptsize$\pm$0.019 & 0.610 \scriptsize$\pm$0.012 & \underline{0.783 \scriptsize$\pm$0.003} & 0.439 \scriptsize$\pm$0.125 & 0.744 \scriptsize$\pm$0.074 & 0.736 \scriptsize$\pm$0.018 & 0.782 \scriptsize$\pm$0.003 & \textbf{0.800 \scriptsize$\pm$0.002} \\
\midrule
\textbf{$1\textsuperscript{st}$-Place Count$\uparrow$}   & 1 & 1 & 0 & 6 & 0 & 0 & 0 & 4 & \textbf{8} \\
\textbf{Avg. Rank $\downarrow$} & 4.85 & 5.80 & 6.10 & 3.80 & 6.30 & 5.60 & 5.55 & 3.30 & \textbf{2.80} \\
\textbf{Avg. AUC $\uparrow$}  & 0.684 & 0.609 & 0.626 & 0.712 & 0.542 & 0.633 & 0.642 & 0.727 & \textbf{0.753} \\

\bottomrule
\end{tabular}}
\end{table*}

\begin{table*}[ht]
\centering
\caption{AUC results for the LCC Growth/Shrinkage prediction task (\cls{classification}). Best results are in \textbf{bold}, second best are \underline{underlined}.}
\label{tab:llc_full_table}
\resizebox{\textwidth}{!}{%
\begin{tabular}{lccccccc|cc}
\toprule
 & \multicolumn{7}{c|}{\textbf{Single Model on Individual Networks}} & \multicolumn{2}{c}{\textbf{Transfer Models}} \\
 \midrule
\textbf{Dataset} & \textbf{HTGN} & \textbf{GC-LSTM} & \textbf{EvolveGCN} & \textbf{GraphPulse} & \textbf{ROLAND} & \textbf{TGCN} & \textbf{WinGNN} & \textbf{MiNT} & \methodname \textbf{(Ours)} \\
\midrule
MIR       & 0.745 \scriptsize (±0.023) & 0.585 \scriptsize (±0.128) & 0.575 \scriptsize (±0.146) & 0.800 \scriptsize (±0.008) & 0.536 \scriptsize (±0.275) & 0.585 \scriptsize (±0.055) & 0.749 \scriptsize (±0.020) & \textbf{0.845 \scriptsize (±0.035)} & \underline{0.815 \scriptsize (±0.007)} \\
DOGE2.0   & 0.446 \scriptsize (±0.164) & 0.387 \scriptsize (±0.294) & 0.583 \scriptsize (±0.115) & 0.333 \scriptsize (±0.042) & 0.411 \scriptsize (±0.232) & 0.464 \scriptsize (±0.182) & 0.595 \scriptsize (±0.176) & \underline{0.661 \scriptsize (±0.047)} & \textbf{0.702 \scriptsize (±0.160)} \\
MUTE      & 0.574 \scriptsize (±0.022) & 0.579 \scriptsize (±0.022) & 0.578 \scriptsize (±0.033) & \underline{0.647 \scriptsize (±0.014)} & 0.624 \scriptsize (±0.037) & 0.567 \scriptsize (±0.007) & 0.641 \scriptsize (±0.061) & 0.582 \scriptsize (±0.078) & \textbf{0.704 \scriptsize (±0.085)} \\
EVERMOON  & 0.494 \scriptsize (±0.127) & 0.512 \scriptsize (±0.112) & 0.548 \scriptsize (±0.152) & 0.463 \scriptsize (±0.034) & 0.491 \scriptsize (±0.157) & \underline{0.624 \scriptsize (±0.004)} & 0.603 \scriptsize (±0.041) & 0.527 \scriptsize (±0.118) & \textbf{0.667 \scriptsize (±0.045)} \\
DERC      & 0.717 \scriptsize (±0.035) & 0.591 \scriptsize (±0.010) & 0.553 \scriptsize (±0.044) & \underline{0.727 \scriptsize (±0.009)} & 0.481 \scriptsize (±0.131) & 0.523 \scriptsize (±0.103) & 0.582 \scriptsize (±0.043) & 0.689 \scriptsize (±0.096) & \textbf{0.808 \scriptsize (±0.022)} \\
ADX       & \textbf{0.753 \scriptsize (±0.013)} & 0.599 \scriptsize (±0.012) & 0.604 \scriptsize (±0.030) & 0.661 \scriptsize (±0.006) & 0.606 \scriptsize (±0.059) & 0.621 \scriptsize (±0.017) & 0.611 \scriptsize (±0.062) & 0.587 \scriptsize (±0.014) & \underline{0.727 \scriptsize (±0.093)} \\
HOICHI    & 0.746 \scriptsize (±0.010) & \underline{0.749 \scriptsize (±0.001)} & 0.745 \scriptsize (±0.003) & 0.730 \scriptsize (±0.017) & 0.360 \scriptsize (±0.121) & \textbf{0.750 \scriptsize (±0.002)} & 0.635 \scriptsize (±0.183) & 0.722 \scriptsize (±0.034) & 0.638 \scriptsize (±0.075) \\
SDEX      & \textbf{0.911 \scriptsize (±0.104)} & 0.721 \scriptsize (±0.138) & 0.601 \scriptsize (±0.105) & 0.808 \scriptsize (±0.050) & \underline{0.825 \scriptsize (±0.047)} & 0.770 \scriptsize (±0.231) & 0.575 \scriptsize (±0.282) & 0.382 \scriptsize (±0.280) & 0.816 \scriptsize (±0.025) \\
BAG       & 0.493 \scriptsize (±0.043) & 0.291 \scriptsize (±0.180) & 0.480 \scriptsize (±0.052) & \underline{0.900 \scriptsize (±0.010)} & 0.463 \scriptsize (±0.019) & 0.463 \scriptsize (±0.141) & 0.490 \scriptsize (±0.080) & 0.893 \scriptsize (±0.074) & \textbf{0.976 \scriptsize (±0.006)} \\
XCN       & 0.566 \scriptsize (±0.199) & 0.481 \scriptsize (±0.160) & 0.533 \scriptsize (±0.257) & 0.681 \scriptsize (±0.005) & 0.569 \scriptsize (±0.204) & 0.638 \scriptsize (±0.045) & 0.549 \scriptsize (±0.133) & \underline{0.827 \scriptsize (±0.025)} & \textbf{0.887 \scriptsize (±0.014)} \\
ETH2x-FLI & 0.561 \scriptsize (±0.037) & 0.529 \scriptsize (±0.017) & 0.547 \scriptsize (±0.009) & \underline{0.653 \scriptsize (±0.047)} & 0.499 \scriptsize (±0.135) & 0.549 \scriptsize (±0.019) & 0.505 \scriptsize (±0.090) & 0.618 \scriptsize (±0.025) & \textbf{0.687 \scriptsize (±0.032)} \\
stkAAVE   & 0.623 \scriptsize (±0.077) & 0.581 \scriptsize (±0.085) & 0.551 \scriptsize (±0.102) & 0.662 \scriptsize (±0.004) & 0.532 \scriptsize (±0.140) & 0.543 \scriptsize (±0.102) & 0.489 \scriptsize (±0.105) & \underline{0.688 \scriptsize (±0.019)} & \textbf{0.748 \scriptsize (±0.012)} \\
GLM       & 0.761 \scriptsize (±0.031) & 0.481 \scriptsize (±0.073) & 0.636 \scriptsize (±0.123) & 0.749 \scriptsize (±0.014) & 0.802 \scriptsize (±0.037) & 0.425 \scriptsize (±0.005) & 0.489 \scriptsize (±0.079) & \underline{0.818 \scriptsize (±0.074)} & \textbf{0.848 \scriptsize (±0.012)} \\
QOM       & 0.658 \scriptsize (±0.150) & 0.509 \scriptsize (±0.100) & 0.562 \scriptsize (±0.022) & \textbf{0.747 \scriptsize (±0.006)} & 0.627 \scriptsize (±0.134) & 0.419 \scriptsize (±0.044) & 0.546 \scriptsize (±0.152) & 0.645 \scriptsize (±0.109) & \underline{0.729 \scriptsize (±0.010)} \\
WOJAK     & 0.378 \scriptsize (±0.028) & 0.489 \scriptsize (±0.133) & 0.394 \scriptsize (±0.079) & \textbf{0.550 \scriptsize (±0.036)} & 0.360 \scriptsize (±0.005) & 0.481 \scriptsize (±0.092) & 0.415 \scriptsize (±0.017) & 0.492 \scriptsize (±0.107) & \underline{0.500 \scriptsize (±0.083)} \\
DINO      & 0.706 \scriptsize (±0.120) & \underline{0.796 \scriptsize (±0.023)} & 0.710 \scriptsize (±0.034) & 0.661 \scriptsize (±0.026) & 0.523 \scriptsize (±0.238) & 0.773 \scriptsize (±0.043) & 0.731 \scriptsize (±0.037) & 0.561 \scriptsize (±0.006) & \textbf{0.818 \scriptsize (±0.049)} \\
Metis     & 0.679 \scriptsize (±0.039) & 0.687 \scriptsize (±0.018) & 0.672 \scriptsize (±0.016) & \textbf{0.783 \scriptsize (±0.007)} & 0.672 \scriptsize (±0.103) & 0.657 \scriptsize (±0.014) & 0.634 \scriptsize (±0.042) & \underline{0.780 \scriptsize (±0.041)} & 0.731 \scriptsize (±0.001) \\
REPv2     & 0.730 \scriptsize (±0.007) & 0.653 \scriptsize (±0.015) & 0.644 \scriptsize (±0.027) & \underline{0.752 \scriptsize (±0.001)} & 0.658 \scriptsize (±0.103) & 0.646 \scriptsize (±0.025) & 0.683 \scriptsize (±0.014) & 0.742 \scriptsize (±0.041) & \textbf{0.764 \scriptsize (±0.020)} \\
TRAC      & 0.733 \scriptsize (±0.009) & 0.629 \scriptsize (±0.005) & 0.623 \scriptsize (±0.004) & 0.686 \scriptsize (±0.001) & 0.606 \scriptsize (±0.117) & 0.620 \scriptsize (±0.005) & 0.599 \scriptsize (±0.026) & \underline{0.762 \scriptsize (±0.028)} & \textbf{0.781 \scriptsize (±0.004)} \\
BEPRO     & 0.694 \scriptsize (±0.009) & 0.595 \scriptsize (±0.008) & 0.557 \scriptsize (±0.058) & \underline{0.725 \scriptsize (±0.004)} & 0.482 \scriptsize (±0.146) & 0.536 \scriptsize (±0.031) & 0.582 \scriptsize (±0.063) & 0.628 \scriptsize (±0.017) & \textbf{0.830 \scriptsize (±0.006)} \\
\midrule
\textbf{$1\textsuperscript{st}$-Place Count$\uparrow$}   & 2 & 0 & 0 & 3 & 0 & 1 & 0 & 1 & \textbf{13} \\
\textbf{Avg. Rank $\downarrow$} & 4.40 & 6.00 & 6.30 & 3.50 & 6.95 & 6.10 & 6.10 & 3.95 & \textbf{1.70} \\
\textbf{Avg. AUC $\uparrow$}  & 0.648 & 0.572 & 0.585 & 0.686 & 0.556 & 0.583 & 0.585 & 0.672 & \textbf{0.759} \\
\bottomrule
\end{tabular}}
\end{table*}

\begin{table*}[ht]
\centering
\caption{AUC results for the Node Growth/Shrinkage prediction task (\cls{classification}). Best results are in \textbf{bold}, second best are \underline{underlined}.}
\label{tab:nod_gs_full_table}
\resizebox{\textwidth}{!}{%
\begin{tabular}{lccccccc|cc}
\toprule
 & \multicolumn{7}{c|}{\textbf{Single Model on Individual Networks}} & \multicolumn{2}{c}{\textbf{Transfer Models}} \\
\toprule
\textbf{Dataset} & \textbf{HTGN} & \textbf{GC-LSTM} & \textbf{EvolveGCN} & \textbf{GraphPulse} & \textbf{ROLAND} & \textbf{TGCN} & \textbf{WinGNN} & \textbf{MiNT} & \methodname \textbf{(Ours)} \\
\midrule
MIR       & 0.545 \scriptsize (±0.030) & 0.537 \scriptsize (±0.033) & 0.528 \scriptsize (±0.081) & \underline{0.633 \scriptsize (±0.066)} & 0.472 \scriptsize (±0.040) & 0.532 \scriptsize (±0.021) & 0.525 \scriptsize (±0.043) & 0.622 \scriptsize (±0.030) & \textbf{0.764 \scriptsize (±0.001)} \\
DOGE2.0   & 0.427 \scriptsize (±0.065) & 0.633 \scriptsize (±0.014) & 0.400 \scriptsize (±0.061) & 0.403 \scriptsize (±0.035) & 0.260 \scriptsize (±0.000) & 0.627 \scriptsize (±0.034) & \underline{0.693 \scriptsize (±0.041)} & \textbf{0.750 \scriptsize (±0.014)} & 0.633 \scriptsize (±0.064) \\
MUTE      & 0.518 \scriptsize (±0.023) & 0.448 \scriptsize (±0.008) & 0.475 \scriptsize (±0.051) & \underline{0.677 \scriptsize (±0.008)} & 0.411 \scriptsize (±0.053) & 0.458 \scriptsize (±0.009) & 0.562 \scriptsize (±0.015) & 0.606 \scriptsize (±0.056) & \textbf{0.748 \scriptsize (±0.025)} \\
EVERMOON  & 0.585 \scriptsize (±0.059) & 0.606 \scriptsize (±0.021) & 0.488 \scriptsize (±0.088) & 0.463 \scriptsize (±0.002) & 0.427 \scriptsize (±0.137) & 0.567 \scriptsize (±0.030) & 0.548 \scriptsize (±0.116) & \underline{0.614 \scriptsize (±0.071)} & \textbf{0.655 \scriptsize (±0.040)} \\
DERC      & \underline{0.662 \scriptsize (±0.051)} & 0.492 \scriptsize (±0.069) & 0.503 \scriptsize (±0.084) & 0.611 \scriptsize (±0.049) & 0.551 \scriptsize (±0.013) & 0.447 \scriptsize (±0.003) & 0.517 \scriptsize (±0.034) & 0.569 \scriptsize (±0.004) & \textbf{0.742 \scriptsize (±0.004)} \\
ADX       & \underline{0.678 \scriptsize (±0.017)} & 0.505 \scriptsize (±0.043) & 0.509 \scriptsize (±0.022) & 0.701 \scriptsize (±0.003) & 0.557 \scriptsize (±0.082) & 0.484 \scriptsize (±0.048) & 0.504 \scriptsize (±0.018) & 0.507 \scriptsize (±0.037) & \textbf{0.718 \scriptsize (±0.051)} \\
HOICHI    & 0.687 \scriptsize (±0.004) & \underline{0.718 \scriptsize (±0.007)} & 0.685 \scriptsize (±0.020) & \textbf{0.745 \scriptsize (±0.006)} & 0.347 \scriptsize (±0.084) & \underline{0.718 \scriptsize (±0.002)} & 0.526 \scriptsize (±0.188) & 0.492 \scriptsize (±0.120) & 0.558 \scriptsize (±0.047) \\
SDEX      & \underline{0.824 \scriptsize (±0.106)} & 0.364 \scriptsize (±0.148) & 0.817 \scriptsize (±0.032) & \textbf{0.865 \scriptsize (±0.011)} & 0.779 \scriptsize (±0.018) & 0.755 \scriptsize (±0.202) & 0.757 \scriptsize (±0.072) & 0.861 \scriptsize (±0.025) & 0.743 \scriptsize (±0.026) \\
BAG       & 0.735 \scriptsize (±0.075) & 0.337 \scriptsize (±0.089) & 0.166 \scriptsize (±0.066) & \underline{0.897 \scriptsize (±0.016)} & 0.390 \scriptsize (±0.088) & 0.391 \scriptsize (±0.219) & 0.515 \scriptsize (±0.008) & 0.685 \scriptsize (±0.038) & \textbf{0.969 \scriptsize (±0.009)} \\
XCN       & 0.476 \scriptsize (±0.012) & 0.466 \scriptsize (±0.012) & 0.407 \scriptsize (±0.176) & \underline{0.671 \scriptsize (±0.020)} & 0.430 \scriptsize (±0.144) & 0.483 \scriptsize (±0.036) & 0.355 \scriptsize (±0.017) & 0.505 \scriptsize (±0.002) & \textbf{0.878 \scriptsize (±0.009)} \\
ETH2x-FLI & \underline{0.628 \scriptsize (±0.022)} & 0.548 \scriptsize (±0.001) & 0.548 \scriptsize (±0.002) & 0.615 \scriptsize (±0.020) & 0.488 \scriptsize (±0.063) & 0.553 \scriptsize (±0.036) & 0.586 \scriptsize (±0.098) & 0.411 \scriptsize (±0.066) & \textbf{0.678 \scriptsize (±0.031)} \\
stkAAVE   & 0.517 \scriptsize (±0.093) & 0.543 \scriptsize (±0.043) & 0.456 \scriptsize (±0.069) & 0.643 \scriptsize (±0.005) & \underline{0.661 \scriptsize (±0.037)} & 0.425 \scriptsize (±0.029) & 0.465 \scriptsize (±0.036) & 0.561 \scriptsize (±0.007) & \textbf{0.779 \scriptsize (±0.007)} \\
GLM       & 0.706 \scriptsize (±0.014) & 0.566 \scriptsize (±0.001) & 0.516 \scriptsize (±0.105) & 0.595 \scriptsize (±0.003) & 0.493 \scriptsize (±0.149) & 0.575 \scriptsize (±0.019) & 0.610 \scriptsize (±0.024) & \underline{0.720 \scriptsize (±0.045)} & \textbf{0.763 \scriptsize (±0.005)} \\
QOM       & 0.647 \scriptsize (±0.094) & 0.492 \scriptsize (±0.003) & 0.485 \scriptsize (±0.001) & \underline{0.705 \scriptsize (±0.002)} & 0.592 \scriptsize (±0.080) & 0.495 \scriptsize (±0.004) & 0.409 \scriptsize (±0.051) & 0.572 \scriptsize (±0.017) & \textbf{0.719 \scriptsize (±0.012)} \\
WOJAK     & 0.417 \scriptsize (±0.143) & 0.338 \scriptsize (±0.068) & 0.357 \scriptsize (±0.104) & \underline{0.500 \scriptsize (±0.000)} & 0.202 \scriptsize (±0.018) & 0.488 \scriptsize (±0.080) & 0.314 \scriptsize (±0.029) & \textbf{0.618 \scriptsize (±0.035)} & 0.412 \scriptsize (±0.060) \\
DINO      & \underline{0.845 \scriptsize (±0.015)} & 0.323 \scriptsize (±0.148) & 0.444 \scriptsize (±0.052) & 0.686 \scriptsize (±0.007) & 0.330 \scriptsize (±0.115) & 0.615 \scriptsize (±0.070) & 0.600 \scriptsize (±0.292) & 0.735 \scriptsize (±0.005) & \textbf{0.910 \scriptsize (±0.013)} \\
Metis     & 0.589 \scriptsize (±0.049) & 0.483 \scriptsize (±0.052) & 0.566 \scriptsize (±0.012) & \underline{0.652 \scriptsize (±0.029)} & 0.574 \scriptsize (±0.040) & 0.549 \scriptsize (±0.011) & 0.510 \scriptsize (±0.005) & 0.616 \scriptsize (±0.012) & \textbf{0.693 \scriptsize (±0.007)} \\
REPv2     & 0.650 \scriptsize (±0.004) & 0.519 \scriptsize (±0.023) & 0.515 \scriptsize (±0.019) & 0.662 \scriptsize (±0.008) & 0.597 \scriptsize (±0.028) & 0.534 \scriptsize (±0.029) & 0.626 \scriptsize (±0.058) & \textbf{0.710 \scriptsize (±0.129)} & \underline{0.689 \scriptsize (±0.013)} \\
TRAC      & \underline{0.670 \scriptsize (±0.031)} & 0.527 \scriptsize (±0.016) & 0.524 \scriptsize (±0.003) & 0.610 \scriptsize (±0.055) & 0.546 \scriptsize (±0.024) & 0.524 \scriptsize (±0.001) & 0.528 \scriptsize (±0.020) & 0.600 \scriptsize (±0.027) & \textbf{0.765 \scriptsize (±0.003)} \\
BEPRO     & 0.500 \scriptsize (±0.055) & 0.332 \scriptsize (±0.010) & 0.356 \scriptsize (±0.015) & \underline{0.707 \scriptsize (±0.005)} & 0.490 \scriptsize (±0.016) & 0.372 \scriptsize (±0.100) & 0.420 \scriptsize (±0.018) & 0.561 \scriptsize (±0.022) & \textbf{0.865 \scriptsize (±0.010)} \\
\midrule
\textbf{$1\textsuperscript{st}$-Place Count$\uparrow$}   & 0 & 0 & 0 & 2 & 0 & 0 & 0 & 3 & \textbf{15} \\
\textbf{Avg. Rank$\downarrow$} & 3.65 & 6.65 & 7.00 & 2.95 & 6.70 & 6.15 & 6.20 & 3.55 & \textbf{1.95} \\
\textbf{Avg. AUC$\uparrow$}  & 0.615 & 0.489 & 0.487 & 0.652 & 0.480 & 0.530 & 0.528 & 0.616 & \textbf{0.734} \\
\bottomrule
\end{tabular}}
\end{table*}

\noindent\textbf{Regression.} Extended \rgs{regression} results are provided for \textit{Edge Count}, \textit{New Node Count}, and \textit{Influential Node Count} in Table~\ref{tab:edg_count_r}, Table~\ref{tab:new_node_reg_mae_no_h64}, and Table~\ref{tab:inf_node_mae_no_h64}, respectively.  
\begin{table*}[ht]
\centering
\caption{MAE results for the Edge Count prediction task (\rgs{regression}). Best results are in \textbf{bold}, second best are \underline{underlined}.}
\label{tab:edg_count_r}
\resizebox{\textwidth}{!}{%
\begin{tabular}{l | *{7}{c} | c }
 \toprule 
    &   \multicolumn{7}{c|}{\textbf{Single Model on Individual Networks}}  &  \multicolumn{1}{c}{\textbf{Transfer Models}} \\
\toprule
\textbf{Dataset} & \textbf{HTGN} & \textbf{TGCN} & \textbf{GCLSTM} & \textbf{ROLAND} & \textbf{EGCN} & \textbf{GraphPulse} & \textbf{WinGNN} & \methodname \textbf{(Ours)} \\
\midrule
MIR       & 0.059 {\scriptsize (±0.007)} & 0.044 {\scriptsize (±0.009)} & 0.047 {\scriptsize (±0.003)} & \underline{0.039 {\scriptsize (±0.000)}} & 0.057 {\scriptsize (±0.015)} & 0.059 {\scriptsize (±0.001)} & 0.046 {\scriptsize (±0.010)} & \textbf{0.016 {\scriptsize (±0.005)}} \\
DOGE2.0   & 0.101 {\scriptsize (±0.035)} & 0.063 {\scriptsize (±0.017)} & 0.106 {\scriptsize (±0.029)} & 0.052 {\scriptsize (±0.003)} & 0.092 {\scriptsize (±0.031)} & \underline{0.046 {\scriptsize (±0.000)}} & \textbf{0.045 {\scriptsize (±0.003)}} & 0.187 {\scriptsize (±0.008)} \\
MUTE      & 0.025 {\scriptsize (±0.006)} & 0.038 {\scriptsize (±0.004)} & \textbf{0.017 {\scriptsize (±0.001)}} & 0.040 {\scriptsize (±0.007)} & 0.049 {\scriptsize (±0.005)} & 0.025 {\scriptsize (±0.002)} & 0.027 {\scriptsize (±0.003)} & \underline{0.021 {\scriptsize (±0.022)}} \\
EVERMOON  & \textbf{0.010 {\scriptsize (±0.001)}} & 0.021 {\scriptsize (±0.013)} & 0.025 {\scriptsize (±0.007)} & \underline{0.016 {\scriptsize (±0.016)}} & 0.030 {\scriptsize (±0.010)} & 0.235 {\scriptsize (±0.005)} & 0.025 {\scriptsize (±0.004)} & 0.017 {\scriptsize (±0.000)} \\
DERC      & 0.038 {\scriptsize (±0.015)} & 0.059 {\scriptsize (±0.011)} & \textbf{0.016 {\scriptsize (±0.005)}} & 0.060 {\scriptsize (±0.008)} & \underline{0.023 {\scriptsize (±0.007)}} & \underline{0.023 {\scriptsize (±0.003)}} & 0.032 {\scriptsize (±0.001)} & 0.021 {\scriptsize (±0.003)} \\
ADX       & \underline{0.017 {\scriptsize (±0.001)}} & 0.018 {\scriptsize (±0.003)} & \textbf{0.016 {\scriptsize (±0.001)}} & \underline{0.017 {\scriptsize (±0.002)}} & 0.021 {\scriptsize (±0.002)} & 0.019 {\scriptsize (±0.001)} & \textbf{0.016 {\scriptsize (±0.000)}} & 0.025 {\scriptsize (±0.008)} \\
HOICHI    & 0.034 {\scriptsize (±0.010)} & \textbf{0.020 {\scriptsize (±0.001)}} & 0.046 {\scriptsize (±0.013)} & \textbf{0.020 {\scriptsize (±0.003)}} & 0.034 {\scriptsize (±0.013)} & 0.044 {\scriptsize (±0.002)} & \underline{0.028 {\scriptsize (±0.005)}} & 0.027 {\scriptsize (±0.020)} \\
SDEX      & \underline{0.080 {\scriptsize (±0.029)}} & 0.128 {\scriptsize (±0.046)} & \textbf{0.058 {\scriptsize (±0.007)}} & 0.121 {\scriptsize (±0.002)} & 0.085 {\scriptsize (±0.025)} & 0.106 {\scriptsize (±0.005)} & 0.128 {\scriptsize (±0.008)} & 0.095 {\scriptsize (±0.043)} \\
BAG       & \textbf{0.022 {\scriptsize (±0.003)}} & \underline{0.023 {\scriptsize (±0.017)}} & 0.025 {\scriptsize (±0.002)} & 0.030 {\scriptsize (±0.016)} & 0.027 {\scriptsize (±0.005)} & 0.063 {\scriptsize (±0.001)} & 0.260 {\scriptsize (±0.064)} & 0.041 {\scriptsize (±0.020)} \\
XCN       & 0.074 {\scriptsize (±0.006)} & 0.112 {\scriptsize (±0.042)} & 0.107 {\scriptsize (±0.024)} & 0.120 {\scriptsize (±0.035)} & 0.121 {\scriptsize (±0.011)} & 0.118 {\scriptsize (±0.000)} & \underline{0.072 {\scriptsize (±0.018)}} & \textbf{0.062 {\scriptsize (±0.015)}} \\
ETH2x-FLI & 0.055 {\scriptsize (±0.015)} & 0.079 {\scriptsize (±0.019)} & 0.177 {\scriptsize (±0.057)} & 0.040 {\scriptsize (±0.026)} & 0.066 {\scriptsize (±0.016)} & 0.144 {\scriptsize (±0.009)} & \underline{0.030 {\scriptsize (±0.011)}} & \textbf{0.023 {\scriptsize (±0.001)}} \\
stkAAVE   & 0.083 {\scriptsize (±0.008)} & 0.087 {\scriptsize (±0.005)} & 0.092 {\scriptsize (±0.017)} & 0.104 {\scriptsize (±0.012)} & 0.079 {\scriptsize (±0.010)} & 0.096 {\scriptsize (±0.006)} & 0.100 {\scriptsize (±0.004)} & \textbf{0.051 {\scriptsize (±0.022)}} \\
GLM       & 0.072 {\scriptsize (±0.010)} & \underline{0.058 {\scriptsize (±0.005)}} & 0.063 {\scriptsize (±0.002)} & \underline{0.058 {\scriptsize (±0.003)}} & 0.060 {\scriptsize (±0.001)} & 0.076 {\scriptsize (±0.001)} & \textbf{0.054 {\scriptsize (±0.003)}} & 0.087 {\scriptsize (±0.008)} \\
QOM       & 0.042 {\scriptsize (±0.005)} & 0.085 {\scriptsize (±0.020)} & 0.053 {\scriptsize (±0.030)} & 0.057 {\scriptsize (±0.030)} & 0.069 {\scriptsize (±0.013)} & 0.055 {\scriptsize (±0.001)} & 0.046 {\scriptsize (±0.006)} & \textbf{0.036 {\scriptsize (±0.013)}} \\
WOJAK     & \underline{0.009 {\scriptsize (±0.002)}} & 0.012 {\scriptsize (±0.003)} & 0.013 {\scriptsize (±0.004)} & 0.016 {\scriptsize (±0.008)} & 0.013 {\scriptsize (±0.007)} & 0.057 {\scriptsize (±0.008)} & \textbf{0.006 {\scriptsize (±0.002)}} & 0.010 {\scriptsize (±0.004)} \\
DINO      & 0.069 {\scriptsize (±0.020)} & 0.025 {\scriptsize (±0.009)} & 0.040 {\scriptsize (±0.019)} & \textbf{0.014 {\scriptsize (±0.004)}} & 0.039 {\scriptsize (±0.011)} & 0.087 {\scriptsize (±0.002)} & 0.021 {\scriptsize (±0.008)} & \underline{0.017 {\scriptsize (±0.007)}} \\
Metis     & 0.038 {\scriptsize (±0.002)} & 0.054 {\scriptsize (±0.001)} & 0.047 {\scriptsize (±0.003)} & 0.057 {\scriptsize (±0.008)} & 0.053 {\scriptsize (±0.004)} & 0.066 {\scriptsize (±0.006)} & 0.043 {\scriptsize (±0.001)} & \textbf{0.034 {\scriptsize (±0.010)}} \\
REPv2     & 0.117 {\scriptsize (±0.013)} & \underline{0.108 {\scriptsize (±0.004)}} & 0.115 {\scriptsize (±0.004)} & \textbf{0.106 {\scriptsize (±0.001)}} & 0.128 {\scriptsize (±0.036)} & 0.119 {\scriptsize (±0.001)} & 0.118 {\scriptsize (±0.001)} & 0.111 {\scriptsize (±0.008)} \\
TRAC      & 0.026 {\scriptsize (±0.004)} & 0.036 {\scriptsize (±0.010)} & 0.061 {\scriptsize (±0.006)} & \underline{0.023 {\scriptsize (±0.004)}} & 0.036 {\scriptsize (±0.003)} & \textbf{0.017 {\scriptsize (±0.000)}} & 0.040 {\scriptsize (±0.014)} & 0.021 {\scriptsize (±0.001)} \\
BEPRO     & \underline{0.009 {\scriptsize (±0.001)}} & \underline{0.009 {\scriptsize (±0.003)}} & \underline{0.009 {\scriptsize (±0.002)}} & 0.015 {\scriptsize (±0.017)} & \textbf{0.007 {\scriptsize (±0.001)}} & \textbf{0.007 {\scriptsize (±0.000)}} & \textbf{0.007 {\scriptsize (±0.002)}} & 0.011 {\scriptsize (±0.008)} \\
\midrule
\textbf{$1\textsuperscript{st}$-Place Count$\uparrow$}     & 2 & 1 & 4 & 3 & 1 & 2 & 5 & \textbf{6} \\
\textbf{Avg. Rank  $\downarrow$ }  & 3.95 & 4.60 & 4.58 & 4.58 & 5.30 & 5.72 & 3.93 & \textbf{3.35} \\
\textbf{Avg. MAE  $\downarrow$ }   & 0.049 & 0.054 & 0.057 & 0.050 & 0.054 & 0.073 & 0.057 & \textbf{0.046} \\
\bottomrule
\end{tabular}}
\end{table*}

\begin{table*}[h]
\centering
\caption{MAE results for the New Node Count prediction task (\rgs{regression}). Best results are in \textbf{bold}, second best are \underline{underlined}.}
\label{tab:new_node_reg_mae_no_h64}
\resizebox{\textwidth}{!}{%
\begin{tabular}{l | *{7}{c} | c }
\toprule
 & \multicolumn{7}{c|}{\textbf{Single Model on Individual Networks}} & \multicolumn{1}{c}{\textbf{Transfer Models}} \\
\toprule
\textbf{Dataset} & \textbf{HTGN} & \textbf{TGCN} & \textbf{GCLSTM} & \textbf{ROLAND} & \textbf{EGCN} & \textbf{GraphPulse} & \textbf{WinGNN} & \methodname \textbf{(ours)} \\
\midrule
MIR & 0.031 {\scriptsize (±0.002)} & 0.028 {\scriptsize (±0.001)} & 0.039 {\scriptsize (±0.005)} & \underline{0.025 {\scriptsize (±0.018)}} & 0.030 {\scriptsize (±0.001)} & \underline{0.025 {\scriptsize (±0.001)}} & 0.037 {\scriptsize (±0.011)} & \textbf{0.013 {\scriptsize (±0.004)}} \\
DOGE2.0 & \underline{0.046 {\scriptsize (±0.009)}} & 0.064 {\scriptsize (±0.021)} & \textbf{0.044 {\scriptsize (±0.014)}} & 0.073 {\scriptsize (±0.019)} & 0.051 {\scriptsize (±0.031)} & 0.157 {\scriptsize (±0.010)} & 0.098 {\scriptsize (±0.009)} & 0.092 {\scriptsize (±0.008)} \\
MUTE & \textbf{0.021 {\scriptsize (±0.003)}} & 0.041 {\scriptsize (±0.005)} & 0.030 {\scriptsize (±0.003)} & 0.048 {\scriptsize (±0.002)} & 0.053 {\scriptsize (±0.003)} & 0.031 {\scriptsize (±0.001)} & 0.051 {\scriptsize (±0.015)} & \underline{0.025 {\scriptsize (±0.005)}} \\
EVERMOON & \textbf{0.010 {\scriptsize (±0.004)}} & 0.017 {\scriptsize (±0.008)} & 0.026 {\scriptsize (±0.009)} & 0.029 {\scriptsize (±0.019)} & 0.028 {\scriptsize (±0.017)} & 0.222 {\scriptsize (±0.005)} & 0.022 {\scriptsize (±0.002)} & \underline{0.012 {\scriptsize (±0.005)}} \\
DERC & 0.028 {\scriptsize (±0.009)} & 0.034 {\scriptsize (±0.018)} & \underline{0.016 {\scriptsize (±0.024)}} & 0.043 {\scriptsize (±0.024)} & 0.019 {\scriptsize (±0.004)} & 0.043 {\scriptsize (±0.011)} & 0.027 {\scriptsize (±0.003)} & \textbf{0.015 {\scriptsize (±0.004)}} \\
ADX & 0.014 {\scriptsize (±0.001)} & \textbf{0.010 {\scriptsize (±0.001)}} & \underline{0.011 {\scriptsize (±0.001)}} & 0.024 {\scriptsize (±0.296)} & 0.012 {\scriptsize (±0.000)} & 0.024 {\scriptsize (±0.002)} & \underline{0.011 {\scriptsize (±0.000)}} & 0.016 {\scriptsize (±0.005)} \\
HOICHI & 0.044 {\scriptsize (±0.004)} & 0.035 {\scriptsize (±0.025)} & 0.053 {\scriptsize (±0.028)} & 0.204 {\scriptsize (±0.028)} & 0.039 {\scriptsize (±0.023)} & 0.066 {\scriptsize (±0.001)} & \textbf{0.027 {\scriptsize (±0.004)}} & \underline{0.029 {\scriptsize (±0.008)}} \\
SDEX & 0.075 {\scriptsize (±0.002)} & 0.093 {\scriptsize (±0.006)} & \underline{0.069 {\scriptsize (±0.002)}} & 0.088 {\scriptsize (±0.002)} & 0.077 {\scriptsize (±0.008)} & 0.087 {\scriptsize (±0.002)} & 0.090 {\scriptsize (±0.025)} & \textbf{0.063 {\scriptsize (±0.006)}} \\
BAG & \underline{0.023 {\scriptsize (±0.007)}} & 0.031 {\scriptsize (±0.008)} & \textbf{0.019 {\scriptsize (±0.007)}} & 0.053 {\scriptsize (±0.006)} & 0.030 {\scriptsize (±0.004)} & 0.054 {\scriptsize (±0.000)} & 0.230 {\scriptsize (±0.062)} & 0.052 {\scriptsize (±0.002)} \\
XCN & 0.014 {\scriptsize (±0.007)} & 0.015 {\scriptsize (±0.006)} & 0.017 {\scriptsize (±0.007)} & \underline{0.012 {\scriptsize (±0.012)}} & 0.017 {\scriptsize (±0.007)} & 0.040 {\scriptsize (±0.001)} & 0.015 {\scriptsize (±0.002)} & \textbf{0.009 {\scriptsize (±0.002)}} \\
ETH2x-FLI & 0.031 {\scriptsize (±0.010)} & 0.041 {\scriptsize (±0.001)} & 0.069 {\scriptsize (±0.010)} & \textbf{0.020 {\scriptsize (±0.004)}} & 0.030 {\scriptsize (±0.002)} & 0.092 {\scriptsize (±0.001)} & \underline{0.028 {\scriptsize (±0.003)}} & 0.030 {\scriptsize (±0.004)} \\
stkAAVE & 0.128 {\scriptsize (±0.018)} & 0.136 {\scriptsize (±0.008)} & 0.128 {\scriptsize (±0.018)} & 0.154 {\scriptsize (±0.005)} & \underline{0.124 {\scriptsize (±0.005)}} & 0.151 {\scriptsize (±0.000)} & 0.147 {\scriptsize (±0.008)} & \textbf{0.078 {\scriptsize (±0.017)}} \\
GLM & \textbf{0.066 {\scriptsize (±0.000)}} & 0.068 {\scriptsize (±0.001)} & 0.068 {\scriptsize (±0.000)} & 0.068 {\scriptsize (±0.002)} & \underline{0.067 {\scriptsize (±0.002)}} & 0.092 {\scriptsize (±0.000)} & \textbf{0.066 {\scriptsize (±0.000)}} & 0.094 {\scriptsize (±0.007)} \\
QOM & 0.035 {\scriptsize (±0.006)} & 0.038 {\scriptsize (±0.010)} & 0.032 {\scriptsize (±0.020)} & \underline{0.018 {\scriptsize (±0.010)}} & 0.035 {\scriptsize (±0.020)} & 0.033 {\scriptsize (±0.004)} & 0.029 {\scriptsize (±0.007)} & \textbf{0.012 {\scriptsize (±0.001)}} \\
WOJAK & \underline{0.008 {\scriptsize (±0.001)}} & 0.009 {\scriptsize (±0.001)} & 0.015 {\scriptsize (±0.003)} & 0.029 {\scriptsize (±0.017)} & 0.014 {\scriptsize (±0.003)} & 0.067 {\scriptsize (±0.005)} & \textbf{0.007 {\scriptsize (±0.001)}} & 0.009 {\scriptsize (±0.000)} \\
DINO & 0.061 {\scriptsize (±0.017)} & 0.024 {\scriptsize (±0.002)} & 0.028 {\scriptsize (±0.019)} & \textbf{0.013 {\scriptsize (±0.005)}} & 0.030 {\scriptsize (±0.005)} & 0.085 {\scriptsize (±0.001)} & 0.051 {\scriptsize (±0.029)} & \underline{0.018 {\scriptsize (±0.007)}} \\
Metis & \textbf{0.034 {\scriptsize (±0.010)}} & 0.045 {\scriptsize (±0.006)} & \underline{0.041 {\scriptsize (±0.006)}} & 0.043 {\scriptsize (±0.005)} & 0.042 {\scriptsize (±0.004)} & 0.054 {\scriptsize (±0.001)} & \textbf{0.034 {\scriptsize (±0.002)}} & \textbf{0.034 {\scriptsize (±0.010)}} \\
REPv2 & \underline{0.061 {\scriptsize (±0.003)}} & \underline{0.061 {\scriptsize (±0.004)}} & 0.075 {\scriptsize (±0.003)} & \textbf{0.055 {\scriptsize (±0.002)}} & 0.063 {\scriptsize (±0.004)} & 0.068 {\scriptsize (±0.000)} & 0.063 {\scriptsize (±0.000)} & 0.066 {\scriptsize (±0.004)} \\
TRAC & 0.043 {\scriptsize (±0.021)} & 0.030 {\scriptsize (±0.003)} & 0.071 {\scriptsize (±0.021)} & \underline{0.021 {\scriptsize (±0.006)}} & 0.025 {\scriptsize (±0.009)} & \textbf{0.018 {\scriptsize (±0.000)}} & 0.044 {\scriptsize (±0.010)} & 0.022 {\scriptsize (±0.005)} \\
BEPRO & 0.012 {\scriptsize (±0.002)} & 0.010 {\scriptsize (±0.002)} & 0.011 {\scriptsize (±0.017)} & 0.011 {\scriptsize (±0.011)} & 0.009 {\scriptsize (±0.000)} & 0.012 {\scriptsize (±0.000)} & \underline{0.006 {\scriptsize (±0.001)}} & \textbf{0.004 {\scriptsize (±0.001)}} \\
\midrule
\textbf{$1\textsuperscript{st}$-Place Count$\uparrow$} & 4 & 1 & 2 & 3 & 0 & 1 & 4 & \textbf{8} \\
\textbf{Avg. Rank $\downarrow$} & 3.45 & 4.10 & 4.00 & 4.35 & 3.90 & 5.75 & 4.05 & \textbf{2.50} \\
\textbf{Avg. MAE $\downarrow$} & 0.039 & 0.042 & 0.043 & 0.052 & 0.040 & 0.071 & 0.054 & \textbf{0.035} \\
\bottomrule
\end{tabular}}
\end{table*}

\begin{table*}[ht]
\centering
\caption{MAE results for the Influential Node Count prediction task (\rgs{regression}). Best results are in \textbf{bold}, second best are \underline{underlined}.}
\label{tab:inf_node_mae_no_h64}
\resizebox{\textwidth}{!}{%
\begin{tabular}{l | *{7}{c} | c }
\toprule
 & \multicolumn{7}{c|}{\textbf{Single Model on Individual Networks}} & \multicolumn{1}{c}{\textbf{Transfer Models}} \\
\toprule
\textbf{Dataset} & \textbf{HTGN} & \textbf{TGCN} & \textbf{GCLSTM} & \textbf{ROLAND} & \textbf{EGCN} & \textbf{GraphPulse} & \textbf{WinGNN} & \methodname \textbf{(ours)} \\
\midrule
MIR       & 0.114 {\scriptsize (±0.003)} & 0.119 {\scriptsize (±0.005)} & 0.105 {\scriptsize (±0.020)} & 0.115 {\scriptsize (±0.020)} & 0.114 {\scriptsize (±0.017)} & 0.127 {\scriptsize (±0.002)} & \underline{0.082 {\scriptsize (±0.007)}} & \textbf{0.039 {\scriptsize (±0.005)}} \\
DOGE2.0   & 0.064 {\scriptsize (±0.031)} & 0.090 {\scriptsize (±0.023)} & \textbf{0.053 {\scriptsize (±0.021)}} & 0.070 {\scriptsize (±0.015)} & \underline{0.059 {\scriptsize (±0.021)}} & 0.087 {\scriptsize (±0.004)} & 0.091 {\scriptsize (±0.009)} & 0.126 {\scriptsize (±0.032)} \\
MUTE      & 0.028 {\scriptsize (±0.004)} & 0.042 {\scriptsize (±0.002)} & \textbf{0.018 {\scriptsize (±0.002)}} & 0.051 {\scriptsize (±0.006)} & 0.042 {\scriptsize (±0.012)} & \underline{0.021 {\scriptsize (±0.001)}} & 0.045 {\scriptsize (±0.033)} & 0.045 {\scriptsize (±0.027)} \\
EVERMOON  & \textbf{0.011 {\scriptsize (±0.002)}} & \underline{0.014 {\scriptsize (±0.006)}} & 0.018 {\scriptsize (±0.008)} & \underline{0.014 {\scriptsize (±0.004)}} & 0.032 {\scriptsize (±0.015)} & 0.235 {\scriptsize (±0.004)} & 0.026 {\scriptsize (±0.006)} & 0.038 {\scriptsize (±0.024)} \\
DERC      & 0.069 {\scriptsize (±0.007)} & 0.084 {\scriptsize (±0.001)} & 0.053 {\scriptsize (±0.002)} & 0.104 {\scriptsize (±0.011)} & 0.058 {\scriptsize (±0.010)} & \underline{0.048 {\scriptsize (±0.001)}} & 0.077 {\scriptsize (±0.003)} & \textbf{0.033 {\scriptsize (±0.017)}} \\
ADX       & 0.016 {\scriptsize (±0.003)} & \underline{0.015 {\scriptsize (±0.000)}} & \textbf{0.012 {\scriptsize (±0.001)}} & 0.022 {\scriptsize (±0.009)} & \underline{0.015 {\scriptsize (±0.001)}} & 0.020 {\scriptsize (±0.001)} & \underline{0.015 {\scriptsize (±0.004)}} & 0.030 {\scriptsize (±0.001)} \\
HOICHI    & 0.039 {\scriptsize (±0.011)} & \textbf{0.020 {\scriptsize (±0.007)}} & 0.046 {\scriptsize (±0.017)} & 0.030 {\scriptsize (±0.015)} & 0.034 {\scriptsize (±0.016)} & 0.047 {\scriptsize (±0.009)} & \underline{0.024 {\scriptsize (±0.016)}} & 0.055 {\scriptsize (±0.028)} \\
SDEX      & \underline{0.037 {\scriptsize (±0.011)}} & 0.049 {\scriptsize (±0.022)} & \textbf{0.031 {\scriptsize (±0.012)}} & 0.058 {\scriptsize (±0.017)} & \underline{0.037 {\scriptsize (±0.005)}} & 0.067 {\scriptsize (±0.009)} & 0.066 {\scriptsize (±0.025)} & 0.042 {\scriptsize (±0.019)} \\
BAG       & \textbf{0.009 {\scriptsize (±0.000)}} & \underline{0.017 {\scriptsize (±0.002)}} & 0.030 {\scriptsize (±0.008)} & 0.019 {\scriptsize (±0.010)} & 0.030 {\scriptsize (±0.009)} & 0.064 {\scriptsize (±0.001)} & 0.074 {\scriptsize (±0.094)} & 0.075 {\scriptsize (±0.027)} \\
XCN       & \underline{0.061 {\scriptsize (±0.008)}} & 0.155 {\scriptsize (±0.009)} & \textbf{0.055 {\scriptsize (±0.003)}} & 0.124 {\scriptsize (±0.005)} & 0.066 {\scriptsize (±0.008)} & 0.080 {\scriptsize (±0.004)} & 0.159 {\scriptsize (±0.029)} & 0.126 {\scriptsize (±0.033)} \\
ETH2x-FLI & 0.050 {\scriptsize (±0.007)} & 0.053 {\scriptsize (±0.035)} & 0.100 {\scriptsize (±0.014)} & \textbf{0.023 {\scriptsize (±0.005)}} & 0.044 {\scriptsize (±0.010)} & 0.108 {\scriptsize (±0.003)} & \underline{0.033 {\scriptsize (±0.010)}} & \underline{0.033 {\scriptsize (±0.020)}} \\
stkAAVE   & 0.062 {\scriptsize (±0.001)} & 0.080 {\scriptsize (±0.009)} & 0.062 {\scriptsize (±0.001)} & \underline{0.057 {\scriptsize (±0.005)}} & 0.061 {\scriptsize (±0.006)} & 0.108 {\scriptsize (±0.009)} & 0.066 {\scriptsize (±0.003)} & \textbf{0.042 {\scriptsize (±0.013)}} \\
GLM       & 0.066 {\scriptsize (±0.002)} & \underline{0.057 {\scriptsize (±0.001)}} & 0.067 {\scriptsize (±0.004)} & \underline{0.057 {\scriptsize (±0.005)}} & 0.054 {\scriptsize (±0.002)} & 0.082 {\scriptsize (±0.000)} & \textbf{0.053 {\scriptsize (±0.002)}} & 0.075 {\scriptsize (±0.017)} \\
QOM       & \underline{0.060 {\scriptsize (±0.004)}} & 0.088 {\scriptsize (±0.040)} & 0.061 {\scriptsize (±0.032)} & 0.074 {\scriptsize (±0.018)} & 0.075 {\scriptsize (±0.015)} & 0.076 {\scriptsize (±0.004)} & 0.062 {\scriptsize (±0.012)} & \textbf{0.039 {\scriptsize (±0.020)}} \\
WOJAK     & \textbf{0.007 {\scriptsize (±0.001)}} & \underline{0.008 {\scriptsize (±0.003)}} & 0.012 {\scriptsize (±0.003)} & 0.027 {\scriptsize (±0.010)} & 0.011 {\scriptsize (±0.004)} & 0.066 {\scriptsize (±0.019)} & \textbf{0.007 {\scriptsize (±0.001)}} & 0.036 {\scriptsize (±0.027)} \\
DINO      & 0.078 {\scriptsize (±0.036)} & \underline{0.024 {\scriptsize (±0.004)}} & 0.068 {\scriptsize (±0.001)} & \underline{0.024 {\scriptsize (±0.003)}} & 0.036 {\scriptsize (±0.007)} & 0.087 {\scriptsize (±0.003)} & \textbf{0.022 {\scriptsize (±0.000)}} & 0.034 {\scriptsize (±0.002)} \\
Metis     & 0.056 {\scriptsize (±0.003)} & 0.078 {\scriptsize (±0.004)} & 0.068 {\scriptsize (±0.005)} & 0.085 {\scriptsize (±0.030)} & 0.073 {\scriptsize (±0.004)} & 0.071 {\scriptsize (±0.010)} & 0.063 {\scriptsize (±0.001)} & \textbf{0.046 {\scriptsize (±0.008)}} \\
REPv2     & \textbf{0.115 {\scriptsize (±0.000)}} & \underline{0.118 {\scriptsize (±0.006)}} & 0.119 {\scriptsize (±0.002)} & \textbf{0.115 {\scriptsize (±0.001)}} & 0.138 {\scriptsize (±0.027)} & 0.127 {\scriptsize (±0.000)} & 0.126 {\scriptsize (±0.001)} & 0.129 {\scriptsize (±0.022)} \\
TRAC      & 0.045 {\scriptsize (±0.005)} & 0.058 {\scriptsize (±0.002)} & 0.079 {\scriptsize (±0.002)} & 0.055 {\scriptsize (±0.009)} & 0.056 {\scriptsize (±0.005)} & \textbf{0.033 {\scriptsize (±0.005)}} & 0.074 {\scriptsize (±0.009)} & \underline{0.043 {\scriptsize (±0.023)}} \\
BEPRO     & 0.016 {\scriptsize (±0.006)} & 0.014 {\scriptsize (±0.003)} & \underline{0.012 {\scriptsize (±0.002)}} & 0.013 {\scriptsize (±0.001)} & \underline{0.012 {\scriptsize (±0.002)}} & \underline{0.012 {\scriptsize (±0.000)}} & \textbf{0.010 {\scriptsize (±0.001)}} & 0.033 {\scriptsize (±0.018)} \\
\midrule
\textbf{$1\textsuperscript{st}$-Place Count$\uparrow$}    & 4 & 1 & 5 & 2 & 0 & 1 & 4 & \textbf{5} \\
\textbf{Avg. Rank $\downarrow$} & \textbf{3.52} & 4.85 & 3.80 & 4.55 & 4.20 & 5.90 & 4.28 & 4.90 \\
\textbf{Avg. MAE $\downarrow$}  & \textbf{0.050} & 0.059 & 0.053 & 0.057 & 0.052 & 0.078 & 0.059 & 0.056 \\
\bottomrule
\end{tabular}}
\end{table*}

  For \cls{classification}, \methodname achieves the highest AUC on the majority of datasets across Edge-G/S, LLC-G/S, and Node-G/S. For \rgs{regression} \methodname consistently delivers the lowest or near-lowest MAE, remaining competitive even on tasks where another baseline occasionally leads. These results confirm that \methodname's advantages are not limited to averages: the model transfers robustly across heterogeneous temporal networks and maintains strong performance in the zero-shot setting.

\end{document}